\documentclass{article}

\usepackage{linguex}
\usepackage{harvard}
\usepackage{amsmath}
\usepackage{amssymb}
\usepackage{latexsym}
\usepackage{graphicx}
\usepackage{proof}
\usepackage{tikz}
\usetikzlibrary{arrows,matrix,graphs,shapes,snakes,automata,backgrounds,petri,positioning,calc,decorations.pathmorphing}
\usepackage{hyperref}

\tikzstyle{reverseclip}=[insert path={(current page.north east) --
  (current page.south east) --
  (current page.south west) --
  (current page.north west) --
  (current page.north east)}
]
\tikzset{pas/.style={fill=gray!60}, 
act/.style={fill=gray!30},
main/.style={draw,fill=white},
ctx/.style={rounded rectangle,minimum size=7mm},
val/.style={rectangle,minimum size=7mm},
cmd/.style={chamfered rectangle,draw,fill=white},
tns/.style={circle,minimum size=4mm,draw,fill=white},
par/.style={circle,minimum size=4mm,draw,fill=black}, 
minipar/.style={circle,minimum size=2.5mm,draw,fill=black}, 
pn/.style={rounded corners, rectangle,fill=blue!30,draw,minimum size=12mm},
gpn/.style={rounded corners, rectangle,fill=gray!30,draw,minimum size=12mm},
medpn/.style={rounded corners, rectangle,fill=blue!30,draw,minimum size=20mm},
bigpn/.style={rounded corners, rectangle,fill=blue!30,draw,minimum size=25mm}}

\newcommand{\editout}[1]{}

\colorlet{lightgray}{gray!10}

\newcommand{\apsnodei}{\centerdot_{\rule{0pt}{1.2ex}}}
\newcommand{\ldr}{\mathbin{/}}
\newcommand{\ldl}{\mathbin{\backslash}}

\newcommand{\grow}[1]{\phantom{(}#1\phantom{)}}

\newcommand{\appl}{\ensuremath{\scriptsize @}}

\newcommand{\ats}{\textit{s}}
\newcommand{\atn}{\textit{n}}
\newcommand{\atnp}{\textit{np}}
\newcommand{\atpp}[1]{\textit{pp}_{\textit{#1}}}

\newcommand{\proofspace}{\vphantom{()}}
\newcommand{\atsr}{\textit{s}\vphantom{()}}
\newcommand{\atnr}{\textit{n}\vphantom{()}}
\newcommand{\atnpr}{\textit{np}\vphantom{()}}
\newcommand{\atppr}[1]{\textit{pp}_{\textit{#1}}\vphantom{()}}

\newcommand{\blb}{\phantom{M}}

\title{Perspectives on neural proof nets}
\author{Richard Moot}
\date{CNRS, LIRMM, Univerist\'{e} de Montpellier}

\begin{document}
\maketitle

\section{Introduction}
Proof nets are a way of representing proofs as a type of (hyper)graph. Originally introduced for linear logic \cite{girard}, proof nets can be seen as a parallelised sequent calculus which removes inessential rule permutations, but also as a multi-conclusion natural deduction which simplifies many of the logical rules (notably the $\multimap E$, $\bullet E$, $\Diamond E$ rules). This make proof nets a good choice for automated theorem proving: avoiding needless rule permutations entails an important reduction of the search space for proofs (compared to sequent calculus, and to a somewhat lesser extent when compared to natural deduction) but still allows us to compute the lambda terms corresponding to our proofs: enumerating all different proof nets for a sequent is equivalent to enumerating all its different lambda terms.

Proof nets can be adapted to different types of type-logical grammars while preserving their good logical properties \cite{moot2021}. This makes them an important tool for testing the predictions of different grammars written in type-logical formalisms. Combining the lambda terms produced by proof search with lambda terms assigned to words in the lexicon places type-logical grammars in the formal semantics tradition initiated by Montague \cite{montague,M95}.

In this paper I will present two different ways of combining proof net proof search with neural networks, using two different ways to split the task into two subtasks. The first approach is the `standard' approach which has been applied to proof search in type-logical grammars in various different forms \cite{kogkalidis2020neural,deepgrail23}. Since this approach has been discussed in other places, I will only present it briefly as a way to contrast it with the second, novel approach, which is the main topic of the paper.

The rest of this paper is structured as follows. Section~\ref{sec:pn} presents the standard view on proof nets for natural language processing, and introduces proof nets in the style of \citeasnoun{mp}, which are quite close to natural deduction proofs, but which allow structures to have multiple conclusions. Section~\ref{sec:npn} shows how the standard approach maps nicely to a neural network architecture: following \citeasnoun{kogkalidis2020neural}, we first map words to formulas, then use a neural network to find the correct matching between atomic formulas in a way similar to a resolution proof. Section~\ref{sec:alt} proposes an alternative way to split the task into two: first we generate the graph structure in a way which guarantees it corresponds to a lambda-term, then we obtain the detailed structure using vertex labelling. Vertex labelling is a well-studied task in graph neural networks, so most of Section~\ref{sec:alt} will be about ways to implement graph generation using neural networks. Some work is needed to fit this task into one of the standard graph neural network paradigms but two possibilities are explored (with a third in Appendix~\ref{app:backward}). Section~\ref{sec:conc} concludes. Unfortunately, we need to leave the implementation and evaluation of these new strategies to future research.

 \section{Proof nets}
\label{sec:pn}

The `standard' approach to using proof nets for natural language processing is the following.

\begin{enumerate}
	\item \emph{Lexical lookup} The lexicon maps words to formulas. In general, a word can have multiple formulas assigned to it, so this step is fundamentally non-deterministic.
	\item \emph{Unfold} We write down the formula decomposition tree.
	\item \emph{Match} We enumerate the 1-1 matchings between atomic formulas of opposite polarity, similar to the way a resolution proof links atoms with their negations.
	\item \emph{Check correctness} We check whether the resulting structure satisfies the correctness conditions.
\end{enumerate}

We will return to these different steps below. This approach is standard because it corresponds naturally to the way we write our grammars in type-logical grammars: the grammar writer provides a lexicon and then tests his grammar on sentences using the words in the lexicon. The task of the theorem prover is then to find all different proofs/lambda terms, and this allows type-logical grammarians to test the predictions of their grammars by computing the different meanings of the sentences in their fragment.

To keep the discussion simple we will start with proof nets for the non-associative Lambek calculus NL. Table~\ref{tab:mmlinks} shows the links for the two implications. 

%Even though there is some variation with the links --- for example, multimodal calculi add labels inside the central circles of the link --- we will mostly ignore these complications since they have little important for what follows.

\begin{table}
\begin{center}
\begin{tikzpicture}[scale=0.75]
% /E
\node (labl) at (11em,8.5em) {$[\ldr E]$};
\node (ab) at (11em,10.8em) {$C$};
\node (a) at (8em,15.6em) {$C/ B$};
\node (aa) at (8.7em,15.2em) {};
\node (b) at (14em,15.75em) {$B$};
\node[tns] (c) at (11em,13.668em) {};
%\node (clab) at (3em,12.668em) {$i$};
\draw (c) -- (ab);
\draw (c) -- (aa);
\draw (c) -- (b);
% /I
\node (labl) at (11em,-2.0em) {$[\ldr I]$};
\node (pa) at (8em,0) {$C/ B$};
\node (pat) at (8.7em,0.44em) {};
\node[par] (pc) at (11em,1.732em) {};
%\node (pclab) at (3em,1.732em) {\textcolor{white}{$i$}};
\node (pb) at (14em,0.15em) {$B$};
\node (pd) at (11em,4.8em) {$C$};
\draw (pc) -- (pb);
\draw (pc) -- (pd);
\path[>=latex,->]  (pc) edge (pat);
% \E
\node (labl) at (23em,8.5em) {$[\ldl E]$};
\node (ab) at (23em,10.8em) {$C$};
\node (a) at (26em,15.6em) {$A\backslash C$};
\node (aa) at (25.3em,15.2em) {};
\node (b) at (20em,15.75em) {$A$};
\node[tns] (c) at (23em,13.668em) {};
%\node (clab) at (23em,12.668em) {$i$};
\draw (c) -- (ab);
\draw (c) -- (aa);
\draw (c) -- (b);
% \I
\node (labl) at (23em,-2.0em) {$[\ldl I]$};
\node (pa) at (26em,0) {$A\backslash C$};
\node (pat) at (25.3em,0.44em) {};
\node[par] (pc) at (23em,1.732em) {};
%\node (pclab) at (23em,1.732em) {\textcolor{white}{$i$}};
\node (pb) at (20em,0.15em) {$A$};
\node (pd) at (23em,4.8em) {$C$};
\draw (pc) -- (pb);
\draw (pc) -- (pd);
\path[>=latex,->]  (pc) edge (pat);
% *E
%\node (labl) at (13em,7.5em) {$[\bullet E]$};
%\node (pa) at (10em,9.8em) {$A$};
%\node (pdt) at (13em,14.3em) {};
%\node[par] (pc) at (13em,11.532em) {};
%\node (pclab) at (13em,11.532em) {\textcolor{white}{$i$}};
%\node (pb) at (16em,9.8em) {$B$};
%\node (pd) at (13em,14.6em) {$A\bullet_i B$};
%\draw (pc) -- (pb);
%\draw (pc) -- (pa);
%\path[>=latex,->]  (pc) edge (pdt);
% *I
%\node (labl) at (13em,-2.0em) {$[\bullet I]$};
%\node (ab) at (13em,0em) {$A\bullet_i B$};
%\node (aba) at (13em,0.3em) {};
%\node (a) at (16em,4.8em) {$B$};
%\node (b) at (10em,4.8em) {$A$};
%\node[tns] (c) at (13em,2.868em) {};
%\node (clab) at (13em,2.868em) {$i$};
%\draw (c) -- (aba);
%\draw (c) -- (a);
%\draw (c) -- (b);
\end{tikzpicture}
\end{center}
\caption{Links for proof nets}
\label{tab:mmlinks}
\end{table}

The table give the names of the corresponding natural deduction rule for each of the links. For the elimination (or \emph{tensor}) links, the correspondence with the modus ponens rule is quite immediate: for $/E$ we have a formula $C/ B$, that is a formula which looks for a $B$ to its right to form a $C$, and we have a formula $B$ to its right, therefore we conclude $C$. Similarly, the formula $A\backslash C$ combines with an $A$ to its left to form a $C$.

The introduction (or \emph{par}) links are somewhat more difficult to explain. We start by remarking that the par links are up-down symmetric with the corresponding tensor link. The way to read the $/ I$ link is that we connect it to a structure with conclusion $C$ and hypothesis $B$ to obtain a structure with conclusion $C/ B$. This way the co-indexing between hypotheses and introduction rule in natural deduction rules is replaced by an edge in the hypergraph.

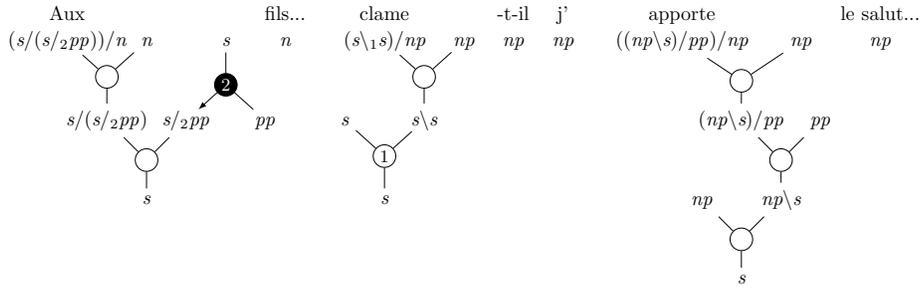
\begin{figure}
\scalebox{0.75}{
\begin{tikzpicture}
\node (auxl) at (10em,11.3em) {Aux\proofspace};
\node (aux) at (10em,10em) {$(\ats/(\ats/_2 \atpp{}))/\atn$};
\node (na) at (14em,10em) {$\atnr$};
\node [tns] (t) at (12em,8.2em) {};
\node (sspp) at (12em,6em) {$\ats/(\ats/_{2} \atpp{} )$};
\node (auxs) at (10em,10em) {\blb};
\draw (t)--(auxs);
\draw (t)--(sspp);
\draw (t)--(na);
\node (spp) at (16em,6em) {$\atsr /_2 \atppr{}$};
\node (auxs) at (14em,2em) {$\ats$};
\node [tns] (tt) at (14em,4.0em) {};
\draw (tt) -- (auxs);
\draw (tt) -- (sspp);
\draw (tt) -- (spp);
\node [par] (par) at (18em,7.8em) {};
\node at (18em,7.8em) {\textcolor{white}{\small 2}};
\node (pp) at (20em,6em) {$\atppr{}$};
\node (ppd) at (20em,6em) {\phantom{x}};
\node (tops) at (18em,10em) {$\ats$};
\draw (par)--(ppd);
\draw (par)--(tops);
\node (sppd) at (16em,6em) {\phantom{x}};
\path[>=latex,->]  (par) edge (sppd);
%%%%
\node (fils) at (21em,10em) {$\atnr$};
\node (filsl) at (21em,11.3em) {fils...\proofspace};
%%%
\node (clamel) at (26em,11.3em) {clame\proofspace};
\node (hs) at (24em,6em) {$\atsr$};
\node (ss) at (28em,6em) {$\atsr \backslash \ats$};
\node (ssd) at (28em,6em) {\phantom{x}};
\node (cs) at (26em,2em) {$\ats$};
\node [tns] (t) at (26em,4.2em){};
\draw (t) -- (hs);
\draw (t) -- (ssd);
\draw (t) -- (cs);
\node (clsu) at (30em,10em) {$\atnpr$};
\node (clsud) at (30em,10em) {\phantom{x}};
\node (clame) at (26em,10em) {$(\ats\backslash_1 \ats)/\atnp$};
\node (clamed) at (26em,10em) {\phantom{M}};
\node [tns] (tc) at (28em,8.2em) {};
\node at (26em,4.2em) {\small 1};
\draw (tc) -- (clamed);
\draw (tc) -- (clsud);
\draw (tc) -- (ss);
%%%
\node (il) at (32.5em,10em) {$\atnpr$};
\node at (32.5em,11.3em) {-t-il\proofspace};
%%%
\node (je) at (35em,10em) {$\atnpr$};
\node at (35em,11.3em) {j'\proofspace};
%%%
\node at (41em,11.3em) {apporte\proofspace};
\node (apporte) at (41em,10em) {$((\atnp\backslash\ats)/\atpp{})/\atnp$};
\node (apob) at (47em,10em) {$\atnpr$};
\node (tvp) at (44em,6em) {$(\atnp\backslash\ats)/\atpp{}$};
\node (appp) at (48em,6em) {$\atppr{}$};
\node (vp) at (46em,2em) {$\atnpr\backslash \ats$};
\node (apsu) at (42em,2em) {$\atnpr$};
\node (apc) at (44em,-2em) {$\ats$};
\node [tns] (t1) at (44em,8em) {};
\draw (t1) -- (apporte);
\draw (t1) -- (apob);
\draw (t1) -- (tvp);
\node [tns] (t2) at (46em,4.0em) {};
\draw (t2) -- (vp);
\draw (t2) -- (tvp);
\draw (t2) -- (appp);
\node [tns] (t3) at (44em,0em) {};
\draw (t3) -- (apc);
\draw (t3) -- (vp);
\draw (t3) -- (apsu);
\node at (51em,11.3em) {le salut...\proofspace};
\node (lesalut) at (51em,10em) {$\atnpr$};
\end{tikzpicture}}
\caption{Formula unfolding}
\label{fig:unfold}
\end{figure}

Figure~\ref{fig:unfold} shows a (slightly simplified) example unfolding for the sentence.

\exg. Aux fils de la R\'{e}volution mexicaine, claime-t-il, j'apport le salut fraternel des fils de la R\'{e}volution fran\c{c}aise. \\
{To the} sons of the revolution Mexican, {he exclaimed}, {I bring} the salvation brotherly {of the} sons of the revolution French. \\
To the sons of the Mexican revolution, he exclaimed, I bring the brotherly salvation of the sons of the French revolution.

In the simplified version of Figure~\ref{fig:unfold}, the noun ``fils de la R\'{e}volution mexicaine'' has been abbreviated as ``fils...''. Similarly, ``le salut fraternel des fils de la R\'{e}volution fran\c{c}aise'' has been abbreviated as the noun phrase ``le salut...''. 

We can verify that each local neighbourhood is an instantiation of one of the links from Table~\ref{tab:mmlinks}. 
Given a formula, like $((np\backslash s)/pp)/np$ for ``apporte'' the unfolding is completely deterministic. We simply apply the logical links until we arrive at the atomic formulas. 
Some of the links have labels inside of the central circle  --- $2$ for the $/ I$ link of ``Aux'', $1$ for the bottom most $\backslash E$ link of ``clame''. 
 These labels are ways to mark certain connections as special; technically, they allows us to introduce controlled versions of structural rules for so-called multimodal versions of the Lambek calculus \cite{moor:mult95} and these extensions can be incorporated into the proof net calculus we present here without problems \cite{mp}. However, the complete formal development of multimodal proof nets is not important for what follows, and the only thing to remember is that the labels can permit some additional operations.

\begin{figure}
\begin{center}
\scalebox{0.75}{
\begin{tikzpicture}
\node (auxl) at (10em,11.0em) {Aux\proofspace};
\node (aux) at (10em,10em) {$\apsnodei$};
\node (na) at (14em,10em) {$\atnr$};
\node [tns] (t) at (12em,8.2em) {};
\node (sspp) at (12em,6em) {$\apsnodei$};
\node (auxs) at (10em,10em) {\blb};
\draw (t)--(auxs);
\draw (t)--(sspp);
\draw (t)--(na);
\node (spp) at (16em,6em) {$\apsnodei$};
\node (auxs) at (14em,2em) {$\ats_1$};
\node [tns] (tt) at (14em,4.0em) {};
\draw (tt) -- (auxs);
\draw (tt) -- (sspp);
\draw (tt) -- (spp);
\node [par] (par) at (18em,7.8em) {};
\node at (18em,7.8em) {\textcolor{white}{\small 2}};
\node (pp) at (20em,6em) {$\atppr{}$};
\node (ppd) at (20em,6em) {\phantom{x}};
\node (tops) at (18em,10em) {$\ats_2$};
\draw (par)--(ppd);
\draw (par)--(tops);
\node (sppd) at (16em,6em) {\phantom{x}};
\path[>=latex,->]  (par) edge (sppd);
%%%%
\node (fils) at (21em,10em) {$\atnr$};
\node (filsl) at (21em,11.0em) {fils...\proofspace};
%%%
\node (clamel) at (26em,11.0em) {clame\proofspace};
\node (hs) at (24em,6em) {$\ats_3$};
\node (ss) at (28em,6em) {$\apsnodei$};
\node (ssd) at (28em,6em) {\phantom{x}};
\node (cs) at (26em,2em) {$\ats_4$};
\node [tns] (t) at (26em,4.2em){};
\draw (t) -- (hs);
\draw (t) -- (ssd);
\draw (t) -- (cs);
\node (clsu) at (30em,10em) {$\atnp_1$};
\node (clsud) at (30em,10em) {\phantom{x}};
\node (clame) at (26em,10em) {$\apsnodei$};
\node (clamed) at (26em,10em) {\phantom{M}};
\node [tns] (tc) at (28em,8.2em) {};
\node at (26em,4.2em) {\small 1};
\draw (tc) -- (clamed);
\draw (tc) -- (clsud);
\draw (tc) -- (ss);
%%%
\node (il) at (33em,10em) {$\atnp_2$};
\node at (33em,11.0em) {-t-il\proofspace};
%%%
\node (je) at (36em,10em) {$\atnp_3$};
\node at (36em,11.0em) {j'\proofspace};
%%%
\node at (40em,11.0em) {apporte\proofspace};
\node (apporte) at (40em,10em) {$\apsnodei$};
\node (apob) at (44em,10em) {$\atnp_5$};
\node (tvp) at (42em,6em) {$\apsnodei$};
\node (appp) at (46em,6em) {$\atppr{}$};
\node (vp) at (44em,2em) {$\apsnodei$};
\node (apsu) at (40em,2em) {$\atnp_4$};
\node (apc) at (42em,-2em) {$\ats_5$};
\node [tns] (t1) at (42em,8em) {};
\draw (t1) -- (apporte);
\draw (t1) -- (apob);
\draw (t1) -- (tvp);
\node [tns] (t2) at (44em,4.0em) {};
\draw (t2) -- (vp);
\draw (t2) -- (tvp);
\draw (t2) -- (appp);
\node [tns] (t3) at (42em,0em) {};
\draw (t3) -- (apc);
\draw (t3) -- (vp);
\draw (t3) -- (apsu);
\node at (49em,11.0em) {le salut...\proofspace};
\node (lesalut) at (49em,10em) {$\atnp_6$};
\node at (53em,10em) {$\ats_6$};
\node at (53em,9em) {Goal};
\end{tikzpicture}}
\scalebox{0.75}{
\begin{tikzpicture}
\draw (0em,0em) -- (0em,6.2em);
\draw (2em,0em) -- (2em,6.2em);
\draw (4em,0em) -- (4em,6.2em);
\draw (6em,0em) -- (6em,6.2em);
\draw (-0.2em,0em) -- (6em,0em);
\draw (-0.2em,2em) -- (6em,2em);
\draw (-0.2em,4em) -- (6em,4em);
\draw (-0.2em,6em) -- (6em,6em);
\draw [fill=gray!40] (0em,4em) -- (0em,6em) --(2em,6em) -- (2em,4em) -- cycle;
\draw [fill=gray!40] (2em,2em) -- (2em,4em) --(4em,4em) -- (4em,2em) -- cycle;
\draw [fill=gray!40] (4em,0em) -- (4em,2em) --(6em,2em) -- (6em,0em) -- cycle;
\draw (10em,0em) -- (10em,6.2em);
\draw (9.8em,0em) -- (16em,0em);
\draw (12em,0em) -- (12em,6.2em);
\draw (14em,0em) -- (14em,6.2em);
\draw (16em,0em) -- (16em,6.2em);
\draw (9.8em,2em) -- (16em,2em);
\draw (9.8em,4em) -- (16em,4em);
\draw (16em,0em) -- (16em,6em);
\draw (9.8em,6em) -- (16em,6em);	
\node at (-1.0em,5em) {$\atnp_2$};
\node at (-1.0em,3em) {$\atnp_3$};
\node at (-1.0em,1em) {$\atnp_6$};
\node at (1em,7em) {$\atnp_1$};
\node at (3em,7em) {$\atnp_4$};
\node at (5em,7em) {$\atnp_5$};
\node at (9.0em,5em) {$\ats_1$};
\node at (9.0em,3em) {$\ats_4$};
\node at (9.0em,1em) {$\ats_5$};
\node at (11em,7em) {$\ats_2$};
\node at (13em,7em) {$\ats_3$};
\node at (15em,7em) {$\ats_6$};
\end{tikzpicture}}
\end{center}
\caption{Abstract proof structure of Figure~\ref{fig:unfold}, with matrices representing the possible $\atnp$ and $\ats$ matchings.}
\label{fig:abstract}
\end{figure}
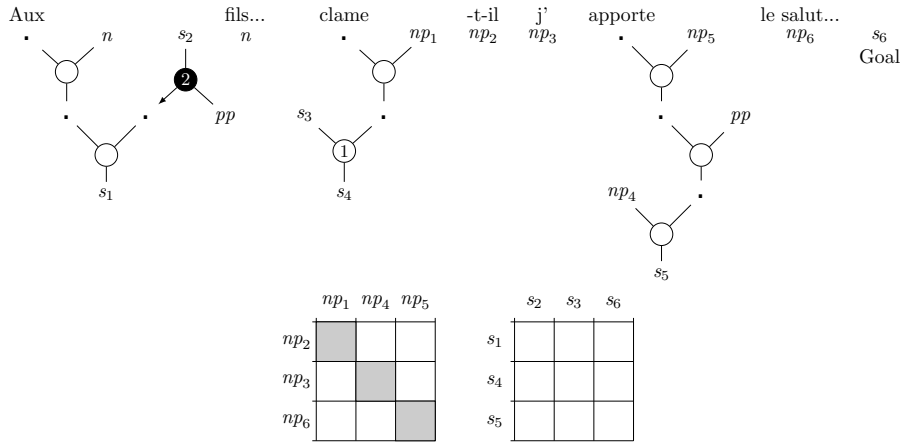

We can simplify Figure~\ref{fig:unfold} further by removing all complex formulas from the vertices, leaving only the atomic formulas. This simplification is information preserving: given that the `root' of each component is indicated by the corresponding word, we can determine the corresponding link for each local neighbourhood, then compute the formulas from the atomic leaves up towards the lexical root. The structure shown in Figure~\ref{fig:abstract} is called an abstract proof structure. In the figure, we have explicitly added the goal formula $\ats$ as well. In addition, the atomic $\atnp$ and $\ats$ formulas have been given integer subscripts. These subscripts are not formally part of the structure and are only there to give us a simple way to refer to the different atomic formulas. 

Proof search now consists of matching the negative formulas --- formulas which are not connected from below such as the formula $\atnp_2$ of ``-t-il'' and $\atn$ of ``fils...'' --- with the positive formulas --- formulas which are not connected from the top, such as $\atnp_4$ and $\atnp_5$ of ``apporte'' and the $\atnp_1$ formula of ``clame''. The two square matrices on the bottom of Figure~\ref{fig:abstract} summarise the possible matchings for the $\atnp$ and $\ats$ formulas (the $\atn$ and $\atpp{}$ formulas have only a single possibility in the structure). The negative (conclusion) atomic formulas are shown at the row labels, the positive (hypothesis) atomic formulas are shown at the column labels. A valid matching is a 1-1 matching between rows (negative atoms) and columns (positive atoms). 

The simplification of ``fils...'' and ``le salut...''  reduces the combinatorics of proof search by quite a bit. For example, there are only two nouns left in the structure, the one of ``fils...'' and the one of ``Aux'' so there is only one way to connect the $n$ formulas, as opposed to $11! = 39\,916\,800$ for the complete sentence. We can similarly use the information about the word order to connect the noun phrase of ``-t-il'' to ``clame'' and the two noun phrases ``j'' and ``le salut...'' to ``apporte'' as subject and object respectively. We should be careful about this kind of identification in the presence of structural rules, which can change the structure of the trees: we will see later how such a rewrite will allow us to put ``clame-t-il'' in the desired position. 

The cells shaded grey (bottom left of Figure~\ref{fig:abstract}) show the correct vertex identifications for the $\atnp$ formulas: vertex $\atnp_1$ with vertex $\atnp_2$, $\atnp_3$ with $\atnp_4$ and $\atnp_5$ with $\atnp_6$.

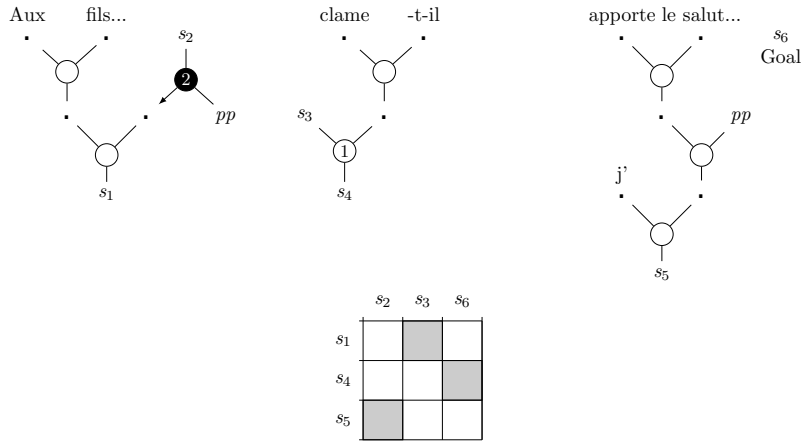
\begin{figure}
\begin{center}
\scalebox{0.75}{
\begin{tikzpicture}
\node (auxl) at (10em,11.0em) {Aux\proofspace};
\node (aux) at (10em,10em) {$\apsnodei$};
\node (na) at (14em,10em) {$\apsnodei$};
\node [tns] (t) at (12em,8.2em) {};
\node (sspp) at (12em,6em) {$\apsnodei$};
\node (auxs) at (10em,10em) {\blb};
\draw (t)--(auxs);
\draw (t)--(sspp);
\draw (t)--(na);
\node (spp) at (16em,6em) {$\apsnodei$};
\node (auxs) at (14em,2em) {$\ats_1$};
\node [tns] (tt) at (14em,4.0em) {};
\draw (tt) -- (auxs);
\draw (tt) -- (sspp);
\draw (tt) -- (spp);
\node [par] (par) at (18em,7.8em) {};
\node at (18em,7.8em) {\textcolor{white}{\small 2}};
\node (pp) at (20em,6em) {$\atppr{}$};
\node (ppd) at (20em,6em) {\phantom{x}};
\node (tops) at (18em,10em) {$\ats_2$};
\draw (par)--(ppd);
\draw (par)--(tops);
\node (sppd) at (16em,6em) {\phantom{x}};
\path[>=latex,->]  (par) edge (sppd);
%%%%
%\node (fils) at (21em,10em) {$\atnr$};
\node (filsl) at (14em,11.0em) {fils...\proofspace};
%%%
\node (clamel) at (26em,11.0em) {clame\proofspace};
\node (hs) at (24em,6em) {$\ats_3$};
\node (ss) at (28em,6em) {$\apsnodei$};
\node (ssd) at (28em,6em) {\phantom{x}};
\node (cs) at (26em,2em) {$\ats_4$};
\node [tns] (t) at (26em,4.2em){};
\draw (t) -- (hs);
\draw (t) -- (ssd);
\draw (t) -- (cs);
\node (clsu) at (30em,10em) {$\apsnodei$};
\node (clsud) at (30em,10em) {\phantom{x}};
\node (clame) at (26em,10em) {$\apsnodei$};
\node (clamed) at (26em,10em) {\phantom{M}};
\node [tns] (tc) at (28em,8.2em) {};
\node at (26em,4.2em) {\small 1};
\draw (tc) -- (clamed);
\draw (tc) -- (clsud);
\draw (tc) -- (ss);
%%%
%\node (il) at (33em,10em) {$\atnpr$};
\node at (30em,11.0em) {-t-il\proofspace};
%%%
%\node (je) at (36em,10em) {$\atnpr$};
\node at (40em,3.0em) {j'\proofspace};
%%%
\node at (40em,11.0em) {apporte\proofspace};
\node (apporte) at (40em,10em) {$\apsnodei$};
\node (apob) at (44em,10em) {$\apsnodei$};
\node (tvp) at (42em,6em) {$\apsnodei$};
\node (appp) at (46em,6em) {$\atppr{}$};
\node (vp) at (44em,2em) {$\apsnodei$};
\node (apsu) at (40em,2em) {$\apsnodei$};
\node (apc) at (42em,-2em) {$\ats_5$};
\node [tns] (t1) at (42em,8em) {};
\draw (t1) -- (apporte);
\draw (t1) -- (apob);
\draw (t1) -- (tvp);
\node [tns] (t2) at (44em,4.0em) {};
\draw (t2) -- (vp);
\draw (t2) -- (tvp);
\draw (t2) -- (appp);
\node [tns] (t3) at (42em,0em) {};
\draw (t3) -- (apc);
\draw (t3) -- (vp);
\draw (t3) -- (apsu);
\node at (44em,11.0em) {le salut...\proofspace};
\node at (48em,10em) {$\ats_6$};
\node at (48em,9em) {Goal};
%\node (lesalut) at (49em,10em) {$\atnpr$};
\end{tikzpicture}}	
\scalebox{0.75}{
\begin{tikzpicture}
\draw [fill=gray!40] (12em,4em) -- (12em,6em) --(14em,6em) -- (14em,4em) -- cycle;
\draw [fill=gray!40] (14em,2em) -- (14em,4em) --(16em,4em) -- (16em,2em) -- cycle;
\draw [fill=gray!40] (10em,0em) -- (10em,2em) --(12em,2em) -- (12em,0em) -- cycle;
\draw (10em,0em) -- (10em,6.2em);
\draw (9.8em,0em) -- (16em,0em);
\draw (12em,0em) -- (12em,6.2em);
\draw (14em,0em) -- (14em,6.2em);
\draw (16em,0em) -- (16em,6.2em);
\draw (9.8em,2em) -- (16em,2em);
\draw (9.8em,4em) -- (16em,4em);
\draw (16em,0em) -- (16em,6em);
\draw (9.8em,6em) -- (16em,6em);	
\node at (9.0em,5em) {$\ats_1$};
\node at (9.0em,3em) {$\ats_4$};
\node at (9.0em,1em) {$\ats_5$};
\node at (11em,7em) {$\ats_2$};
\node at (13em,7em) {$\ats_3$};
\node at (15em,7em) {$\ats_6$};
\end{tikzpicture}}
\end{center}
	\caption{Abstract proof structure of Figure~\ref{fig:abstract} after identification of the nouns and noun phrases.}
	\label{fig:nnp}
\end{figure}

\begin{figure}	
\begin{center}
\!\!\!\!\!\hspace{-3.0ex}
\scalebox{0.75}{
\begin{tikzpicture}
\node (auxl) at (10em,11.0em) {Aux\proofspace};
\node (aux) at (10em,10em) {$\apsnodei$};
\node (na) at (14em,10em) {$\apsnodei$};
\node [tns] (t) at (12em,8.2em) {};
\node (sspp) at (12em,6em) {$\apsnodei$};
\node (auxs) at (10em,10em) {\blb};
\draw (t)--(auxs);
\draw (t)--(sspp);
\draw (t)--(na);
\node (spp) at (16em,6em) {$\apsnodei$};
\node (auxs) at (14em,2em) {$\ats$};
\node [tns] (tt) at (14em,4.0em) {};
\draw (tt) -- (auxs);
\draw (tt) -- (sspp);
\draw (tt) -- (spp);
\node [par] (par) at (18em,7.8em) {};
\node at (18em,7.8em) {\textcolor{white}{\small 2}};
\node (pp) at (20em,6em) {$\atppr{}$};
\node (ppd) at (20em,6em) {\phantom{x}};
\node (tops) at (18em,10em) {$\apsnodei$};
\draw (par)--(ppd);
\draw (par)--(tops);
\node (sppd) at (16em,6em) {\phantom{x}};
\path[>=latex,->]  (par) edge (sppd);
%%%%
%\node (fils) at (21em,10em) {$\atnr$};
\node (filsl) at (14em,11.0em) {fils...\proofspace};
%%%
\node (clamel) at (26em,11.0em) {clame\proofspace};
\node (hs) at (24em,6em) {$\atsr$};
\node (ss) at (28em,6em) {$\apsnodei$};
\node (ssd) at (28em,6em) {\phantom{x}};
\node (cs) at (26em,2em) {\phantom{$\ats$}};
\node [tns] (t) at (26em,4.2em){};
\draw (t) -- (hs);
\draw (t) -- (ssd);
\draw (t) -- (cs);
\node (clsu) at (30em,10em) {$\apsnodei$};
\node (clsud) at (30em,10em) {\phantom{x}};
\node (clame) at (26em,10em) {$\apsnodei$};
\node (clamed) at (26em,10em) {\phantom{M}};
\node [tns] (tc) at (28em,8.2em) {};
\node at (26em,4.2em) {\small 1};
\draw (tc) -- (clamed);
\draw (tc) -- (clsud);
\draw (tc) -- (ss);
%%%
%\node (il) at (33em,10em) {$\atnpr$};
\node at (30em,11.0em) {-t-il\proofspace};
%%%
%\node (je) at (36em,10em) {$\atnpr$};
\node at (16em,15.0em) {j'\proofspace};
%%%
\node at (16em,23.0em) {apporte\proofspace};
\node (apporte) at (16em,22em) {$\apsnodei$};
\node (apob) at (20em,22em) {$\apsnodei$};
\node (tvp) at (18em,18em) {$\apsnodei$};
\node (appp) at (22em,18em) {$\atppr{}$};
\node (vp) at (20em,14em) {$\apsnodei$};
\node (apsu) at (16em,14em) {$\apsnodei$};
\node (apc) at (18em,10em) {\phantom{x}};
% = 18,10.  -24,-12
%
\node [tns] (t1) at (18em,20em) {};
\draw (t1) -- (apporte);
\draw (t1) -- (apob);
\draw (t1) -- (tvp);
\node [tns] (t2) at (20em,16.0em) {};
\draw (t2) -- (vp);
\draw (t2) -- (tvp);
\draw (t2) -- (appp);
\node [tns] (t3) at (18em,12em) {};
\draw (t3) -- (apc);
\draw (t3) -- (vp);
\draw (t3) -- (apsu);
\node at (20em,23.0em) {le salut...\proofspace};
%\node (lesalut) at (49em,10em) {$\atnpr$};
\node at (26em,2em) {Goal};
\end{tikzpicture}}
\raisebox{7em}{$\leadsto$}
\scalebox{0.75}{
\begin{tikzpicture}
\node (auxl) at (10em,11.0em) {Aux\proofspace};
\node (aux) at (10em,10em) {$\apsnodei$};
\node (na) at (14em,10em) {$\apsnodei$};
\node [tns] (t) at (12em,8.2em) {};
\node (sspp) at (12em,6em) {$\apsnodei$};
\node (auxs) at (10em,10em) {\blb};
\draw (t)--(auxs);
\draw (t)--(sspp);
\draw (t)--(na);
\node (spp) at (16em,6em) {$\apsnodei$};
\node (auxs) at (14em,2em) {$\ats$};
\node [tns] (tt) at (14em,4.0em) {};
\draw (tt) -- (auxs);
\draw (tt) -- (sspp);
\draw (tt) -- (spp);
\node [par] (par) at (18em,7.8em) {};
\node at (18em,7.8em) {\textcolor{white}{\small 2}};
%\node (pp) at (20em,6em) {$\atppr{}$};
%\node (ppd) at (20em,6em) {\phantom{x}};
\node (tops) at (18em,10em) {$\apsnodei$};
%\draw (par)--(ppd);
\draw (par)--(tops);
\node (sppd) at (16em,6em) {\phantom{x}};
\path[>=latex,->]  (par) edge (sppd);
%%%%
%\node (fils) at (21em,10em) {$\atnr$};
\node (filsl) at (14em,11.0em) {fils...\proofspace};
%%%
\node (clamel) at (26em,11.0em) {clame\proofspace};
\node (hs) at (24em,6em) {$\atsr$};
\node (ss) at (28em,6em) {$\apsnodei$};
\node (ssd) at (28em,6em) {\phantom{x}};
\node (cs) at (26em,2em) {\phantom{$\ats$}};
\node [tns] (t) at (26em,4.2em){};
\draw (t) -- (hs);
\draw (t) -- (ssd);
\draw (t) -- (cs);
\node (clsu) at (30em,10em) {$\apsnodei$};
\node (clsud) at (30em,10em) {\phantom{x}};
\node (clame) at (26em,10em) {$\apsnodei$};
\node (clamed) at (26em,10em) {\phantom{M}};
\node [tns] (tc) at (28em,8.2em) {};
\node at (26em,4.2em) {\small 1};
\draw (tc) -- (clamed);
\draw (tc) -- (clsud);
\draw (tc) -- (ss);
%%%
%\node (il) at (33em,10em) {$\atnpr$};
\node at (30em,11.0em) {-t-il\proofspace};
%%%
%\node (je) at (36em,10em) {$\atnpr$};
\node at (16em,15.0em) {j'\proofspace};
%%%
\node at (16em,23.0em) {apporte\proofspace};
\node (apporte) at (16em,22em) {$\apsnodei$};
\node (apob) at (20em,22em) {$\apsnodei$};
\node (tvp) at (18em,18em) {$\apsnodei$};
\node (appp) at (22em,18em) {$\apsnodei$};
\node (vp) at (20em,14em) {$\apsnodei$};
\node (apsu) at (16em,14em) {$\apsnodei$};
\node (apc) at (18em,10em) {\phantom{x}};
% = 18,10.  -24,-12
%
\node [tns] (t1) at (18em,20em) {};
\draw (t1) -- (apporte);
\draw (t1) -- (apob);
\draw (t1) -- (tvp);
\node [tns] (t2) at (20em,16.0em) {};
\draw (t2) -- (vp);
\draw (t2) -- (tvp);
\draw (t2) -- (appp);
\node [tns] (t3) at (18em,12em) {};
\draw (t3) -- (apc);
\draw (t3) -- (vp);
\draw (t3) -- (apsu);
\node at (20em,23.0em) {le salut...\proofspace};
\draw (appp) to [out=50,in=330] (par);
%\node (lesalut) at (49em,10em) {$\atnpr$};
\node at (26em,2em) {Goal};
\end{tikzpicture}}
\!\!\!\!\!\hspace{-3.0ex}
\end{center}
\caption{Connecting the $pp$ hypothesis}
\label{fig:apspp}
\end{figure}
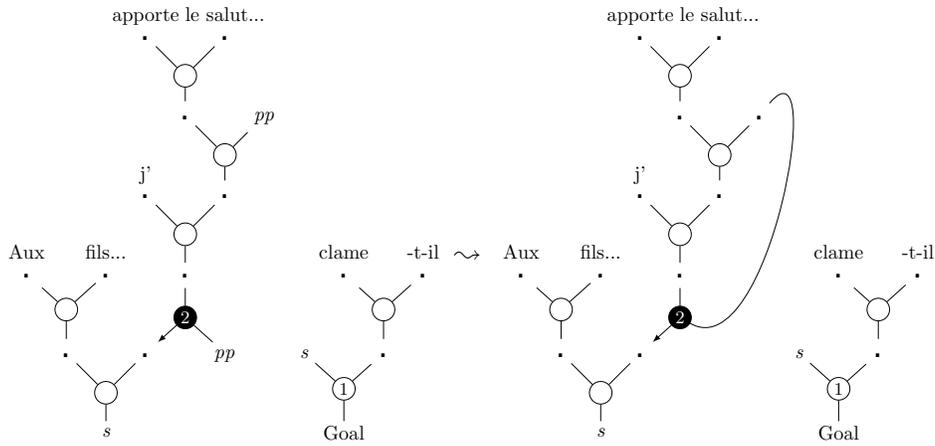

Performing the vertex identifications for the nouns and noun phrases produces the abstract proof structure shown in Figure~\ref{fig:nnp}. For the $\ats$ vertices, there are three possibilities for $\ats_5$ corresponding to the tree ``j'apport le salut $\atpp{}$''. We can identify it with the goal formula $\ats_6$, connect it to the tree for ``clame-t-il'' at $\ats_3$, or to the tree for ``aux fils...'' at $\ats_2$. In the current context, we prefer the identification $\ats_5 - \ats_2$ and the identification $\ats_6 - \ats_4$ as shown on the left of Figure~\ref{fig:apspp}.

For connecting the $pp$ vertices, there is only one possibility, shown on the right of Figure~\ref{fig:apspp}. The curved line plays the same role as the co-indexing of hypotheses and introduction rules in natural deduction proofs. The $pp$ hypothesis in the abstract proof structure on the left of the figure is withdrawn, and the curved line connects it to the introduction rule (represented by the filled circle marked `2') withdrawing it. 

\begin{figure}
\begin{center}
\scalebox{0.75}{
\begin{tikzpicture}
\node (auxl) at (10em,11.0em) {Aux\proofspace};
\node (aux) at (10em,10em) {$\apsnodei$};
\node (na) at (14em,10em) {$\apsnodei$};
\node [tns] (t) at (12em,8.2em) {};
\node (sspp) at (12em,6em) {$\apsnodei$};
\node (auxs) at (10em,10em) {\blb};
\draw (t)--(auxs);
\draw (t)--(sspp);
\draw (t)--(na);
\node (spp) at (16em,6em) {$\apsnodei$};
\node (auxs) at (14em,2em) {$\apsnodei$};
\node [tns] (tt) at (14em,4.0em) {};
\draw (tt) -- (auxs);
\draw (tt) -- (sspp);
\draw (tt) -- (spp);
\node [par] (par) at (18em,7.8em) {};
\node at (18em,7.8em) {\textcolor{white}{\small 2}};
%\node (pp) at (20em,6em) {$\atppr{}$};
%\node (ppd) at (20em,6em) {\phantom{x}};
\node (tops) at (18em,10em) {$\apsnodei$};
%\draw (par)--(ppd);
\draw (par)--(tops);
\node (sppd) at (16em,6em) {\phantom{x}};
\path[>=latex,->]  (par) edge (sppd);
%%%%
%\node (fils) at (21em,10em) {$\atnr$};
\node (filsl) at (14em,11.0em) {fils...\proofspace};
%%%
\node (clamel) at (22em,7.0em) {clame\proofspace};
%\node (hs) at (20em,2em) {$\atsr$};
\node (ss) at (24em,2em) {$\apsnodei$};
\node (ssd) at (24em,2em) {\phantom{x}};
\node (cs) at (19em,-2em) {\phantom{$\atsr$}};
\node [tns] (t) at (19em,0.2em){};
\draw (t) -- (auxs);
\draw (t) -- (ssd);
\draw (t) -- (cs);
\node (clsu) at (26em,6em) {$\apsnodei$};
\node (clsud) at (26em,6em) {$\apsnodei$};
\node (clame) at (22em,6em) {$\apsnodei$};
\node (clamed) at (22em,6em) {\phantom{M}};
\node [tns] (tc) at (24em,4.2em) {};
\node at (19em,0.2em) {\small 1};
\draw (tc) -- (clamed);
\draw (tc) -- (clsud);
\draw (tc) -- (ss);
%%%
%\node (il) at (33em,10em) {$\atnpr$};
\node at (26em,7.0em) {-t-il\proofspace};
%%%
%\node (je) at (36em,10em) {$\atnpr$};
\node at (16em,15.0em) {j'\proofspace};
%%%
\node at (16em,23.0em) {apporte\proofspace};
\node (apporte) at (16em,22em) {$\apsnodei$};
\node (apob) at (20em,22em) {$\apsnodei$};
\node (tvp) at (18em,18em) {$\apsnodei$};
\node (appp) at (22em,18em) {$\apsnodei$};
\node (vp) at (20em,14em) {$\apsnodei$};
\node (apsu) at (16em,14em) {$\apsnodei$};
\node (apc) at (18em,10em) {\phantom{x}};
% = 18,10.  -24,-12
%
\node [tns] (t1) at (18em,20em) {};
\draw (t1) -- (apporte);
\draw (t1) -- (apob);
\draw (t1) -- (tvp);
\node [tns] (t2) at (20em,16.0em) {};
\draw (t2) -- (vp);
\draw (t2) -- (tvp);
\draw (t2) -- (appp);
\node [tns] (t3) at (18em,12em) {};
\draw (t3) -- (apc);
\draw (t3) -- (vp);
\draw (t3) -- (apsu);
\node at (20em,23.0em) {le salut...\proofspace};
\draw (appp) to [out=50,in=330] (par);
%\node (lesalut) at (49em,10em) {$\atnpr$};
\node at (19em,-2em) {Goal};
\end{tikzpicture}}
\end{center}
\caption{Continuing from the abstract proof structure of Figure~\ref{fig:apspp} to obtain the completed linking for the abstract proof structure of Figure~\ref{fig:abstract}}
\label{fig:aps:linked}
\end{figure}
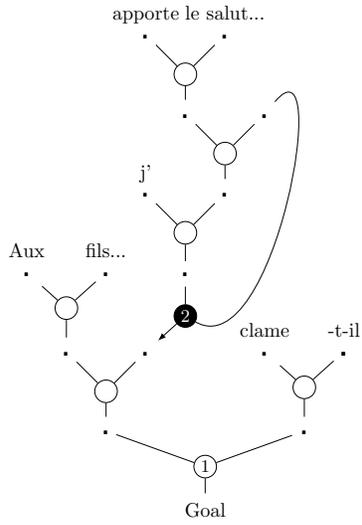

We complete the matching stage by connecting ``clame-t-il'' to ``Aux fils ... j'apport le salut...''. In this matching, ``clame-t-il'' takes the rest of the sentence as its argument, which corresponds to the intended reading for the complete sentence, shown as Figure~\ref{fig:aps:linked}.

It is important to note here that the graph of Figure~\ref{fig:aps:linked} is uniquely determined by the formula unfolding of Figure~\ref{fig:abstract} together with the matching of the atomic formulas, and therefore uniquely determined by just the sequence of formulas together with the matching of the atoms.

\subsection{Contractions}

While proof structures are good for representing the search space in automated theorem proving contexts --- proof search amounts simply to finding matchings of atomic formulas --- proof structures do not necessarily correspond to proofs. Following \citeasnoun{mp}, we will define proof nets as proof structures whose abstract proof structures contract to a tree. 

\begin{figure}
\begin{center}
\hspace{-1em}
\begin{tikzpicture}[scale=0.75]
\node [pn] at (6em,-3.3em) {$\Delta$};
\node [pn] at (6em,12.4em) {$\Gamma$};
\node (ab) at (3em,4.8em) {$\apsnodei$};
\node (a) at (6em,9.6em) {$\apsnodei$};
\node (b) at (0em,9.6em) {$\apsnodei$};
\node[tns] (c) at (3em,7.668em) {};
%\node (clab) at (3em,7.668em) {$i$};
\draw (c) -- (ab);
\draw (c) -- (a);
\draw (c) -- (b);
\node (pa) at (6em,0) {$\apsnodei$};
\node[par] (pc) at (3em,1.732em) {};
%\node (pclab) at (3em,1.732em) {\textcolor{white}{$i$}};
\draw (pc) -- (ab);
\path[>=latex,->]  (pc) edge (pa);
\draw (b) to [out=130,in=210] (pc);
%\node (labl) at (3em,-2.5em) {$[\ldr I]$};
%%%%
\node at (10em,4.8em) {$\rightarrow$};
%%%
\node at (14em,4.8em) {$\apsnodei$};
\node [pn] at (14em,1.8em) {$\Delta$};
\node [pn] at (14em,7.7em) {$\Gamma$};
\end{tikzpicture}
\qquad\qquad 
\begin{tikzpicture}[scale=0.75]
\node [pn] at (0em,-3.3em) {$\Delta$};
\node [pn] at (0em,12.4em) {$\Gamma$};
%\node (labl) at (3em,-2.5em) {$[\ldr I]$};
\node (ab) at (3em,4.8em) {$\apsnodei$};
\node (a) at (0,9.6em) {$\apsnodei$};
\node (b) at (6em,9.6em) {$\apsnodei$};
\node[tns] (c) at (3em,7.668em) {};
%\node (clab) at (3em,7.668em) {$i$};
\draw (c) -- (ab);
\draw (c) -- (a);
\draw (c) -- (b);
\node (pa) at (0,0) {$\apsnodei$};
\node[par] (pc) at (3em,1.732em) {};
%\node (pclab) at (3em,1.732em) {\textcolor{white}{$i$}};
\draw (pc) -- (ab);
\path[>=latex,->]  (pc) edge (pa);
\draw (b) to [out=50,in=330] (pc);
%%%%
\node at (10em,4.8em) {$\rightarrow$};
%%%
\node at (14em,4.8em) {$\apsnodei$};
\node [pn] at (14em,1.8em) {$\Delta$};
\node [pn] at (14em,7.7em) {$\Gamma$};
\end{tikzpicture}
\end{center}
\caption{Contractions for the non-associative Lambek calculus}
\label{fig:contr:nl}
\end{figure}
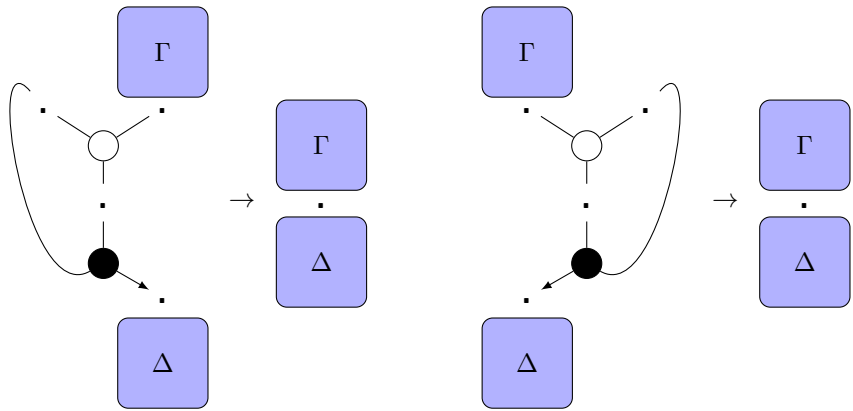

We will present a simplified version of the contraction condition from \citeasnoun{mp}, which is nonetheless powerful enough for the rest of this paper. The simplest contractions are those for the non-associative Lambek calculus NL shown in Figure~\ref{fig:contr:nl}.

The contractions have the same general pattern. They take a pair of links --- one corresponding to an elimination (tensor) rule and one corresponding to an introduction (par) rule --- connected to each other at two vertices. The vertex of the par link with the arrow is not attached to the tensor link, but the other two vertices are. Moreover, the connections respect up/down and left/right. 

The contraction for $\backslash I$, shown on the left of Figure~\ref{fig:contr:nl}, amounts to withdrawing the hypothesis which is the immediate left daughter of the root node, whereas the contraction for $/ I$, shown on the right, withdraws the immediate right daughter.

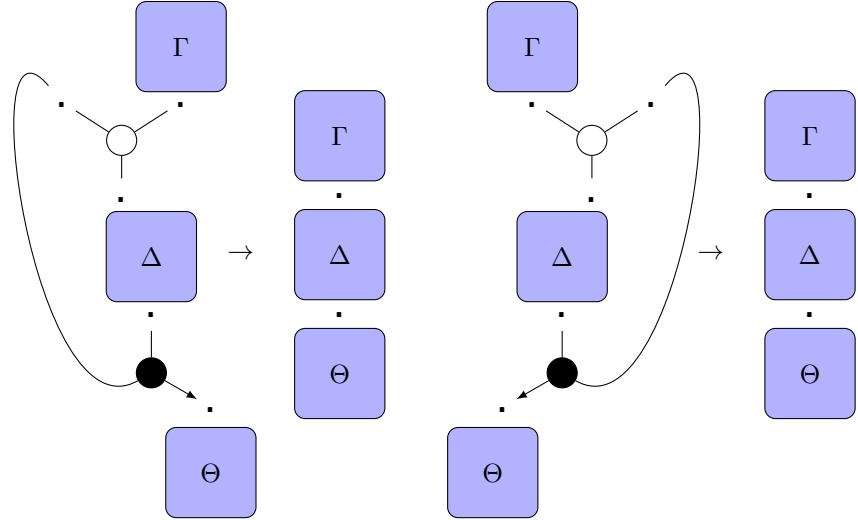
\begin{figure}
\begin{center}
\hspace{-2em}
\begin{tikzpicture}[scale=0.75]
\node [pn] at (7.5em,-3.3em) {$\Theta$};
\node [pn] at (6em,18.2em) {$\Gamma$};
\node [pn] at (4.5em,7.6em) {$\Delta$};
%\node [pn] at (3em,7.6em) {$\Delta$};
\node (ab2) at (3em,10.6em) {$\apsnodei$};
\node (ab) at (4.5em,4.8em) {$\apsnodei$};
\node (a) at (6em,15.4em) {$\apsnodei$};
\node (b) at (0em,15.4em) {$\apsnodei$};
\node[tns] (c) at (3em,13.468em) {};
%\node (clab) at (3em,7.668em) {$i$};
\draw (c) -- (ab2);
\draw (c) -- (a);
\draw (c) -- (b);
\node (pa) at (7.5em,0em) {$\apsnodei$};
\node[par] (pc) at (4.5em,1.732em) {};
%\node (pclab) at (3em,1.732em) {\textcolor{white}{$i$}};
\draw (pc) -- (ab);
\path[>=latex,->]  (pc) edge (pa);
\draw (b) to [out=130,in=210] (pc);
%\node (labl) at (3em,-2.5em) {$[\ldr I]$};
%%%%
\node at (9em,7.7em) {$\rightarrow$};
%%%
\node at (14em,4.8em) {$\apsnodei$};
\node at (14em,10.8em) {$\apsnodei$};
\node [pn] at (14em,1.7em) {$\Theta$};
\node [pn] at (14em,7.7em) {$\Delta$};
\node [pn] at (14em,13.7em) {$\Gamma$};
\end{tikzpicture}
\qquad
\begin{tikzpicture}[scale=0.75]
\node [pn] at (1em,-3.3em) {$\Theta$};
\node [pn] at (3em,18.2em) {$\Gamma$};
\node [pn] at (4.5em,7.6em) {$\Delta$};
%\node [pn] at (3em,7.6em) {$\Delta$};
\node (ab2) at (6em,10.6em) {$\apsnodei$};
\node (ab) at (4.5em,4.8em) {$\apsnodei$};
\node (a) at (9em,15.4em) {$\apsnodei$};
\node (b) at (3em,15.4em) {$\apsnodei$};
\node[tns] (c) at (6em,13.468em) {};
%\node (clab) at (3em,7.668em) {$i$};
\draw (c) -- (ab2);
\draw (c) -- (a);
\draw (c) -- (b);
\node (pa) at (1.5em,0em) {$\apsnodei$};
\node[par] (pc) at (4.5em,1.732em) {};
%\node (pclab) at (3em,1.732em) {\textcolor{white}{$i$}};
\draw (pc) -- (ab);
\path[>=latex,->]  (pc) edge (pa);
%\draw (b) to [out=130,in=210] (pc);
\draw (a) to [out=50,in=330] (pc);
%\node (labl) at (3em,-2.5em) {$[\ldr I]$};
%%%%
\node at (12em,7.7em) {$\rightarrow$};
%%%
\node at (17em,4.8em) {$\apsnodei$};
\node at (17em,10.8em) {$\apsnodei$};
\node [pn] at (17em,1.7em) {$\Theta$};
\node [pn] at (17em,7.7em) {$\Delta$};
\node [pn] at (17em,13.7em) {$\Gamma$};
\end{tikzpicture}
\end{center}
\caption{Contractions for the associative Lambek calculus.}
\label{fig:contr:l}
\end{figure}

Looking back at the abstract proof structure for our example in Figure~\ref{fig:aps:linked}, we see that it is not of the appropriate form to apply the $/ I$ contraction for NL. When we look at the par link, it is connected top-bottom to a tensor link, but the curved line is connected the right branch of a different tensor link. We can therefore not apply the $/ I$ contraction. In a non-associative context, this is the correct behaviour. 

Moving to the associative Lambek calculus, we change the contractions to the ones shown in Figure~\ref{fig:contr:l}. Where for the NL contractions, the top node of the par link was connected to the bottom node of the tensor link, for the L contractions a structure $\Delta$ has been inserted between the two. The contractions now have the conditions that for $\backslash I$, shown on the left of the figure, the path through $\Delta$ only passes through tensor links and moves left each time. For $/ I$, shown on the right of the figure, the path through $\Delta$ only passes through tensor links and moves right each time. 

This means that the $\backslash I$ contraction withdraws the leftmost hypothesis, whereas the $/ I$ contraction withdraws the rightmost hypothesis, which is the desired behaviour for the Lambek calculus introduction rule.

We can relax the contractions even further, keeping only the constraint that the path through $\Delta$ doesn't pass any par links. This produces the left branch and right branch extraction package of \citeasnoun{MMCandR} and \citeasnoun{oehrle11multi}.

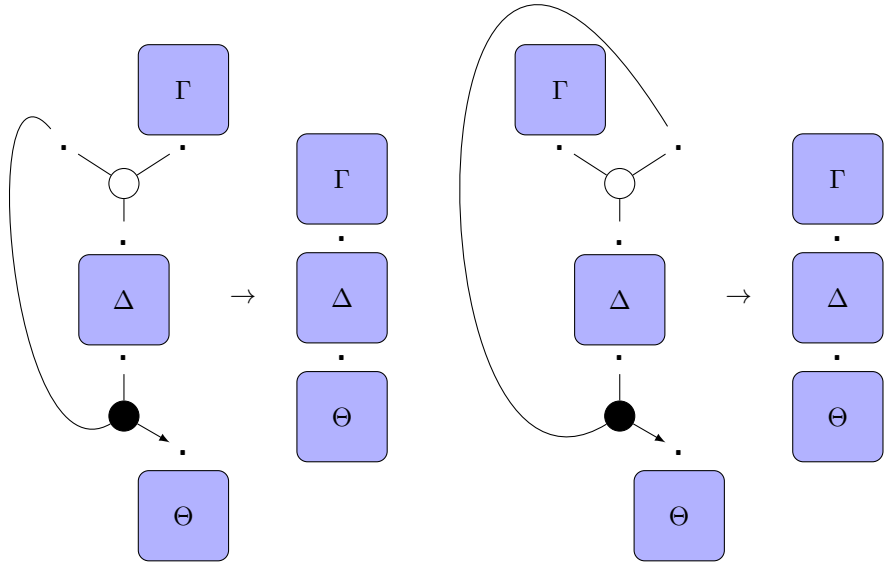
\begin{figure}
\begin{center}
\hspace{-2em}
\begin{tikzpicture}[scale=0.75]
\node [pn] at (6em,-3.3em) {$\Theta$};
\node [pn] at (6em,18.2em) {$\Gamma$};
\node [pn] at (3em,7.6em) {$\Delta$};
\node (ab2) at (3em,10.6em) {$\apsnodei$};
\node (ab) at (3em,4.8em) {$\apsnodei$};
\node (a) at (6em,15.4em) {$\apsnodei$};
\node (b) at (0em,15.4em) {$\apsnodei$};
\node[tns] (c) at (3em,13.468em) {};
%\node (clab) at (3em,7.668em) {$i$};
\draw (c) -- (ab2);
\draw (c) -- (a);
\draw (c) -- (b);
\node (pa) at (6em,0em) {$\apsnodei$};
\node[par] (pc) at (3em,1.732em) {};
%\node (pclab) at (3em,1.732em) {\textcolor{white}{$i$}};
\draw (pc) -- (ab);
\path[>=latex,->]  (pc) edge (pa);
\draw (b) to [out=130,in=210] (pc);
%\node (labl) at (3em,-2.5em) {$[\ldr I]$};
%%%%
\node at (9em,7.7em) {$\rightarrow$};
%%%
\node at (14em,4.8em) {$\apsnodei$};
\node at (14em,10.8em) {$\apsnodei$};
\node [pn] at (14em,1.7em) {$\Theta$};
\node [pn] at (14em,7.7em) {$\Delta$};
\node [pn] at (14em,13.7em) {$\Gamma$};
\end{tikzpicture}
\!
\begin{tikzpicture}[scale=0.75]
\node [pn] at (6em,-3.3em) {$\Theta$};
\node [pn] at (0em,18.2em) {$\Gamma$};
%\node [pn] at (4.5em,7.6em) {$\Delta$};
\node [pn] at (3em,7.6em) {$\Delta$};
\node (ab2) at (3em,10.6em) {$\apsnodei$};
\node (ab) at (3em,4.8em) {$\apsnodei$};
\node (a) at (6em,15.4em) {$\apsnodei$};
\node (b) at (0em,15.4em) {$\apsnodei$};
\node[tns] (c) at (3em,13.468em) {};
%\node (clab) at (3em,7.668em) {$i$};
\draw (c) -- (ab2);
\draw (c) -- (a);
\draw (c) -- (b);
\node (pa) at (6em,0em) {$\apsnodei$};
\node[par] (pc) at (3em,1.732em) {};
%\node (pclab) at (3em,1.732em) {\textcolor{white}{$i$}};
\draw (pc) -- (ab);
\path[>=latex,->]  (pc) edge (pa);
\draw (a) ..controls(-8em,38em)and(-8em,-5em).. (pc);
%\node (labl) at (3em,-2.5em) {$[\ldr I]$};
%%%%
\node at (9em,7.7em) {$\rightarrow$};
%%%
\node at (14em,4.8em) {$\apsnodei$};
\node at (14em,10.8em) {$\apsnodei$};
\node [pn] at (14em,1.7em) {$\Theta$};
\node [pn] at (14em,7.7em) {$\Delta$};
\node [pn] at (14em,13.7em) {$\Gamma$};
\end{tikzpicture}
\end{center}
\caption{Contractions for the associative, commutative Lambek calculus LP}
\label{fig:contr:lp}
\end{figure}

Finally, moving to a fully commutative system makes to the implications interderivable. In the commutative context, we keep only the tensor link for $/$ and write it $\multimap$ as its linear logic counterpart. For the par link, we now treat the conclusions as unordered, which turns the two par links into different ways of writing the same link. Figure~\ref{fig:contr:lp} shows the contractions for $\multimap$. The difference with Figure~\ref{fig:contr:l} is that we now have two contractions for a single connective, depending on whether the hypothesis is on a left or right branch.  

If we denote a move left from the current node (and through a tensor link) by $l$ and right by $r$, then starting from the root node, the path constraints for the contractions look as follows for the logics discussed. 

\medskip
\begin{tabular}{cc}
Left branch $\backslash$ & Right branch $/$ \\ 
\begin{tabular}{l|l}
path & logic \\ \hline
$l$ & NL \\
$l^+$ & L \\
$(lr)^* l$ & left branch extraction \\
$(lr)^+$ & LP	
\end{tabular}
&
\begin{tabular}{l|l}
path & logic \\ \hline
$r$ & NL \\
$r^+$ & L \\
$(lr)^* r$ & right branch extraction \\
$(lr)^+$ & LP	
\end{tabular}
\end{tabular}
\medskip

For NL, we must arrive at the node corresponding to the withdrawn hypothesis after one step (left for $\backslash$, right for $/$). For L, we take at least one step, but all steps must be in the same direction. For the left/right branch extraction package, we only verify the last step is left or right.  For LP, we take at least one step, but there are no constraints on the direction. 

\begin{figure}
\begin{center}	
\scalebox{0.75}{
\begin{tikzpicture}
\node (auxl) at (10em,11.0em) {Aux\proofspace};
\node (aux) at (10em,10em) {$\apsnodei$};
\node (na) at (14em,10em) {$\apsnodei$};
\node [tns] (t) at (12em,8.2em) {};
\node (sspp) at (12em,6em) {$\apsnodei$};
\node (auxs) at (10em,10em) {\blb};
\draw (t)--(auxs);
\draw (t)--(sspp);
\draw (t)--(na);
\node (spp) at (16em,6em) {$\apsnodei$};
\node (auxs) at (14em,2em) {$\apsnodei$};
\node [tns] (tt) at (14em,4.0em) {};
\draw (tt) -- (auxs);
\draw (tt) -- (sspp);
\draw (tt) -- (spp);
\node [par] (par) at (18em,7.8em) {};
\node at (18em,7.8em) {\textcolor{white}{\small 2}};
%\node (pp) at (20em,6em) {$\atppr{}$};
%\node (ppd) at (20em,6em) {\phantom{x}};
\node (tops) at (18em,10em) {$\apsnodei$};
%\draw (par)--(ppd);
\draw (par)--(tops);
\node (sppd) at (16em,6em) {\phantom{x}};
\path[>=latex,->]  (par) edge (sppd);
%%%%
%\node (fils) at (21em,10em) {$\atnr$};
\node (filsl) at (14em,11.0em) {fils...\proofspace};
%%%
\node (clamel) at (22em,7.0em) {clame\proofspace};
%\node (hs) at (20em,2em) {$\atsr$};
\node (ss) at (24em,2em) {$\apsnodei$};
\node (ssd) at (24em,2em) {\phantom{x}};
\node (cs) at (19em,-2em) {$\atsr$};
\node [tns] (t) at (19em,0.2em){};
\draw (t) -- (auxs);
\draw (t) -- (ssd);
\draw (t) -- (cs);
\node (clsu) at (26em,6em) {$\apsnodei$};
\node (clsud) at (26em,6em) {$\apsnodei$};
\node (clame) at (22em,6em) {$\apsnodei$};
\node (clamed) at (22em,6em) {\phantom{M}};
\node [tns] (tc) at (24em,4.2em) {};
\node at (19em,0.2em) {\small 1};
\draw (tc) -- (clamed);
\draw (tc) -- (clsud);
\draw (tc) -- (ss);
%%%
%\node (il) at (33em,10em) {$\atnpr$};
\node at (26em,7.0em) {-t-il\proofspace};
%%%
%\node (je) at (36em,10em) {$\atnpr$};
\node at (16em,15.0em) {j'\proofspace};
%%%
\node at (16em,23.0em) {apporte\proofspace};
\node (apporte) at (16em,22em) {$\apsnodei$};
\node (apob) at (20em,22em) {$\apsnodei$};
\node (tvp) at (18em,18em) {$\apsnodei$};
\node (appp) at (22em,18em) {$\apsnodei$};
\node (vp) at (20em,14em) {$\apsnodei$};
\node (apsu) at (16em,14em) {$\apsnodei$};
\node (apc) at (18em,10em) {\phantom{x}};
% = 18,10.  -24,-12
%
\node [tns] (t1) at (18em,20em) {};
\draw (t1) -- (apporte);
\draw (t1) -- (apob);
\draw (t1) -- (tvp);
\node [tns] (t2) at (20em,16.0em) {};
\draw (t2) -- (vp);
\draw (t2) -- (tvp);
\draw (t2) -- (appp);
\node [tns] (t3) at (18em,12em) {};
\draw (t3) -- (apc);
\draw (t3) -- (vp);
\draw (t3) -- (apsu);
\node at (20em,23.0em) {le salut...\proofspace};
\draw (appp) to [out=50,in=330] (par);
%\node (lesalut) at (49em,10em) {$\atnpr$};
\end{tikzpicture}}
\raisebox{7em}{$\;\;\;\;\;\leadsto$}
\scalebox{0.75}{
\begin{tikzpicture}
\node (auxl) at (10em,11.0em) {Aux\proofspace};
\node (aux) at (10em,10em) {$\apsnodei$};
\node (na) at (14em,10em) {$\apsnodei$};
\node [tns] (t) at (12em,8.2em) {};
\node (sspp) at (12em,6em) {$\apsnodei$};
\node (auxs) at (10em,10em) {\blb};
\draw (t)--(auxs);
\draw (t)--(sspp);
\draw (t)--(na);
\node (spp) at (18em,6em) {$\apsnodei$};
\node (auxs) at (15em,2em) {$\apsnodei$};
\node [tns] (tt) at (15em,4.2em) {};
\draw (tt) -- (auxs);
\draw (tt) -- (sspp);
\draw (tt) -- (spp);
%\node [par] (par) at (18em,7.8em) {};
%\node at (18em,7.8em) {\textcolor{white}{\small 2}};
%\node (pp) at (20em,6em) {$\atppr{}$};
%\node (ppd) at (20em,6em) {\phantom{x}};
%\node (tops) at (18em,10em) {$\apsnodei$};
%\draw (par)--(ppd);
%\draw (par)--(tops);
\node (sppd) at (16em,6em) {\phantom{x}};
%\path[>=latex,->]  (par) edge (sppd);
%%%%
%\node (fils) at (21em,10em) {$\atnr$};
\node (filsl) at (14em,11.0em) {fils...\proofspace};
%%%
\node (clamel) at (22em,7.0em) {clame\proofspace};
%\node (hs) at (20em,2em) {$\atsr$};
\node (ss) at (24em,2em) {$\apsnodei$};
\node (ssd) at (24em,2em) {\phantom{x}};
\node (cs) at (19em,-2em) {$\atsr$};
\node [tns] (t) at (19em,0.2em){};
\draw (t) -- (auxs);
\draw (t) -- (ssd);
\draw (t) -- (cs);
\node (clsu) at (26em,6em) {$\apsnodei$};
\node (clsud) at (26em,6em) {$\apsnodei$};
\node (clame) at (22em,6em) {$\apsnodei$};
\node (clamed) at (22em,6em) {\phantom{M}};
\node [tns] (tc) at (24em,4.2em) {};
\node at (19em,0.2em) {\small 1};
\draw (tc) -- (clamed);
\draw (tc) -- (clsud);
\draw (tc) -- (ss);
%%%
%\node (il) at (33em,10em) {$\atnpr$};
\node at (26em,7.0em) {-t-il\proofspace};
%%%
%\node (je) at (36em,10em) {$\atnpr$};
\node at (16em,11.0em) {j'\proofspace};
%%%
\node at (18em,15.0em) {apporte\proofspace};
%\node (apporte) at (16em,22em) {$\apsnodei$};
%\node (apob) at (20em,22em) {$\apsnodei$};
\node (tvp) at (18em,14em) {$\apsnodei$};
\node (appp) at (22em,14em) {$\apsnodei$};
\node (vp) at (20em,10em) {$\apsnodei$};
\node (apsu) at (16em,10em) {$\apsnodei$};
\node (apc) at (18em,6em) {\phantom{x}};
% = 18,10.  -24,-12
%
%\node [tns] (t1) at (18em,20em) {};
%\draw (t1) -- (apporte);
%\draw (t1) -- (apob);%
%\draw (t1) -- (tvp);
\node [tns] (t2) at (20em,12.0em) {};
\draw (t2) -- (vp);
\draw (t2) -- (tvp);
\draw (t2) -- (appp);
\node [tns] (t3) at (18em,8em) {};
\draw (t3) -- (apc);
\draw (t3) -- (vp);
\draw (t3) -- (apsu);
\node at (22em,15.0em) {le salut...\proofspace};
%\draw (appp) to [out=50,in=330] (par);
%\node (lesalut) at (49em,10em) {$\atnpr$};
\end{tikzpicture}}
\end{center}
\caption{Right branch contraction for the abstract proof structure of Figure~\ref{fig:aps:linked}}
\label{fig:rbcontr}
\end{figure}
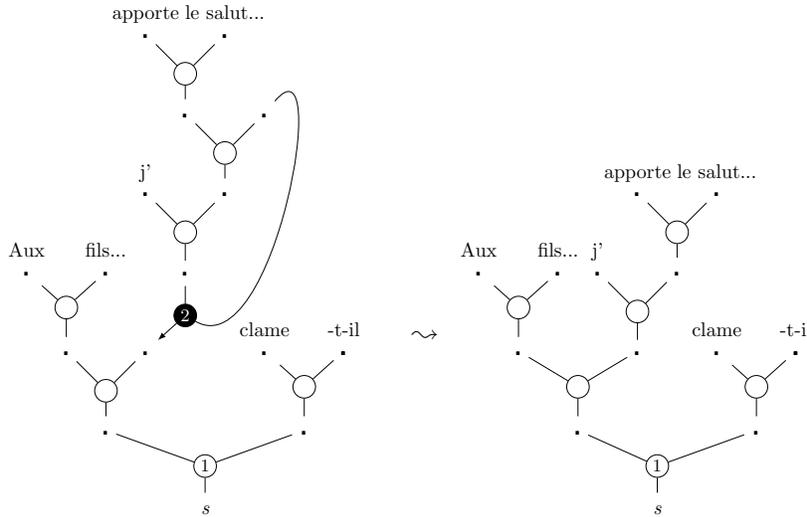

Going back to our previous example, repeated on the right of Figure~\ref{fig:rbcontr}, we can now state more clearly why we have marked the par link with `2'. The marking indicates that a right branch contraction is allowed here\footnote{For this example, a Lambek calculus $/I$ contraction would also work, but for more complicated cases we need the right branch extraction rule.}. 

Verifying that the path from the root to the withdraw hypotheses doesn't pass any other par links (trivially true, since there are no others) and that the withdrawn hypothesis is on a right branch, we apply the contraction to obtain the structure shown on the right of Figure~\ref{fig:rbcontr}.

\begin{figure}
\begin{center}
\scalebox{0.75}{
\begin{tikzpicture}
\node (auxl) at (10em,11.0em) {Aux\proofspace};
\node (aux) at (10em,10em) {$\apsnodei$};
\node (na) at (14em,10em) {$\apsnodei$};
\node [tns] (t) at (12em,8.2em) {};
\node (sspp) at (12em,6em) {$\apsnodei$};
\node (auxs) at (10em,10em) {\blb};
\draw (t)--(auxs);
\draw (t)--(sspp);
\draw (t)--(na);
\node (spp) at (18em,6em) {$\apsnodei$};
\node (auxs) at (15em,2em) {$\apsnodei$};
\node [tns] (tt) at (15em,4.2em) {};
\draw (tt) -- (auxs);
\draw (tt) -- (sspp);
\draw (tt) -- (spp);
%\node [par] (par) at (18em,7.8em) {};
%\node at (18em,7.8em) {\textcolor{white}{\small 2}};
%\node (pp) at (20em,6em) {$\atppr{}$};
%\node (ppd) at (20em,6em) {\phantom{x}};
%\node (tops) at (18em,10em) {$\apsnodei$};
%\draw (par)--(ppd);
%\draw (par)--(tops);
\node (sppd) at (16em,6em) {\phantom{x}};
%\path[>=latex,->]  (par) edge (sppd);
%%%%
%\node (fils) at (21em,10em) {$\atnr$};
\node (filsl) at (14em,11.0em) {fils...\proofspace};
%%%
\node (clamel) at (22em,7.0em) {clame\proofspace};
%\node (hs) at (20em,2em) {$\atsr$};
\node (ss) at (24em,2em) {$\apsnodei$};
\node (ssd) at (24em,2em) {\phantom{x}};
\node (cs) at (19em,-2em) {$\atsr$};
\node [tns] (t) at (19em,0.2em){};
\draw (t) -- (auxs);
\draw (t) -- (ssd);
\draw (t) -- (cs);
\node (clsu) at (26em,6em) {$\apsnodei$};
\node (clsud) at (26em,6em) {$\apsnodei$};
\node (clame) at (22em,6em) {$\apsnodei$};
\node (clamed) at (22em,6em) {\phantom{M}};
\node [tns] (tc) at (24em,4.2em) {};
\node at (19em,0.2em) {\small 1};
\draw (tc) -- (clamed);
\draw (tc) -- (clsud);
\draw (tc) -- (ss);
%%%
%\node (il) at (33em,10em) {$\atnpr$};
\node at (26em,7.0em) {-t-il\proofspace};
%%%
%\node (je) at (36em,10em) {$\atnpr$};
\node at (16em,11.0em) {j'\proofspace};
%%%
\node at (18em,15.0em) {apporte\proofspace};
%\node (apporte) at (16em,22em) {$\apsnodei$};
%\node (apob) at (20em,22em) {$\apsnodei$};
\node (tvp) at (18em,14em) {$\apsnodei$};
\node (appp) at (22em,14em) {$\apsnodei$};
\node (vp) at (20em,10em) {$\apsnodei$};
\node (apsu) at (16em,10em) {$\apsnodei$};
\node (apc) at (18em,6em) {\phantom{x}};
% = 18,10.  -24,-12
%
%\node [tns] (t1) at (18em,20em) {};
%\draw (t1) -- (apporte);
%\draw (t1) -- (apob);%
%\draw (t1) -- (tvp);
\node [tns] (t2) at (20em,12.0em) {};
\draw (t2) -- (vp);
\draw (t2) -- (tvp);
\draw (t2) -- (appp);
\node [tns] (t3) at (18em,8em) {};
\draw (t3) -- (apc);
\draw (t3) -- (vp);
\draw (t3) -- (apsu);
\node at (22em,15.0em) {le salut...\proofspace};
%\draw (appp) to [out=50,in=330] (par);
%\node (lesalut) at (49em,10em) {$\atnpr$};
\end{tikzpicture}}
\raisebox{7em}{$\;\;\;\;\;\leadsto$}%\hspace{-3em}
%%%%%%%%
\scalebox{0.75}{
\begin{tikzpicture}
\node (auxl) at (10em,11.0em) {Aux\proofspace};
\node (aux) at (10em,10em) {$\apsnodei$};
\node (na) at (14em,10em) {$\apsnodei$};
\node [tns] (t) at (12em,8.2em) {};
\node (sspp) at (12em,6em) {$\apsnodei$};
\node (auxs) at (10em,10em) {\blb};
\draw (t)--(auxs);
\draw (t)--(sspp);
\draw (t)--(na);
\node (spp) at (18em,6em) {$\apsnodei$};
\node (auxs) at (15em,2em) {$\apsnodei$};
\node [tns] (tt) at (15em,4.2em) {};
\draw (tt) -- (auxs);
\draw (tt) -- (sspp);
\draw (tt) -- (spp);
%\node [par] (par) at (18em,7.8em) {};
%\node at (18em,7.8em) {\textcolor{white}{\small 2}};
%\node (pp) at (20em,6em) {$\atppr{}$};
%\node (ppd) at (20em,6em) {\phantom{x}};
%\node (tops) at (18em,10em) {$\apsnodei$};
%\draw (par)--(ppd);
%\draw (par)--(tops);
\node (sppd) at (16em,6em) {\phantom{x}};
%\path[>=latex,->]  (par) edge (sppd);
%%%%
%\node (fils) at (21em,10em) {$\atnr$};
\node (filsl) at (13.7em,11.0em) {fils...\proofspace};
%%%
\node (clamel) at (16.3em,11.0em) {clame\proofspace};
%\node (hs) at (20em,2em) {$\atsr$};
\node (ss) at (24em,2em) {$\apsnodei$};
\node (ssd) at (24em,2em) {\phantom{x}};
\node (cs) at (19em,-2em) {$\atsr$};
\node (til) at (20em,10em) {$\apsnodei$};
\node [tns] (t) at (19em,0.2em){};
\draw (t) -- (auxs);
\draw (t) -- (ssd);
\draw (t) -- (cs);
\node (clsu) at (26em,6em) {$\apsnodei$};
\node (clsud) at (26em,6em) {$\apsnodei$};
\node (clame) at (22em,6em) {$\apsnodei$};
\node (clamed) at (22em,6em) {\phantom{M}};
\node [tns] (tc) at (24em,4.2em) {};
\node at (15em,4.2em) {\small 1};
\draw (tc) -- (clamed);
\draw (tc) -- (clsud);
\draw (tc) -- (ss);
%%%
%\node (il) at (33em,10em) {$\atnpr$};
\node at (20em,11.0em) {-t-il\proofspace};
%%%
%\node (je) at (36em,10em) {$\atnpr$};
\node at (22em,7.0em) {j'\proofspace};
%%%
\node at (24em,11.0em) {apporte\proofspace};
%\node (apporte) at (16em,22em) {$\apsnodei$};
%\node (apob) at (20em,22em) {$\apsnodei$};
\node (tvp) at (24em,10em) {$\apsnodei$};
\node (appp) at (28em,10em) {$\apsnodei$};
\node (vp) at (26em,6em) {$\apsnodei$};
\node (apsu) at (16em,10em) {$\apsnodei$};
\node (apc) at (18em,6em) {\phantom{x}};
% = 18,10.  -24,-12
%
%\node [tns] (t1) at (18em,20em) {};
%\draw (t1) -- (apporte);
%\draw (t1) -- (apob);%
%\draw (t1) -- (tvp);
\node [tns] (t2) at (26em,8.0em) {};
\draw (t2) -- (vp);
\draw (t2) -- (tvp);
\draw (t2) -- (appp);
\node [tns] (t3) at (18em,8em) {};
\draw (t3) -- (apc);
\draw (t3) -- (til);
\draw (t3) -- (apsu);
\node at (28em,11.0em) {le salut...\proofspace};
%\draw (appp) to [out=50,in=330] (par);
%\node (lesalut) at (49em,10em) {$\atnpr$};
\end{tikzpicture}}
\end{center}
\caption{Tree rewrite to insert ``clame-t-il'' at its correct place}
\label{fig:rewrend} 
\end{figure}

While we have contracted the initial proof structure to a tree, the yield of the tree is different from the input sentence. An additional tree rewrite, shown in Figure~\ref{fig:rewrend}, allows us to move structures on the right of branches labeled `1' inside. This produces a tree with the correct yield, as shown on the right of Figure~\ref{fig:rewrend}. This is a fairly standard type of rule in multimodal type-logical grammars \cite{M95}.

\subsection{Proof nets and lambda terms}

Proof nets, like natural deduction proofs, correspond to lambda terms. These are linear lambda terms, because we operate in a fragment of multiplicative linear logic. Since multiplicative linear logic with just the `$\multimap$' connective is our semantic language, we can see the derivational meaning as a homomorphism from Lambek calculus proofs (or proofs in any of the variants of the Lambek calculus) into proofs of multiplicative linear logic.

Concretely, this means we replace the Lambek calculus implications `$/$' and `$\backslash$' (and their variants, like `$\backslash_1$' and `$/_2$') with linear implication `$\multimap$'. Meanings are computed at the level of proof nets, that is, proof structures whose abstract proof structures can be contracted using the LP contractions or any restriction of these contractions. However, the actual contractions do not influence the meaning, only the initial proof structure does.

\begin{figure}
\begin{center}
\scalebox{0.75}{
\begin{tikzpicture}
\node (auxl) at (10em,11.0em) {Aux\proofspace};
\node (aux) at (10em,10em) {$\apsnodei$};
\node (na) at (14em,10em) {$\apsnodei$};
\node [tns] (t) at (12em,8.2em) {};
\node at (12em,8.2em) {$/$};
\node (sspp) at (12em,6em) {$\apsnodei$};
\node (auxs) at (10em,10em) {\blb};
\draw (t)--(auxs);
\draw (t)--(sspp);
\draw (t)--(na);
\node (spp) at (16em,6em) {$\apsnodei$};
\node (auxs) at (14em,2em) {$\apsnodei$};
\node [tns] (tt) at (14em,4.0em) {};
\node at (14em,4em) {$/$};
\draw (tt) -- (auxs);
\draw (tt) -- (sspp);
\draw (tt) -- (spp);
\node [par] (par) at (18em,7.8em) {};
\node at (18em,7.8em) {\textcolor{white}{\small /}};
%\node (pp) at (20em,6em) {$\atppr{}$};
%\node (ppd) at (20em,6em) {\phantom{x}};
\node (tops) at (18em,10em) {$\apsnodei$};
%\draw (par)--(ppd);
\draw (par)--(tops);
\node (sppd) at (16em,6em) {\phantom{x}};
\path[>=latex,->]  (par) edge (sppd);
%%%%
%\node (fils) at (21em,10em) {$\atnr$};
\node (filsl) at (14em,11.0em) {fils...\proofspace};
%%%
\node (clamel) at (22em,7.0em) {clame\proofspace};
%\node (hs) at (20em,2em) {$\atsr$};
\node (ss) at (24em,2em) {$\apsnodei$};
\node (ssd) at (24em,2em) {\phantom{x}};
\node (cs) at (19em,-2em) {$\atsr$};
\node [tns] (t) at (19em,0.2em){};
\node at (19em,0.2em) {$\backslash$};
\draw (t) -- (auxs);
\draw (t) -- (ssd);
\draw (t) -- (cs);
\node (clsu) at (26em,6em) {$\apsnodei$};
\node (clsud) at (26em,6em) {$\apsnodei$};
\node (clame) at (22em,6em) {$\apsnodei$};
\node (clamed) at (22em,6em) {\phantom{M}};
\node [tns] (tc) at (24em,4.2em) {};
\node at (24em,4.2em) {$/$};
\draw (tc) -- (clamed);
\draw (tc) -- (clsud);
\draw (tc) -- (ss);
%%%
%\node (il) at (33em,10em) {$\atnpr$};
\node at (26em,7.0em) {-t-il\proofspace};
%%%
%\node (je) at (36em,10em) {$\atnpr$};
\node at (16em,15.0em) {j'\proofspace};
%%%
\node at (16em,23.0em) {apporte\proofspace};
\node (apporte) at (16em,22em) {$\apsnodei$};
\node (apob) at (20em,22em) {$\apsnodei$};
\node (tvp) at (18em,18em) {$\apsnodei$};
\node (appp) at (22em,18em) {$\apsnodei$};
\node (vp) at (20em,14em) {$\apsnodei$};
\node (apsu) at (16em,14em) {$\apsnodei$};
\node (apc) at (18em,10em) {\phantom{x}};
% = 18,10.  -24,-12
%
\node [tns] (t1) at (18em,20em) {};
\node at (18em,20em) {$/$};
\draw (t1) -- (apporte);
\draw (t1) -- (apob);
\draw (t1) -- (tvp);
\node [tns] (t2) at (20em,16.0em) {};
\node at (20em,16em) {$/$};
\draw (t2) -- (vp);
\draw (t2) -- (tvp);
\draw (t2) -- (appp);
\node [tns] (t3) at (18em,12em) {};
\node at (18em,12em) {$\backslash$};
\draw (t3) -- (apc);
\draw (t3) -- (vp);
\draw (t3) -- (apsu);
\node at (20em,23.0em) {le salut...\proofspace};
\draw (appp) to [out=50,in=330] (par);
%\node (lesalut) at (49em,10em) {$\atnpr$};
\end{tikzpicture}}
\raisebox{7em}{$\;\;\;\;\;\leadsto$}
\scalebox{0.75}{
\begin{tikzpicture}
\node (auxl) at (10em,11.0em) {Aux\proofspace};
\node (aux) at (10em,10em) {$\apsnodei$};
\node (na) at (14em,10em) {$\apsnodei$};
\node [tns] (t) at (12em,8.2em) {};
\node at (12em,8.2em) {\appl};
\node (sspp) at (12em,6em) {$\apsnodei$};
\node (auxs) at (10em,10em) {\blb};
\draw (t)--(auxs);
\draw (t)--(sspp);
\draw (t)--(na);
\node (spp) at (16em,6em) {$\apsnodei$};
\node (auxs) at (14em,2em) {$\apsnodei$};
\node [tns] (tt) at (14em,4.0em) {};
\node at (14em,4em) {\appl};
\draw (tt) -- (auxs);
\draw (tt) -- (sspp);
\draw (tt) -- (spp);
\node [par] (par) at (18em,7.8em) {};
\node at (18em,7.8em) {$\textcolor{white}{\small \lambda}$};
%\node (pp) at (20em,6em) {$\atppr{}$};
%\node (ppd) at (20em,6em) {\phantom{x}};
\node (tops) at (18em,10em) {$\apsnodei$};
%\draw (par)--(ppd);
\draw (par)--(tops);
\node (sppd) at (16em,6em) {\phantom{x}};
\path[>=latex,->]  (par) edge (sppd);
%%%%
%\node (fils) at (21em,10em) {$\atnr$};
\node (filsl) at (14em,11.0em) {fils...\proofspace};
%%%
\node (clamel) at (02em,7.0em) {clame\proofspace};
%\node (hs) at (20em,2em) {$\atsr$};
\node (ss) at (04em,2em) {$\apsnodei$};
\node (ssd) at (4em,2em) {\phantom{x}};
\node (cs) at (9em,-2em) {$\apsnodei$};
\node [tns] (t) at (9em,0.2em){};
\node at (9em,0.2em){\appl};
\draw (t) -- (auxs);
\draw (t) -- (ssd);
\draw (t) -- (cs);
\node (clsu) at (06em,6em) {$\apsnodei$};
\node (clsud) at (06em,6em) {$\apsnodei$};
\node (clame) at (02em,6em) {$\apsnodei$};
\node (clamed) at (02em,6em) {\phantom{M}};
\node [tns] (tc) at (04em,4.2em) {};
\node at (4em,4.2em) {\appl};
\draw (tc) -- (clamed);
\draw (tc) -- (clsud);
\draw (tc) -- (ss);
%%%
%\node (il) at (33em,10em) {$\atnpr$};
\node at (06em,7.0em) {-t-il\proofspace};
%%%
%\node (je) at (36em,10em) {$\atnpr$};
\node at (19.8em,15.0em) {j'\proofspace};
%%%
\node at (12em,23.0em) {apporte\proofspace};
\node (apporte) at (12em,22em) {$\apsnodei$};
\node (apob) at (16em,22em) {$\apsnodei$};
\node (tvp) at (14em,18em) {$\apsnodei$};
\node (appp) at (18em,18em) {$\apsnodei$};
\node (vp) at (16em,14em) {$\apsnodei$};
\node (apsu) at (20em,14em) {$\apsnodei$};
\node (apc) at (18em,10em) {\phantom{x}};
% = 18,10.  -24,-12
%
\node [tns] (t1) at (14em,20.2em) {};
\node at (14em,20.2em) {\appl};
\draw (t1) -- (apporte);
\draw (t1) -- (apob);
\draw (t1) -- (tvp);
\node [tns] (t2) at (16em,16.2em) {};
\node at (16em,16.2em) {\appl};
\draw (t2) -- (vp);
\draw (t2) -- (tvp);
\draw (t2) -- (appp);
\node [tns] (t3) at (18em,12em) {};
\node at (18em,12em) {\appl};
\draw (t3) -- (apc);
\draw (t3) -- (vp);
\draw (t3) -- (apsu);
\node at (16em,23.0em) {le salut...\proofspace};
\draw (appp) to [out=50,in=330] (par);
%\node (lesalut) at (49em,10em) {$\atnpr$};
\end{tikzpicture}}
\end{center}
\caption{Semantic version of the proof net of Figure~\ref{fig:aps:linked}}
\label{fig:sem}
\end{figure}
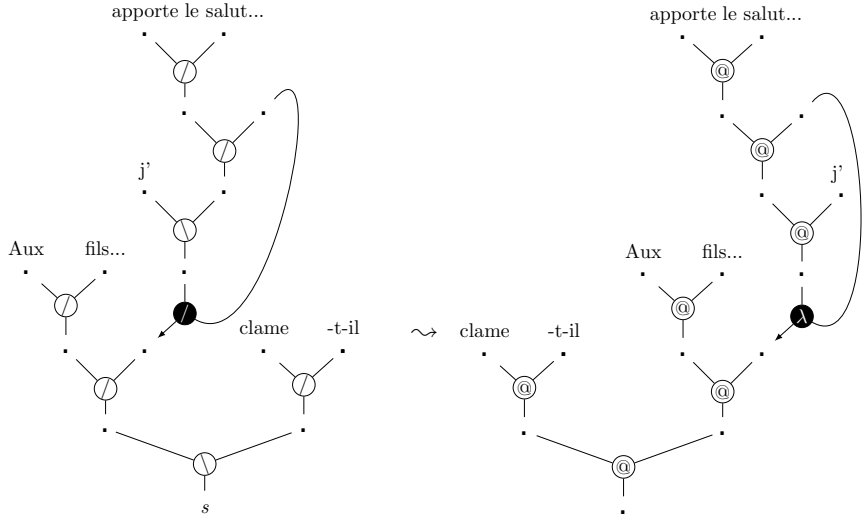

Going back to the abstract proof structure of Figure~\ref{fig:aps:linked}, before any contractions, we take this previous figure but annotate each link with the direction of the implication. This gives the figure shown on the left of Figure~\ref{fig:sem}. Even though the directional information is superfluous, it makes the transformation to a linear logic proof more easy to explain. To transform this abstract proof structure into a structure representing the meaning, we simply swap the order of the premisses of the tensor links marked $\backslash$. Since, by convention, we treat the conclusions of the LP par link as unordered, nothing has to be done for the par link. This produces the structure shown on the right of Figure~\ref{fig:sem}.

We have changed the labels on the tensor links to $@$ for application ($\multimap E$), and the par links to $\lambda$ for abstraction ($\multimap I$), indicating the interpretation in terms of meaning composition. 

\begin{figure}
\begin{center}
\scalebox{0.75}{
\begin{tikzpicture}
\node (auxl) at (10em,11.0em) {Aux\proofspace};
\node (aux) at (10em,10em) {$u$};
\node (na) at (14em,10em) {$f$};
\node [tns] (t) at (12em,8.2em) {};
\node at (12em,8.2em) {\appl};
\node (sspp) at (12em,6em) {$u\,f$};
\node (auxs) at (10em,10em) {\blb};
\draw (t)--(auxs);
\draw (t)--(sspp);
\draw (t)--(na);
\node (spp) at (16em,6em) {$\quad\lambda x. (((a\,s)\,x)\,j)$};
\node (auxs) at (14em,2em) {$(u\,f)\,\lambda x. (((a\,s)\,x)\,j)$};
\node [tns] (tt) at (14em,4.0em) {};
\node at (14em,4em) {\appl};
\draw (tt) -- (auxs);
\draw (tt) -- (sspp);
\draw (tt) -- (spp);
\node [par] (par) at (18em,7.8em) {};
\node at (18em,7.8em) {$\textcolor{white}{\small \lambda}$};
%\node (pp) at (20em,6em) {$\atppr{}$};
%\node (ppd) at (20em,6em) {\phantom{x}};
\node (tops) at (18em,10em) {$((a\,s)\,x)\,j$};
%\draw (par)--(ppd);
\draw (par)--(tops);
\node (sppd) at (16em,6em) {\phantom{x}};
\path[>=latex,->]  (par) edge (sppd);
%%%%
%\node (fils) at (21em,10em) {$\atnr$};
\node (filsl) at (14em,11.0em) {fils...\proofspace};
%%%
\node (clamel) at (02em,7.0em) {clame\proofspace};
%\node (hs) at (20em,2em) {$\atsr$};
\node (ss) at (04em,2em) {$c\,i$};
\node (ssd) at (4em,2em) {\phantom{x}};
\node (cs) at (9em,-2em) {$(c\,i)\, ((u\,f)\,\lambda x. (((a\,s)\,x)\,j) )$};
\node [tns] (t) at (9em,0.2em){};
\node at (9em,0.2em){\appl};
\draw (t) -- (auxs);
\draw (t) -- (ssd);
\draw (t) -- (cs);
\node (clsu) at (06em,6em) {$i$};
\node (clsud) at (06em,6em) {$i$};
\node (clame) at (02em,6em) {$c$};
\node (clamed) at (02em,6em) {\phantom{M}};
\node [tns] (tc) at (04em,4.2em) {};
\node at (4em,4.2em) {\appl};
\draw (tc) -- (clamed);
\draw (tc) -- (clsud);
\draw (tc) -- (ss);
%%%
%\node (il) at (33em,10em) {$\atnpr$};
\node at (06em,7.0em) {-t-il\proofspace};
%%%
%\node (je) at (36em,10em) {$\atnpr$};
\node at (19.8em,15.0em) {j'\proofspace};
%%%
\node at (12em,23.0em) {apporte\proofspace};
\node (apporte) at (12em,22em) {$a$};
\node (apob) at (16em,22em) {$s$};
\node (tvp) at (14em,18em) {$a\, s$};
\node (appp) at (18em,18em) {$x$};
\node (vp) at (16em,14em) {$(a\,s)\,x$};
\node (apsu) at (20em,14em) {$j$};
\node (apc) at (18em,10em) {\phantom{x}};
% = 18,10.  -24,-12
%
\node [tns] (t1) at (14em,20.2em) {};
\node at (14em,20.2em) {\appl};
\draw (t1) -- (apporte);
\draw (t1) -- (apob);
\draw (t1) -- (tvp);
\node [tns] (t2) at (16em,16.2em) {};
\node at (16em,16.2em) {\appl};
\draw (t2) -- (vp);
\draw (t2) -- (tvp);
\draw (t2) -- (appp);
\node [tns] (t3) at (18em,12em) {};
\node at (18em,12em) {\appl};
\draw (t3) -- (apc);
\draw (t3) -- (vp);
\draw (t3) -- (apsu);
\node at (16em,23.0em) {le salut...\proofspace};
\draw (appp) to [out=50,in=330] (par);
%\node (lesalut) at (49em,10em) {$\atnpr$};
\end{tikzpicture}}
\end{center}
\caption{Figure~\ref{fig:sem} with terms at the vertices}
\label{fig:semterm}
\end{figure}
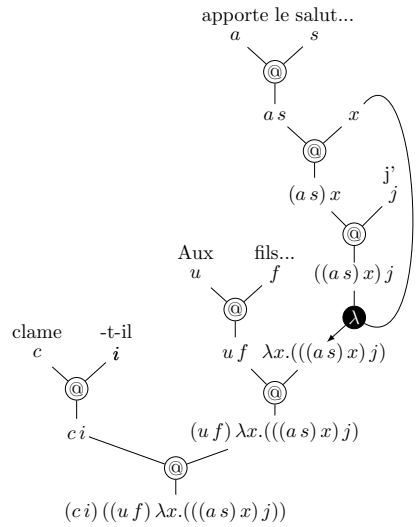

Using variables $u$ for ``Aux'', $f$ for ``fils'', $c$, for ``clame'', etc., we can then compute the meaning corresponding to this proof as shown in Figure~\ref{fig:semterm}. The computed term 
\[
(c\,i)\, ((u\,f)\,\lambda x. (((a\,s)\,x)\,j) )
\]
\noindent is just a flat representation of the information in the LP proof net. Given a sequence of variables $u, f, c, i, j, a, s$ (we will generally just use $x_1,\ldots, x_n$ for an $n$-word sentence) we can --- by the Curry-Howard isomorphism between LP proofs and linear lambda terms --- uniquely reconstruct the proof net from the lambda term and vice versa\footnote{On the condition that the term has no closed subterms.}.

\section{Neural proof nets}
\label{sec:npn}

Proof enumeration for type-logical grammars works well in a research context, with relatively small lexicons and sentences. But how can we `scale' type-logical grammar parsing to use these grammars for natural language understanding tasks? The standard approach has two main bottlenecks. First, lexical ambiguity (that is, words can be assigned many different formulas) becomes a major problem. Second, the matching step is formally equivalent to finding a correct permutation (under some constraints). For both of these bottlenecks, we are no longer interesting in enumerating the different possibilities, but in finding the correct solution, where we mean ``correct'' in the sense that it corresponds to the intended interpretation of the sentence (more prosaically, this means it should be the same as the proof of the sentence we find in a given corpus).   

Different machine learning approaches have been successfully applied to the first task under the label `supertagging' \cite{bj11super}. As in many other research areas, deep learning has greatly improved the accuracy on this task. For example, for the French TLGbank \cite{moot15tlgbank}, a maximum-entropy supertagger obtains 88.8\% of the assigned formulas correct whereas the current state-of-the-art using neural networks is at 95.9\% \cite{kogkalidis2022geometry}.

The second task, linking the atomic formulas, has only recently been implemented using a neural network \cite{kogkalidis2020neural}. This provides a complete neural network solution for parsing arbitrary text using type-logical grammars: neural proof nets. Neural proof nets have been successfully applied to the Dutch \AE{}thel dataset \cite{kogkalidis2020aethel}, and the French TLG\-bank \cite{deepgrail23}.

%\begin{enumerate}
%\item predict, for each word in the sentence, the correct formula,
%\item predict the correct perfect matching between the atomic sub-formulas of the formulas of the previous step.  
%\end{enumerate}

\section{Different perspectives}
\label{sec:alt}

While we have seen important improvements in supertagging, it remains an important bottleneck for neural proof nets: a single error in the formula assignment can cause a proof to fail and not produce a lambda term at all. In addition, compared to standard parser evaluation metrics, it is hard to measure partial successes in neural proof nets: we can calculate the percentage of words assigned the correct formula, and, given the correct formulas, we can calculate the percentage of correct links, but it is not obvious how to calculate a score combining these two percentages into a single useful metric other than the extremely severe `percentage of sentences without any errors'.

%For similar reasons, it is not obvious how to produce the $k$-best proof nets using neural proof nets. We can produce the $k$-best sequences of formulas, and combine it with a strategy proposing multiple linkings, but it is unclear how to combine the two tasks to produce a single, ranked metric. We can interpret the supertagger results as probabilities over formulas, and similarly interpret the different axiom links as all having a certain probability 
% Ideally, the linking step should be able to downvote the best supertagger choice

I therefore propose a different perspective on neural proof nets: instead of dividing the task into a supertagging and a linking task, we divide the task in a graph generation and graph labelling task. The graph generation task is set up in such a way it generates a proof net, step by step, ensuring at each step that the structure is valid. The graph labelling task then assigns labels to the graph vertices, where the labels are the logical connectives and atomic formulas. 

The advantage of this setup is that it ensures by construction we obtain a proof net (and therefore a lambda term) for our input sentence, and that errors in the labelling phrase will have a relatively minor impact on downstream tasks. We can also see the inductive proof net construction steps as parser actions and explore multiple solutions (for example, using beam search). This makes it easy for a model to predict the $k$-best proof nets for a sentence.

We can set up this alternative vision of neural proof nets in such a way that each proof net is generated by a unique tree of parser actions. This also provides a way to solve the proof comparison problem for neural proof nets. If each proof net can be uniquely described by the actions used to create it, then we can compare two sets of actions and apply standard statistical measures (such as f-scores) for comparing these two sets.

\subsection{Graph generation}
\label{sec:gg}

The graph generation task is the most important one, given that it constructs the linear logic proof representing the meaning of the phrase, which is generally what interests us for applications in natural language processing. Graph generation is also the more complicated task given that it doesn't fit as neatly into a well-understood graph neural network architecture.

While we can set up the graph generation component as a form of backward chaining proof search (as we will see in Appendix~\ref{app:backward}), it is most convenient to set it up as a form of forward chaining proof search. For forward chaining, the starting point is a set of isolated vertices, one for each word in the sentence. There is, of course, an implicit linear order on these vertices, which corresponds to the linear order of the corresponding words in the sentence.

The invariant we want to maintain is that each component of the graph corresponds to a proof and a lambda term. Among other things, this means that each component has a unique root node. In our initial structure, we have an axiom $A_i \vdash A_i$ and a free variable $x_i$ for each of the initial vertices.

\begin{figure}
\begin{center}
\begin{tikzpicture}[scale=0.75]
\node [pn] at (2em,18.4em) {$\Gamma$};
\node [pn] at (10em,18.4em) {$\Delta$};
\node (a) at (2em,15.4em) {$\apsnodei$};
\node (b) at (10em,15.4em) {$\apsnodei$};
%%%
\node [pn] at (20em,18.4em) {$\Gamma$};
%\node [pn] at (4.5em,7.6em) {$\Delta$};
\node [pn] at (26em,18.4em) {$\Delta$};
%\node (a) at (4em,1em) {$\apsnodei$};
%\node (b) at (12em,1em) {$\apsnodei$};
%
\node (ab2) at (23em,10.6em) {$\apsnodei$};
\node (a) at (26em,15.4em) {$\apsnodei$};
\node (b) at (20em,15.4em) {$\apsnodei$};
\node[tns] (c) at (23em,13.468em) {};
%\node (clab) at (3em,7.668em) {$i$};
\draw (c) -- (ab2);
\draw (c) -- (a);
\draw (c) -- (b);
\node at (15em,18.4em) {$\rightarrow$};
\end{tikzpicture}
\end{center}
\caption{Graph composition}
\label{fig:grcomp}
\end{figure}
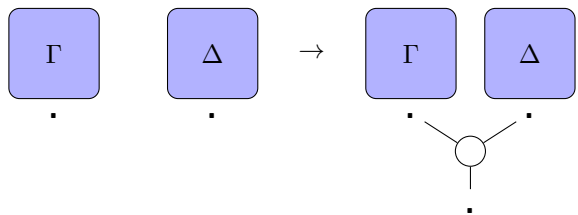

The basic operation is graph composition, shown in Figure~\ref{fig:grcomp}. It takes the root node of two disjoint structures, and combines them into a single structure. Logically, this operation corresponds to the $\multimap E$ rule without explicit formula labels, and to application on the level of terms. Since $\Gamma$ and $\Delta$ are disjoint, this is a valid rule application and one producing a linear lambda term.

The introduction rule and abstraction operation are a bit trickier: we abstract over a variable, but this cannot be one of the variables corresponding to one of the words in the sentence, since these need to be free in the final term we compute. The solution is therefore to introduce a new variable as the sister of a node in the given structure, then introduce an abstraction at an ancestor node of this new variable. There are then two cases, depending on whether the newly introduced (and abstracted) variable is a functor or an argument\footnote{The attentive reader will note that this excludes terms of the form $\lambda x.x$. If we need these terms, we can add a specific graph rewrite for this case. However, type-logical grammar proofs are generally restrict to proofs without empty antecedents, and for proofs without empty antecedents the two given cases suffice.}.

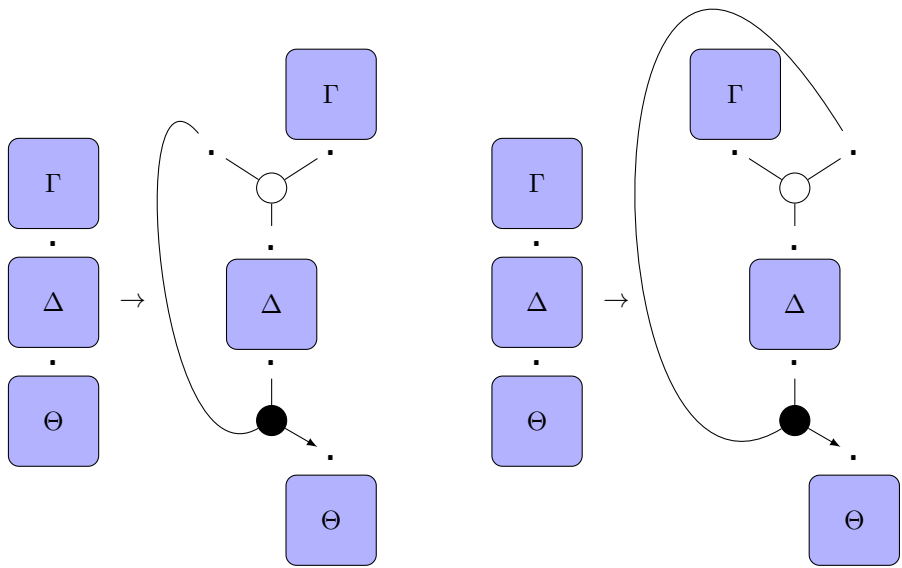
\begin{figure}
\begin{tikzpicture}[scale=0.75]
\node [pn] at (6em,-3.3em) {$\Theta$};
\node [pn] at (6em,18.2em) {$\Gamma$};
%\node [pn] at (4.5em,7.6em) {$\Delta$};
\node [pn] at (3em,7.6em) {$\Delta$};
\node (ab2) at (3em,10.6em) {$\apsnodei$};
\node (ab) at (3em,4.8em) {$\apsnodei$};
\node (a) at (6em,15.4em) {$\apsnodei$};
\node (b) at (0em,15.4em) {$\apsnodei$};
\node[tns] (c) at (3em,13.468em) {};
%\node (clab) at (3em,7.668em) {$i$};
\draw (c) -- (ab2);
\draw (c) -- (a);
\draw (c) -- (b);
\node (pa) at (6em,0em) {$\apsnodei$};
\node[par] (pc) at (3em,1.732em) {};
%\node (pclab) at (3em,1.732em) {\textcolor{white}{$i$}};
\draw (pc) -- (ab);
\path[>=latex,->]  (pc) edge (pa);
%\draw (a) ..controls(-8em,38em)and(-8em,-5em).. (pc);
\draw (b) to [out=130,in=210] (pc);
%\node (labl) at (3em,-2.5em) {$[\ldr I]$};
%%%%
\node at (-4em,7.7em) {$\rightarrow$};
%%%
\node at (-8em,4.8em) {$\apsnodei$};
\node at (-8em,10.8em) {$\apsnodei$};
\node [pn] at (-8em,1.7em) {$\Theta$};
\node [pn] at (-8em,7.7em) {$\Delta$};
\node [pn] at (-8em,13.7em) {$\Gamma$};
\end{tikzpicture}
\qquad\qquad
\begin{tikzpicture}[scale=0.75]
\node [pn] at (6em,-3.3em) {$\Theta$};
\node [pn] at (0em,18.2em) {$\Gamma$};
%\node [pn] at (4.5em,7.6em) {$\Delta$};
\node [pn] at (3em,7.6em) {$\Delta$};
\node (ab2) at (3em,10.6em) {$\apsnodei$};
\node (ab) at (3em,4.8em) {$\apsnodei$};
\node (a) at (6em,15.4em) {$\apsnodei$};
\node (b) at (0em,15.4em) {$\apsnodei$};
\node[tns] (c) at (3em,13.468em) {};
%\node (clab) at (3em,7.668em) {$i$};
\draw (c) -- (ab2);
\draw (c) -- (a);
\draw (c) -- (b);
\node (pa) at (6em,0em) {$\apsnodei$};
\node[par] (pc) at (3em,1.732em) {};
%\node (pclab) at (3em,1.732em) {\textcolor{white}{$i$}};
\draw (pc) -- (ab);
\path[>=latex,->]  (pc) edge (pa);
\draw (a) ..controls(-8em,38em)and(-8em,-5em).. (pc);
%\node (labl) at (3em,-2.5em) {$[\ldr I]$};
%%%%
\node at (-6em,7.7em) {$\rightarrow$};
%%%
\node at (-10em,4.8em) {$\apsnodei$};
\node at (-10em,10.8em) {$\apsnodei$};
\node [pn] at (-10em,1.7em) {$\Theta$};
\node [pn] at (-10em,7.7em) {$\Delta$};
\node [pn] at (-10em,13.7em) {$\Gamma$};
\end{tikzpicture}
\caption{Graph expansions}
\label{fig:rewrexp}
\end{figure}
%\begin{enumerate}
%	\item predict the correct graph structure,
%	\item label the vertices in the graph to produce the correct proof.
%\end{enumerate}

Figure~\ref{fig:rewrexp} shows the two graph rewrite operations corresponding the introduction rule, with abstraction over a functor shown on the left of the figure and abstraction over an argument shown on the right of the figure.

Given that the graph expansions could apply indefinitely, there is a final graph operation `stop', which can apply only when the graph is connected, and which ends the graph generation process.  

The reader will surely have remarked that the graph rewrite operations of Figure~\ref{fig:rewrexp} are simply the graph contractions of the Lambek-van Benthem calculus LP of Figure~\ref{fig:contr:lp} but applied as graph expansions. This makes it very easy to see the rewrite operations are correct. To show that a graph is a correct LP proof, we need to show it contracts to a tree using the contractions of Figure~\ref{fig:contr:lp}. Take a graph $G$ constructed by application of the graph rewrites of Figures~\ref{fig:grcomp} and~\ref{fig:rewrexp}. We can order the rewrites such that all composition operations precede all expansion operations. The composition operations produce a tree, and the expansions produce a structure which can be contracted to a tree, because the expansions are simply the inverses of the contractions\footnote{The condition that the path through $\Delta$ does not pass through another par link is superfluous in LP without the tensor product `$\otimes$', since we can always reorder the operations in such a way that it is respected by starting at the outermost abstraction, then moving inside. However, this no longer holds when we add the product or move to the Lambek calculus.}.

We can also look at the graph rewrite operations from the perspective of the lambda calculus, where they correspond to the following (slightly odd) term construction rules.
\begin{align}
\label{exp:appl} M, N & \leadsto (M\, N)& \text{$M$, $N$ do not share variables}	 \\
P[M[N]] & \leadsto P[\lambda x. M[(x\, N)]] \\
\label{exp:arg} P[M[N]] & \leadsto P[\lambda x. M[(N\, x)]]
\end{align}

The term construction rules which are missing from a combinatorial perspective are the following.
\begin{align}
\label{notbeta}M[N]] & \leadsto M[(\lambda x.x)\, N)]]& \phantom{\text{$M$, $N$ do not share variables}} \\
\label{rewr:empty} M[N]] & \leadsto M[(N\, (\lambda x. x))]]
\end{align}
We can exclude~\ref{notbeta} because it only produces terms which are not beta normal, but the more important reason to exclude these two construction rules is that they are only needed to produce closed subterms, which correspond to empty antecedent derivations, and these are generally excluded in type-logical grammars. If want empty subterms, we can simply add~\ref{rewr:empty} and the corresponding expansion\footnote{The attentive reader will remark that this still excludes the case where, rather than just some of its subterms, the complete term assigned to a sentence is closed. However, we start from the assumption that a sentence contains at least one word, and therefore the computed term must have at least one variable.}.
 
\subsection{Example}

\begin{figure}
\begin{center}
\scalebox{0.75}{
\begin{tikzpicture}
\node (auxl) at (-6em,23.0em) {Aux\proofspace};
\node (aux) at (-6em,22em) {$\apsnodei$};
\node (na) at (-2em,22em) {$\apsnodei$};
%\node [tns] (t) at (12em,8.2em) {};
%\node at (12em,8.2em) {\appl};
%\node (sspp) at (12em,6em) {$\apsnodei$};
%\node (auxs) at (10em,10em) {\blb};
%\draw (t)--(auxs);
%\draw (t)--(sspp);
%\draw (t)--(na);
%\node (spp) at (16em,6em) {$\apsnodei$};
%\node (auxs) at (14em,2em) {$\apsnodei$};
%\node [tns] (tt) at (14em,4.0em) {};
%\node at (14em,4em) {\appl};
%\draw (tt) -- (auxs);
%\draw (tt) -- (sspp);
%\draw (tt) -- (spp);
%\node [par] (par) at (18em,7.8em) {};
%\node at (18em,7.8em) {$\textcolor{white}{\small \lambda}$};
%\node (pp) at (20em,6em) {$\atppr{}$};
%\node (ppd) at (20em,6em) {\phantom{x}};
%\node (tops) at (18em,10em) {$\apsnodei$};
%\draw (par)--(ppd);
%\draw (par)--(tops);
%\node (sppd) at (16em,6em) {\phantom{x}};
%\path[>=latex,->]  (par) edge (sppd);
%%%%
%\node (fils) at (21em,10em) {$\atnr$};
\node (filsl) at (-2em,23.0em) {fils...\proofspace};
%%%
\node (clamel) at (02em,23.0em) {clame\proofspace};
%\node (hs) at (20em,2em) {$\atsr$};
%\node (ss) at (04em,2em) {$\apsnodei$};
%\node (ssd) at (4em,2em) {\phantom{x}};
%\node (cs) at (9em,-2em) {$\apsnodei$};
%\node [tns] (t) at (9em,0.2em){};
%\node at (9em,0.2em){\appl};
%\draw (t) -- (auxs);
%\draw (t) -- (ssd);
%\draw (t) -- (cs);
%\node (clsu) at (06em,6em) {$\apsnodei$};
\node (clsud) at (06em,22em) {$\apsnodei$};
\node (clame) at (02em,22em) {$\apsnodei$};
%\node (clamed) at (02em,6em) {\phantom{M}};
%\node [tns] (tc) at (04em,4.2em) {};
%\node at (4em,4.2em) {\appl};
%\draw (tc) -- (clamed);
%\draw (tc) -- (clsud);
%\draw (tc) -- (ss);
%%%
%\node (il) at (33em,10em) {$\atnpr$};
\node at (06em,23.0em) {-t-il\proofspace};
%%%
%\node (je) at (36em,10em) {$\atnpr$};
\node at (8.8em,23.0em) {j'\proofspace};
\node at (8.8em,22em) {$\apsnodei$};
%%%
\node at (12em,23.0em) {apporte\proofspace};
\node (apporte) at (12em,22em) {$\apsnodei$};
\node (apob) at (16em,22em) {$\apsnodei$};
%\node (tvp) at (14em,18em) {$\apsnodei$};
%\node (appp) at (18em,18em) {$\apsnodei$};
%\node (vp) at (16em,14em) {$\apsnodei$};
%\node (apsu) at (20em,14em) {$\apsnodei$};
%\node (apc) at (18em,10em) {\phantom{x}};
% = 18,10.  -24,-12
%
%\node [tns] (t1) at (14em,20.2em) {};
%\node at (14em,20.2em) {\appl};
%\draw (t1) -- (apporte);
%\draw (t1) -- (apob);
%\draw (t1) -- (tvp);
%\node [tns] (t2) at (16em,16.2em) {};
%\node at (16em,16.2em) {\appl};
%\draw (t2) -- (vp);
%\draw (t2) -- (tvp);
%\draw (t2) -- (appp);
%\node [tns] (t3) at (18em,12em) {};
%\node at (18em,12em) {\appl};
%\draw (t3) -- (apc);
%\draw (t3) -- (vp);
%\draw (t3) -- (apsu);
%%
\node at (16em,23.0em) {le salut...\proofspace};
%\draw (appp) to [out=50,in=330] (par);
%\node (lesalut) at (49em,10em) {$\atnpr$};
\end{tikzpicture}}
\end{center}
\caption{Initial configuration for the proof net generation}
\label{fig:bu1}
\end{figure}

Returning to our previous example, we will now show how to generate the semantic proof net of Figure~\ref{fig:sem} using the graph rewrite operations. Our initial configuration is shown in Figure~\ref{fig:bu1}. In this configuration there are 42 possible applications of the composition/application rule and 14 possibilities for the expansion/abstraction rules. The expansion rules can apply to single vertices, where they correspond to replacing a variable $y$ by either $\lambda x.(x\, y)$ (this operation is generally called lifting) or $\lambda x.(y\, x)$ (eta expansion).

\begin{figure}
\begin{center}	
\scalebox{0.75}{
\begin{tikzpicture}
\node (auxl) at (-6em,23.0em) {Aux\proofspace};
\node (aux) at (-6em,22em) {$\apsnodei$};
\node (na) at (-2em,22em) {$\apsnodei$};
\node [tns] (t) at (-4em,20.2em) {};
\node at (-4em,20.2em) {\appl};
\node (sspp) at (-4em,18em) {$\apsnodei$};
%\node (auxs) at (10em,10em) {\blb};
\draw (t)--(aux);
\draw (t)--(sspp);
\draw (t)--(na);
%\node (spp) at (16em,6em) {$\apsnodei$};
%\node (auxs) at (14em,2em) {$\apsnodei$};
%\node [tns] (tt) at (14em,4.0em) {};
%\node at (14em,4em) {\appl};
%\draw (tt) -- (auxs);
%\draw (tt) -- (sspp);
%\draw (tt) -- (spp);
%\node [par] (par) at (18em,7.8em) {};
%\node at (18em,7.8em) {$\textcolor{white}{\small \lambda}$};
%\node (pp) at (20em,6em) {$\atppr{}$};
%\node (ppd) at (20em,6em) {\phantom{x}};
%\node (tops) at (18em,10em) {$\apsnodei$};
%\draw (par)--(ppd);
%\draw (par)--(tops);
%\node (sppd) at (16em,6em) {\phantom{x}};
%\path[>=latex,->]  (par) edge (sppd);
%%%%
%\node (fils) at (21em,10em) {$\atnr$};
\node (filsl) at (-2em,23.0em) {fils...\proofspace};
%%%
\node (clamel) at (02em,23.0em) {clame\proofspace};
%\node (hs) at (20em,2em) {$\atsr$};
\node (ss) at (04em,18em) {$\apsnodei$};
%\node (ssd) at (4em,2em) {\phantom{x}};
%\node (cs) at (9em,-2em) {$\apsnodei$};
%\node [tns] (t) at (9em,0.2em){};
%\node at (9em,0.2em){\appl};
%\draw (t) -- (auxs);
%\draw (t) -- (ssd);
%\draw (t) -- (cs);
%\node (clsu) at (06em,6em) {$\apsnodei$};
\node (clsud) at (06em,22em) {$\apsnodei$};
\node (clame) at (02em,22em) {$\apsnodei$};
%\node (clamed) at (02em,6em) {\phantom{M}};
\node [tns] (tc) at (04em,20.2em) {};
\node at (4em,20.2em) {\appl};
\draw (tc) -- (clame);
\draw (tc) -- (clsud);
\draw (tc) -- (ss);
%%%
%\node (il) at (33em,10em) {$\atnpr$};
\node at (06em,23.0em) {-t-il\proofspace};
%%%
%\node (je) at (36em,10em) {$\atnpr$};
\node at (8.8em,23.0em) {j'\proofspace};
\node at (8.8em,22em) {$\apsnodei$};
%%%
\node at (12em,23.0em) {apporte\proofspace};
\node (apporte) at (12em,22em) {$\apsnodei$};
\node (apob) at (16em,22em) {$\apsnodei$};
\node (tvp) at (14em,18em) {$\apsnodei$};
%\node (appp) at (18em,18em) {$\apsnodei$};
%\node (vp) at (16em,14em) {$\apsnodei$};
%\node (apsu) at (20em,14em) {$\apsnodei$};
%\node (apc) at (18em,10em) {\phantom{x}};
% = 18,10.  -24,-12
%
\node [tns] (t1) at (14em,20.2em) {};
\node at (14em,20.2em) {\appl};
\draw (t1) -- (apporte);
\draw (t1) -- (apob);
\draw (t1) -- (tvp);
%\node [tns] (t2) at (16em,16.2em) {};
%\node at (16em,16.2em) {\appl};
%\draw (t2) -- (vp);
%\draw (t2) -- (tvp);
%\draw (t2) -- (appp);
%\node [tns] (t3) at (18em,12em) {};
%\node at (18em,12em) {\appl};
%\draw (t3) -- (apc);
%\draw (t3) -- (vp);
%\draw (t3) -- (apsu);
%%
\node at (16em,23.0em) {le salut...\proofspace};
%\draw (appp) to [out=50,in=330] (par);
%\node (lesalut) at (49em,10em) {$\atnpr$};
\end{tikzpicture}}

\bigskip

\scalebox{0.75}{
\begin{tikzpicture}
\node (auxl) at (-6em,23.0em) {Aux\proofspace};
\node (aux) at (-6em,22em) {$\apsnodei$};
\node (na) at (-2em,22em) {$\apsnodei$};
\node [tns] (t) at (-4em,20.2em) {};
\node at (-4em,20.2em) {\appl};
\node (sspp) at (-4em,18em) {$\apsnodei$};
%\node (auxs) at (10em,10em) {\blb};
\draw (t)--(aux);
\draw (t)--(sspp);
\draw (t)--(na);
%\node (spp) at (16em,6em) {$\apsnodei$};
%\node (auxs) at (14em,2em) {$\apsnodei$};
%\node [tns] (tt) at (14em,4.0em) {};
%\node at (14em,4em) {\appl};
%\draw (tt) -- (auxs);
%\draw (tt) -- (sspp);
%\draw (tt) -- (spp);
%\node [par] (par) at (18em,7.8em) {};
%\node at (18em,7.8em) {$\textcolor{white}{\small \lambda}$};
%\node (pp) at (20em,6em) {$\atppr{}$};
%\node (ppd) at (20em,6em) {\phantom{x}};
%\node (tops) at (18em,10em) {$\apsnodei$};
%\draw (par)--(ppd);
%\draw (par)--(tops);
%\node (sppd) at (16em,6em) {$\apsnodei$};
%\path[>=latex,->]  (par) edge (sppd);
%%%%
%\node (fils) at (21em,10em) {$\atnr$};
\node (filsl) at (-2em,23.0em) {fils...\proofspace};
%%%
\node (clamel) at (02em,23.0em) {clame\proofspace};
%\node (hs) at (20em,2em) {$\atsr$};
\node (ss) at (04em,18em) {$\apsnodei$};
%\node (ssd) at (4em,2em) {\phantom{x}};
%\node (cs) at (9em,-2em) {$\apsnodei$};
%\node [tns] (t) at (9em,0.2em){};
%\node at (9em,0.2em){\appl};
%\draw (t) -- (auxs);
%\draw (t) -- (ssd);
%\draw (t) -- (cs);
%\node (clsu) at (06em,6em) {$\apsnodei$};
\node (clsud) at (06em,22em) {$\apsnodei$};
\node (clame) at (02em,22em) {$\apsnodei$};
%\node (clamed) at (02em,6em) {\phantom{M}};
\node [tns] (tc) at (04em,20.2em) {};
\node at (4em,20.2em) {\appl};
\draw (tc) -- (clame);
\draw (tc) -- (clsud);
\draw (tc) -- (ss);
%%%
%\node (il) at (33em,10em) {$\atnpr$};
\node at (06em,23.0em) {-t-il\proofspace};
%%%
%\node (je) at (36em,10em) {$\atnpr$};
\node at (18.0em,19.0em) {j'\proofspace};
%\node at (8.8em,22em) {$\apsnodei$};
%%%
\node at (12em,23.0em) {apporte\proofspace};
\node (apporte) at (12em,22em) {$\apsnodei$};
\node (apob) at (16em,22em) {$\apsnodei$};
\node (tvp) at (14em,18em) {$\apsnodei$};
\node (appp) at (18em,18em) {$\apsnodei$};
\node (vp) at (16em,14em) {$\apsnodei$};
%\node (apsu) at (20em,14em) {$\apsnodei$};
%\node (apc) at (18em,10em) {\phantom{x}};
% = 18,10.  -24,-12
%
\node [tns] (t1) at (14em,20.2em) {};
\node at (14em,20.2em) {\appl};
\draw (t1) -- (apporte);
\draw (t1) -- (apob);
\draw (t1) -- (tvp);
\node [tns] (t2) at (16em,16.2em) {};
\node at (16em,16.2em) {\appl};
\draw (t2) -- (vp);
\draw (t2) -- (tvp);
\draw (t2) -- (appp);
%\node [tns] (t3) at (18em,12em) {};
%\node at (18em,12em) {\appl};
%\draw (t3) -- (apc);
%\draw (t3) -- (vp);
%\draw (t3) -- (apsu);
%%
\node at (16em,23.0em) {le salut...\proofspace};
%\draw (appp) to [out=50,in=330] (par);
%\node (lesalut) at (49em,10em) {$\atnpr$};
\end{tikzpicture}}
\end{center}
\caption{The top row shows first series of applications starting from Figure~\ref{fig:bu1}. On the bottom row, an additional application has been performed.}
\label{fig:bu2}
\end{figure}
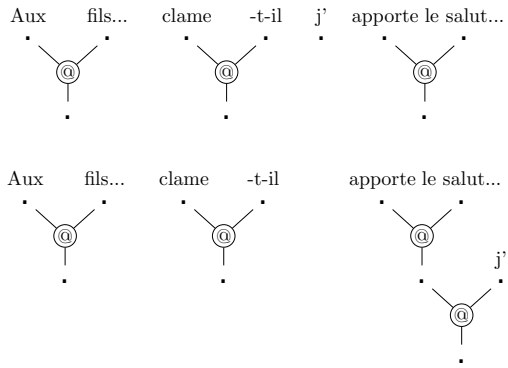
In this configuration, the correct applications are shown at the top row of Figure~\ref{fig:bu2}. The bottom row shows how we can then combine the tree for ``apporte le salut...'' with the tree for ``j''. The three applications on the top row are unordered. We can apply them one at a time, or all at once. The application on the bottom row can only be correctly performed after ``apporte le salut...'' has been created.

However, there is nothing which forbids any of the other connections. It will be up to the neural model to learn that combining ``-t-il'' as functor with ``Aux'' is not a very promising start for the graph generation process.

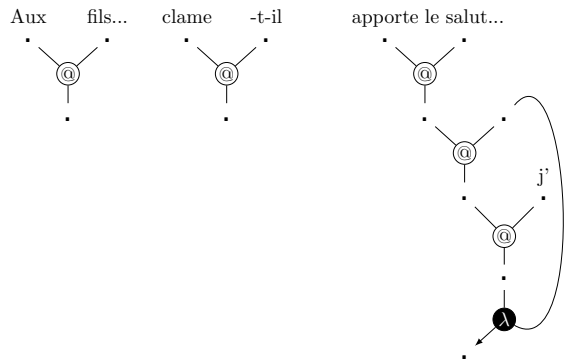
\begin{figure}
\begin{center}
\scalebox{0.75}{
\begin{tikzpicture}
\node (auxl) at (-6em,23.0em) {Aux\proofspace};
\node (aux) at (-6em,22em) {$\apsnodei$};
\node (na) at (-2em,22em) {$\apsnodei$};
\node [tns] (t) at (-4em,20.2em) {};
\node at (-4em,20.2em) {\appl};
\node (sspp) at (-4em,18em) {$\apsnodei$};
%\node (auxs) at (10em,10em) {\blb};
\draw (t)--(aux);
\draw (t)--(sspp);
\draw (t)--(na);
%\node (spp) at (16em,6em) {$\apsnodei$};
%\node (auxs) at (14em,2em) {$\apsnodei$};
%\node [tns] (tt) at (14em,4.0em) {};
%\node at (14em,4em) {\appl};
%\draw (tt) -- (auxs);
%\draw (tt) -- (sspp);
%\draw (tt) -- (spp);
\node [par] (par) at (18em,7.8em) {};
\node at (18em,7.8em) {$\textcolor{white}{\small \lambda}$};
%\node (pp) at (20em,6em) {$\atppr{}$};
%\node (ppd) at (20em,6em) {\phantom{x}};
\node (tops) at (18em,10em) {$\apsnodei$};
%\draw (par)--(ppd);
\draw (par)--(tops);
\node (sppd) at (16em,6em) {$\apsnodei$};
\path[>=latex,->]  (par) edge (sppd);
%%%%
%\node (fils) at (21em,10em) {$\atnr$};
\node (filsl) at (-2em,23.0em) {fils...\proofspace};
%%%
\node (clamel) at (02em,23.0em) {clame\proofspace};
%\node (hs) at (20em,2em) {$\atsr$};
\node (ss) at (04em,18em) {$\apsnodei$};
%\node (ssd) at (4em,2em) {\phantom{x}};
%\node (cs) at (9em,-2em) {$\apsnodei$};
%\node [tns] (t) at (9em,0.2em){};
%\node at (9em,0.2em){\appl};
%\draw (t) -- (auxs);
%\draw (t) -- (ssd);
%\draw (t) -- (cs);
%\node (clsu) at (06em,6em) {$\apsnodei$};
\node (clsud) at (06em,22em) {$\apsnodei$};
\node (clame) at (02em,22em) {$\apsnodei$};
%\node (clamed) at (02em,6em) {\phantom{M}};
\node [tns] (tc) at (04em,20.2em) {};
\node at (4em,20.2em) {\appl};
\draw (tc) -- (clame);
\draw (tc) -- (clsud);
\draw (tc) -- (ss);
%%%
%\node (il) at (33em,10em) {$\atnpr$};
\node at (06em,23.0em) {-t-il\proofspace};
%%%
%\node (je) at (36em,10em) {$\atnpr$};
\node at (20.0em,15.0em) {j'\proofspace};
%\node at (8.8em,22em) {$\apsnodei$};
%%%
\node at (12em,23.0em) {apporte\proofspace};
\node (apporte) at (12em,22em) {$\apsnodei$};
\node (apob) at (16em,22em) {$\apsnodei$};
\node (tvp) at (14em,18em) {$\apsnodei$};
\node (appp) at (18em,18em) {$\apsnodei$};
\node (vp) at (16em,14em) {$\apsnodei$};
5\node (apsu) at (20em,14em) {$\apsnodei$};
%\node (apc) at (18em,10em) {\phantom{x}};
% = 18,10.  -24,-12
%
\node [tns] (t1) at (14em,20.2em) {};
\node at (14em,20.2em) {\appl};
\draw (t1) -- (apporte);
\draw (t1) -- (apob);
\draw (t1) -- (tvp);
\node [tns] (t2) at (16em,16.2em) {};
\node at (16em,16.2em) {\appl};
\draw (t2) -- (vp);
\draw (t2) -- (tvp);
\draw (t2) -- (appp);
\node [tns] (t3) at (18em,12em) {};
\node at (18em,12em) {\appl};
\draw (t3) -- (apc);
\draw (t3) -- (vp);
\draw (t3) -- (apsu);
\node at (16em,23.0em) {le salut...\proofspace};
\draw (appp) to [out=50,in=330] (par);
%\node (lesalut) at (49em,10em) {$\atnpr$};
\end{tikzpicture}}
\end{center}
\caption{The structure of Figure~\ref{fig:bu2} after an argument expansion}
\label{fig:bu3}
\end{figure}

Given the graph on the bottom of Figure~\ref{fig:bu2}, we can apply one of the expansion rules, choosing the sister node of ``j'' to expand, creating a two new links, a tensor link with its root as the sister of ``j'' and a par link at the root of the complete structure. This par link connects to the right daughter of the newly introduced link, and is therefore an instance of an abstraction over an argument (rather than a function).

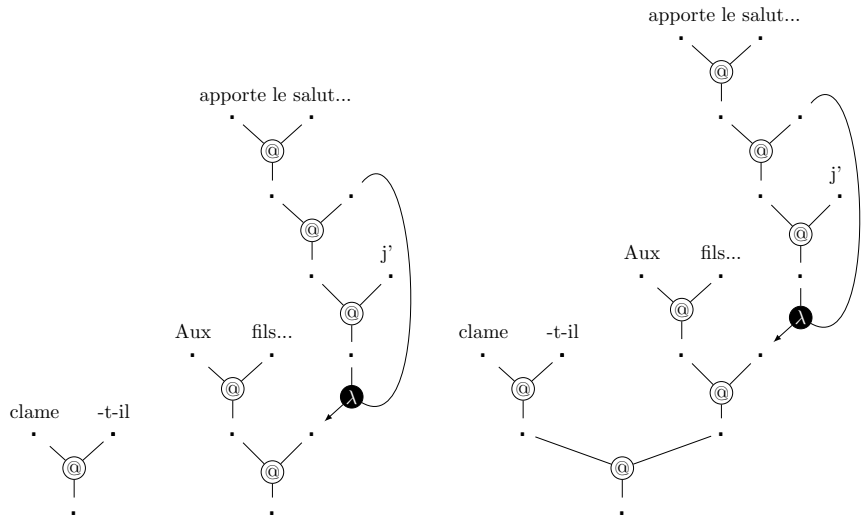
\begin{figure}
\begin{center}
\scalebox{0.75}{
\begin{tikzpicture}
\node (auxl) at (10em,11.0em) {Aux\proofspace};
\node (aux) at (10em,10em) {$\apsnodei$};
\node (na) at (14em,10em) {$\apsnodei$};
\node [tns] (t) at (12em,8.2em) {};
\node at (12em,8.2em) {\appl};
\node (sspp) at (12em,6em) {$\apsnodei$};
\node (auxs) at (10em,10em) {\blb};
\draw (t)--(auxs);
\draw (t)--(sspp);
\draw (t)--(na);
\node (spp) at (16em,6em) {$\apsnodei$};
\node (auxs) at (14em,2em) {$\apsnodei$};
\node [tns] (tt) at (14em,4.0em) {};
\node at (14em,4em) {\appl};
\draw (tt) -- (auxs);
\draw (tt) -- (sspp);
\draw (tt) -- (spp);
\node [par] (par) at (18em,7.8em) {};
\node at (18em,7.8em) {$\textcolor{white}{\small \lambda}$};
%\node (pp) at (20em,6em) {$\atppr{}$};
%\node (ppd) at (20em,6em) {\phantom{x}};
\node (tops) at (18em,10em) {$\apsnodei$};
%\draw (par)--(ppd);
\draw (par)--(tops);
\node (sppd) at (16em,6em) {\phantom{x}};
\path[>=latex,->]  (par) edge (sppd);
%%%%
%\node (fils) at (21em,10em) {$\atnr$};
\node (filsl) at (14em,11.0em) {fils...\proofspace};
%%%
\node (clamel) at (02em,7.0em) {clame\proofspace};
%\node (hs) at (20em,2em) {$\atsr$};
\node (ss) at (04em,2em) {$\apsnodei$};
\node (ssd) at (4em,2em) {\phantom{x}};
%\node (cs) at (9em,-2em) {$\apsnodei$};
%\node [tns] (t) at (9em,0.2em){};
%\node at (9em,0.2em){\appl};
%\draw (t) -- (auxs);
%\draw (t) -- (ssd);
%\draw (t) -- (cs);
\node (clsu) at (06em,6em) {$\apsnodei$};
\node (clsud) at (06em,6em) {$\apsnodei$};
\node (clame) at (02em,6em) {$\apsnodei$};
\node (clamed) at (02em,6em) {\phantom{M}};
\node [tns] (tc) at (04em,4.2em) {};
\node at (4em,4.2em) {\appl};
\draw (tc) -- (clamed);
\draw (tc) -- (clsud);
\draw (tc) -- (ss);
%%%
%\node (il) at (33em,10em) {$\atnpr$};
\node at (06em,7.0em) {-t-il\proofspace};
%%%
%\node (je) at (36em,10em) {$\atnpr$};
\node at (19.8em,15.0em) {j'\proofspace};
%%%
\node at (12em,23.0em) {apporte\proofspace};
\node (apporte) at (12em,22em) {$\apsnodei$};
\node (apob) at (16em,22em) {$\apsnodei$};
\node (tvp) at (14em,18em) {$\apsnodei$};
\node (appp) at (18em,18em) {$\apsnodei$};
\node (vp) at (16em,14em) {$\apsnodei$};
\node (apsu) at (20em,14em) {$\apsnodei$};
\node (apc) at (18em,10em) {\phantom{x}};
% = 18,10.  -24,-12
%
\node [tns] (t1) at (14em,20.2em) {};
\node at (14em,20.2em) {\appl};
\draw (t1) -- (apporte);
\draw (t1) -- (apob);
\draw (t1) -- (tvp);
\node [tns] (t2) at (16em,16.2em) {};
\node at (16em,16.2em) {\appl};
\draw (t2) -- (vp);
\draw (t2) -- (tvp);
\draw (t2) -- (appp);
\node [tns] (t3) at (18em,12em) {};
\node at (18em,12em) {\appl};
\draw (t3) -- (apc);
\draw (t3) -- (vp);
\draw (t3) -- (apsu);
\node at (16em,23.0em) {le salut...\proofspace};
\draw (appp) to [out=50,in=330] (par);
%\node (lesalut) at (49em,10em) {$\atnpr$};
\end{tikzpicture}}
\scalebox{0.75}{
\begin{tikzpicture}
\node (auxl) at (10em,11.0em) {Aux\proofspace};
\node (aux) at (10em,10em) {$\apsnodei$};
\node (na) at (14em,10em) {$\apsnodei$};
\node [tns] (t) at (12em,8.2em) {};
\node at (12em,8.2em) {\appl};
\node (sspp) at (12em,6em) {$\apsnodei$};
\node (auxs) at (10em,10em) {\blb};
\draw (t)--(auxs);
\draw (t)--(sspp);
\draw (t)--(na);
\node (spp) at (16em,6em) {$\apsnodei$};
\node (auxs) at (14em,2em) {$\apsnodei$};
\node [tns] (tt) at (14em,4.0em) {};
\node at (14em,4em) {\appl};
\draw (tt) -- (auxs);
\draw (tt) -- (sspp);
\draw (tt) -- (spp);
\node [par] (par) at (18em,7.8em) {};
\node at (18em,7.8em) {$\textcolor{white}{\small \lambda}$};
%\node (pp) at (20em,6em) {$\atppr{}$};
%\node (ppd) at (20em,6em) {\phantom{x}};
\node (tops) at (18em,10em) {$\apsnodei$};
%\draw (par)--(ppd);
\draw (par)--(tops);
\node (sppd) at (16em,6em) {\phantom{x}};
\path[>=latex,->]  (par) edge (sppd);
%%%%
%\node (fils) at (21em,10em) {$\atnr$};
\node (filsl) at (14em,11.0em) {fils...\proofspace};
%%%
\node (clamel) at (02em,7.0em) {clame\proofspace};
%\node (hs) at (20em,2em) {$\atsr$};
\node (ss) at (04em,2em) {$\apsnodei$};
\node (ssd) at (4em,2em) {\phantom{x}};
\node (cs) at (9em,-2em) {$\apsnodei$};
\node [tns] (t) at (9em,0.2em){};
\node at (9em,0.2em){\appl};
\draw (t) -- (auxs);
\draw (t) -- (ssd);
\draw (t) -- (cs);
\node (clsu) at (06em,6em) {$\apsnodei$};
\node (clsud) at (06em,6em) {$\apsnodei$};
\node (clame) at (02em,6em) {$\apsnodei$};
\node (clamed) at (02em,6em) {\phantom{M}};
\node [tns] (tc) at (04em,4.2em) {};
\node at (4em,4.2em) {\appl};
\draw (tc) -- (clamed);
\draw (tc) -- (clsud);
\draw (tc) -- (ss);
%%%
%\node (il) at (33em,10em) {$\atnpr$};
\node at (06em,7.0em) {-t-il\proofspace};
%%%
%\node (je) at (36em,10em) {$\atnpr$};
\node at (19.8em,15.0em) {j'\proofspace};
%%%
\node at (12em,23.0em) {apporte\proofspace};
\node (apporte) at (12em,22em) {$\apsnodei$};
\node (apob) at (16em,22em) {$\apsnodei$};
\node (tvp) at (14em,18em) {$\apsnodei$};
\node (appp) at (18em,18em) {$\apsnodei$};
\node (vp) at (16em,14em) {$\apsnodei$};
\node (apsu) at (20em,14em) {$\apsnodei$};
\node (apc) at (18em,10em) {\phantom{x}};
% = 18,10.  -24,-12
%
\node [tns] (t1) at (14em,20.2em) {};
\node at (14em,20.2em) {\appl};
\draw (t1) -- (apporte);
\draw (t1) -- (apob);
\draw (t1) -- (tvp);
\node [tns] (t2) at (16em,16.2em) {};
\node at (16em,16.2em) {\appl};
\draw (t2) -- (vp);
\draw (t2) -- (tvp);
\draw (t2) -- (appp);
\node [tns] (t3) at (18em,12em) {};
\node at (18em,12em) {\appl};
\draw (t3) -- (apc);
\draw (t3) -- (vp);
\draw (t3) -- (apsu);
\node at (16em,23.0em) {le salut...\proofspace};
\draw (appp) to [out=50,in=330] (par);
%\node (lesalut) at (49em,10em) {$\atnpr$};
\end{tikzpicture}}
\end{center}
\caption{The final two applications}
\label{fig:buend}
\end{figure}

We can them combine the three components with two more graph composition/application rules to obtain the graph shown on the right of Figure~\ref{fig:buend}. This is the same graph as the semantic graph of Figure~\ref{fig:sem} and it therefore encodes the correct semantics.

Before continuing, let's give some basic analysis of the combinatorics for the different steps. Given $c$ components, there are exactly $c(c-1)$ ways to combine them using the application rule. 
 For the expansion rules, given a component with $n$ nodes, there are $\leq n^2 + n - 1$ ways to select a node and one of its descendants\footnote{The worst case occurs when the tree is maximally unbalanced, such as the subtree ``j'apporte le salut...'' of Figure~\ref{fig:buend}, since it maximises the number of nodes related by the descendant relation. The best case occurs when all leaves of the tree are at the same depth, which minimises the nodes related by the descendant relation and results in less than $\lceil n \log_2 n \rceil$ possibilities for selection of a vertex and one of its descendants.} and double that number once we distinguish between abstraction over a functor and an argument. For the graph shown on the right of  Figure~\ref{fig:buend}, there are 16 vertices and a total of 60 possibilities of node-descendant combination, which is considerably less than the maximum of 3840 for the number of vertices.

Given the combinatorics of the expansion step, it makes sense to restrict its application as much as possible. One simple but useful restriction is that we never do expansions when the root of the expansion is on the left branch on an application link. These expansions produce a beta redex, and when we perform expansions from the root node upwards, the beta redex will remain in the final term\footnote{For example, we can remove a beta redex $(\lambda x. M)\, N$ by a functor expansion as follows $\lambda y. ((y\, \lambda x.M)\, N)$, but an outermost expansion would have produced $\lambda y. ((y M')\, N)$ first (where $M'$ is $\lambda x.M$ before the expansion introducing $x$), followed by the expansion of $M'$ to $\lambda x. M$.}.

We could also be tempted to think we can restrict expansions to the root node of components. However, although this is true for a system with expansions~\ref{exp:appl}-\ref{exp:arg} and \ref{rewr:empty}, such a restriction would need more careful analysis in a system with only expansions~\ref{exp:appl}-\ref{exp:arg}. To illustrate the potential pitfall, consider the term $\lambda x. f (\lambda y.(y\, x))$. Terms of this form are sometimes called `argument lowering', corresponding at the formula level to $((A\multimap B)\multimap B) \multimap C \vdash A\multimap C$. When we want to produce this term starting from $f$, the only valid step we can take is argument expansion to produce $\lambda x. (f\,x)$. To produce the final term, we need to perform function expansion at $x$, but this produces an abstraction which is not at the root node. There appear to be ways around this, for example by considering newly introduced variables such as $x$ in our example terms as a type of root node as well, allowing expansions to apply there.

\subsection{Graph neural networks} 
 
We have seen how graph generation can be done in a way which generates only (linear logic) proof nets, and that these graphs/proofs correspond to the derivational meaning of the sentence. What remains to be shown is how we can integrate graph generation with neural networks.

Graph neural networks are an active research area, and would seem a priori well-suited for generating graph-based representations of proofs. However, our graphs have a number of particularities which restrict the architectures we can use. First of all, there is a linear order on the initial sequence of vertices. We will assume in what follows that the positional information that is part of the BERT encodings for each word \cite{devlin2018bert} suffices for taking word order into account, but we could envisage architectures encoding word order differently. We also need some way to distinguish the three vertices adjacent to a hyperlink in the graph, which seems to require some form of edge labelling, for example representing the link shown below on the left as shown on the right. 

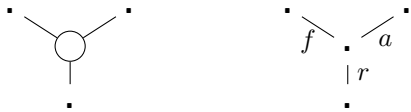
\begin{center}
\begin{tikzpicture}[scale=0.75]
\node (ab) at (3em,4.8em) {$\apsnodei$};
\node (a) at (0,9.6em) {$\apsnodei$};
\node (b) at (6em,9.6em) {$\apsnodei$};
\node[tns] (c) at (3em,7.668em) {};
\draw (c) -- (ab);
\draw (c) -- (a);
\draw (c) -- (b);
%
%%%%
\node (ab) at (17em,4.8em) {$\apsnodei$};
\node (a) at (14em,9.6em) {$\apsnodei$};
\node (b) at (20em,9.6em) {$\apsnodei$};
\node (c) at (17em,7.668em) {$\apsnodei$};
\draw (c) -- (ab) node[midway,right] {$r$};
\draw (c) -- (a);
\draw (c) -- (b);
\node at (15.0em,7.9em) {$f\proofspace$};
\node at (18.9em,8em) {$a\proofspace$};
\end{tikzpicture}
\end{center}

This means we are using heterogeneous graphs, with (at least) two types of vertices, the first type corresponding to the vertices of the original abstract proof structure, and the second to the hyperedges of the abstract proof structure. This is a fairly common way of interpreting a hypergraph as a standard graph.

Transforming the graph generation mechanism into a neural network requires us to make choices for how to represent the initial state, how to represent the possible parse actions, implement a way to select an action among the alternatives, and finally a way to label the vertices in order to obtain the detailed formula information.
We will discuss each of these in turn.

\subsubsection*{Initial states}

 The simplest way to do represent the initial state is to represent each word by an isolated vertex, with BERT embedding of the word --- a vector of floating point values encoding information about the context and relative word positions --- at the corresponding vertex. %We can imagine more complicated strategies, providing initial connections between the words to allow for message passing between vertices from the initial stage.
%For the initial state is simply a BERT embedding for each word. 

\begin{center}
\scalebox{0.75}{
\begin{tikzpicture}
\node (auxl) at (-6em,23.0em) {Aux\proofspace};
\node (aux) at (-6em,22em) {$\apsnodei$};
\node (na) at (-2em,22em) {$\apsnodei$};
%%%%
\node (filsl) at (-2em,23.0em) {fils...\proofspace};
%%%
\node at (2.0em,23.0em) {j'\proofspace};
\node at (2.0em,22em) {$\apsnodei$};
%%%
\node at (6em,23.0em) {apporte\proofspace};
\node (apporte) at (6em,22em) {$\apsnodei$};
\node (apob) at (10em,22em) {$\apsnodei$};
\node at (10em,23.0em) {le salut...\proofspace};
\end{tikzpicture}}
\end{center}

We can connect each word to its neighbours, producing the following initial configuration. %However, this requires us to use multigraphs 
\begin{center}
\scalebox{0.75}{
\begin{tikzpicture}
\node (auxl) at (-6em,23.0em) {Aux\proofspace};
\node (aux) at (-6em,22em) {$\apsnodei$};
\node (na) at (-2em,22em) {$\apsnodei$};
%%%%
\node (filsl) at (-2em,23.0em) {fils...\proofspace};
%%%
\node at (2.0em,23.0em) {j'\proofspace};
\node (jn) at (2.0em,22em) {$\apsnodei$};
%%%
\node at (6em,23.0em) {apporte\proofspace};
\node (apporte) at (6em,22em) {$\apsnodei$};
\node (apob) at (10em,22em) {$\apsnodei$};
\node at (10em,23.0em) {le salut...\proofspace};
\draw (aux) -- (na);
\draw (na) -- (jn);
\draw (jn) -- (apporte);
\draw (apporte) -- (apob);
\end{tikzpicture}}
\end{center}

We can also use a fully connected graph. 

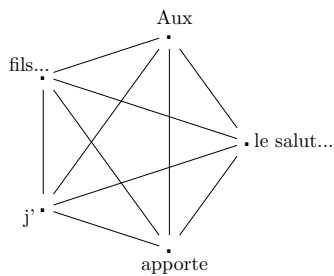
\begin{center}
\scalebox{0.75}{
\begin{tikzpicture}
\tikzstyle{EdgeStyle}=[->,>=stealth,thick,bend left=20]
\foreach \phi in {1,...,5}{
    \node (v\phi) at (360/5 * \phi:2cm) {$\apsnodei$};
};
\draw (v1) -- (v2);
\draw (v1) -- (v3);
\draw (v1) -- (v4);
\draw (v1) -- (v5);
\draw (v2) -- (v3);
\draw (v2) -- (v4);
\draw (v2) -- (v5);
\draw (v3) -- (v4);
\draw (v3) -- (v5);
\draw (v4) -- (v5);
\node (a) at (360/5 * 1:2.3cm) {\grow{Aux}};
\node (f) at (360/5 * 2:2.3cm) {\grow{fils...}};
\node (j) at (360/5 * 3:2.3cm) {\grow{j'}};
\node (ap) at (360/5 * 4:2.3cm) {\grow{apporte}};
\node (s) at (360/5 * 5:2.3cm) {\grow{\quad\qquad le salut...}};
\end{tikzpicture}}
\end{center}

The advantage of the initial graph being more connected rather than a set of isolated vertices is that we can use a graph convolution or message passing step to aggregate the combined information of the BERT vectors of our words. 

The drawback of having connections in the initial graph is that when we want to cast the problem as an edge prediction task, we must use multigraphs (i.e.\ graphs allowing multiple connections between the same vertices), otherwise we cannot predict connections at all for the case above (because all connections are already there) or exclude connections between neighbouring vertices for the line graph, which would exclude many correct linkings. 

Given that few graph neural network architectures support multigraphs, we can therefore choose to use the following structure for the line graph instead. 

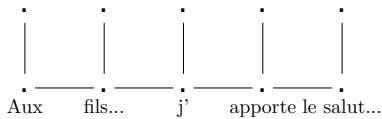
\begin{center}
\scalebox{0.75}{
\begin{tikzpicture}
\node (auxl) at (-6em,17.0em) {Aux\proofspace};
\node (auxt) at (-6em,22em) {$\apsnodei$};
\node (nat) at (-2em,22em) {$\apsnodei$};
%%%%
\node (filsl) at (-2em,17.0em) {fils...\proofspace};
%%%
\node at (2.0em,17.0em) {j'\proofspace};
\node (jnt) at (2.0em,22em) {$\apsnodei$};
%%%
\node at (6em,17.0em) {apporte\proofspace};
\node (apportet) at (6em,22em) {$\apsnodei$};
\node (apobt) at (10em,22em) {$\apsnodei$};
\node at (10em,17.0em) {le salut...\proofspace};
\node (aux) at (-6em,18em) {$\apsnodei$};
\node (na) at (-2em,18em) {$\apsnodei$};
\node (jn) at (2.0em,18em) {$\apsnodei$};
\node (apporte) at (6em,18em) {$\apsnodei$};
\node (apob) at (10em,18em) {$\apsnodei$};
\draw (aux) -- (auxt);
\draw (na) -- (nat);
\draw (jn) -- (jnt);
\draw (apportet) -- (apporte);
\draw (apob) -- (apobt);
\draw (aux) -- (na);
\draw (na) -- (jn);
\draw (jn) -- (apporte);
\draw (apporte) -- (apob);
\end{tikzpicture}}
\end{center}

And the structure below for the fully connected graph. %The additional connections in these graphs are then simply dense layers in the neural network.

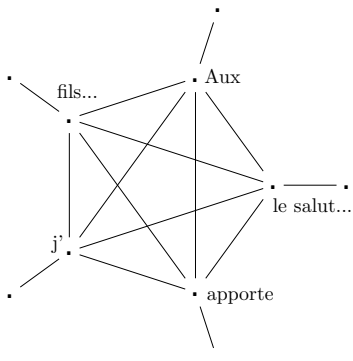
\begin{center}
\scalebox{0.75}{
\begin{tikzpicture}
\tikzstyle{EdgeStyle}=[->,>=stealth,thick,bend left=20]
\foreach \phi in {1,...,5}{
    \node (v\phi) at (360/5 * \phi:2cm) {$\apsnodei$};
};
\draw (v1) -- (v2);
\draw (v1) -- (v3);
\draw (v1) -- (v4);
\draw (v1) -- (v5);
\draw (v2) -- (v3);
\draw (v2) -- (v4);
\draw (v2) -- (v5);
\draw (v3) -- (v4);
\draw (v3) -- (v5);
\draw (v4) -- (v5);
\node (a) at (60:2.2cm) {\grow{Aux}};
\node (f) at (132:2.2cm) {\grow{fils...}};
\node (j) at (210:2.1cm) {\grow{j'}};
\node (ap) at (306:2.4cm) {\grow{apporte}};
\node (s) at (350:2.2cm) {\grow{\quad\qquad le salut...}};
\tikzstyle{EdgeStyle}=[->,>=stealth,thick,bend left=20]
\foreach \phi in {1,...,5}{
    \node (w\phi) at (360/5 * \phi:3.3cm) {$\apsnodei$};
};
\draw (v1) -- (w1);
\draw (v2) -- (w2);
\draw (v3) -- (w3);
\draw (v4) -- (w4);
\draw (v5) -- (w5);
\end{tikzpicture}}
\end{center}

In these structures, we keep the advantages of allowing message passing between the different word representations, but can still cast the problem as an edge prediction task without having to use multigraphs. 

\subsubsection*{Parser actions as edge prediction}

The easiest way to implement the graph generation parser actions as graph neural networks is to implement them as directed edge predictions. In this type of graph neural network, we represent graph composition as a directed edge from functor to argument. The expansion/abstraction rule is slightly more complicated. As we have seen in Section~\ref{sec:gg}, we need to select two vertices, one the ancestor of the other. There are two complications: the first is that we need to distinguish between the two different expansion rules, and the second is that we can select the same vertex twice. We can distinguish between two edges between the same pair of nodes using labelled edges in multigraphs, and we can represent the root and leaf node being identical by a self-loop (again labeled) in the graph, but there are relatively few graph neural networks permitting both self-loops and multi-graphs.

We can avoid these complications by adding new vertices representing possible abstractions to the graph, one for each root of a component and one for each introduced variable at a previous expansion step.  

The figure below sketches what this would look like. The simplest possible situation is shown on the left. Here we have a single vertex (at the top of the figure) representing variable $f$ and the abstraction root below it. The two possibilities for abstraction are then represented by the directed arrows. The arrow in green transforms $f$ into $\lambda x. (f x)$ ($f$ is a functor in this situation, and the arrow goes from functor to argument), whereas the other arrow transforms $f$ to $\lambda x. (x\,f)$ (where $f$ is an argument and the arrow goes from functor to argument).
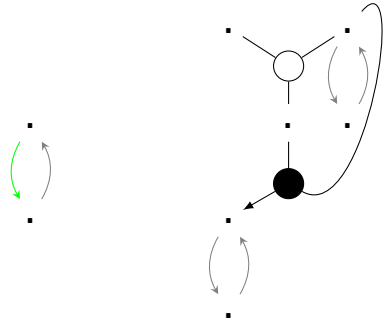
\begin{center}
\begin{tikzpicture}[scale=0.75]
%\node [pn] at (0em,-3.3em) {$\Delta$};
%\node [pn] at (0em,12.4em) {$\Gamma$};
%\node (labl) at (3em,-2.5em) {$[\ldr I]$};
\node (ab) at (3em,4.8em) {$\apsnodei$};
\node (a) at (0,9.6em) {$\apsnodei$};
\node (b) at (6em,9.6em) {$\apsnodei$};
\node[tns] (c) at (3em,7.668em) {};
%\node (clab) at (3em,7.668em) {$i$};
\draw (c) -- (ab);
\draw (c) -- (a);
\draw (c) -- (b);
\node (pa) at (0,0) {$\apsnodei$};
\node[par] (pc) at (3em,1.732em) {};
%\node (pclab) at (3em,1.732em) {\textcolor{white}{$i$}};
\draw (pc) -- (ab);
\path[>=latex,->]  (pc) edge (pa);
\draw (b) to [out=50,in=330] (pc);
%%%%
%\node at (10em,4.8em) {$\rightarrow$};
%%%
\node (x) at (-10em,4.8em) {$\apsnodei$};
%%%% lambda nodes
\node (y) at (-10em,0em) {$\apsnodei$};
\path[->,>=stealth,color=green] (x) edge[bend right] node [right] {} (y);
\path[->,>=stealth,opacity=0.5] (y) edge[bend right] node [right] {} (x);
%\node [pn] at (14em,1.8em) {$\Delta$};
%\node [pn] at (14em,7.7em) {$\Gamma$};
\node (qa) at (0em,-4.8em) {$\apsnodei$};
\path[->,>=stealth,opacity=0.5] (pa) edge[bend right] node [right] {} (qa);
\path[->,>=stealth,opacity=0.5] (qa) edge[bend right] node [right] {} (pa);
%%%
\node (z) at (6em,4.8em) {$\apsnodei$};
\path[->,>=stealth,opacity=0.5] (z) edge[bend right] node [right] {} (b);
\path[->,>=stealth,opacity=0.5] (b) edge[bend right] node [right] {} (z);
%
%\path[->,>=stealth,opacity=0.5] (b) edge[bend right] node [right] {} (qa);
%\path[->,>=stealth,opacity=0.5] (ab) edge[bend right] node [right] {} (qa);
%\path[->,>=stealth,opacity=0.5] (a) edge[bend right] node [right] {} (qa);
%
%\path[->,>=stealth,opacity=0.5] (qa) edge[bend right] node [right] {} (b);
%\path[->,>=stealth,opacity=0.5] (qa) edge[bend right] node [right] {} (a);
%\path[->,>=stealth,opacity=0.5] (qa) edge[bend right] node [right] {} (ab);
\end{tikzpicture}
\end{center}
On the right of the figure, we see the result after $f$ has been transformed into $\lambda x. (f\, x)$. We can now perform abstraction at the root vertex and at the newly created $x$ node. From the root node, all its descendants (the four vertices corresponding to $f$, $x$, $f\, x$ and $\lambda x. (f\, x)$) are potential sisters to a newly introduced variable, but only the root node has been shown to avoid cluttering the figure. From $x$, given that it's a leaf node, there are only two possibilities, corresponding respectively to $\lambda x. (f\, \lambda y. (y\, x))$ and $\lambda x. (f\, \lambda y. (x\, y))$.

Using this trick, all possible parsing actions are directed edges. The task for the neural network is then to assign weights to all edges, maximising the weights assigned to edges representing correct actions, and minimising the weights assigned to incorrect ones.

From the starting situation in the `isolated vertices' scenario, all connections are possible an we should maximise those shown in green below, which indicate that `Aux' is functor of `fils...' and `apporte' of `le salut...'. The connections for the abstractions are not shown in the figure, but should be represented as possiblities.  
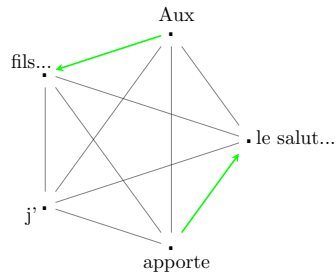
\begin{center}
\scalebox{0.75}{
\begin{tikzpicture}
\tikzstyle{EdgeStyle}=[->,>=stealth,thick,bend left=20]
\foreach \phi in {1,...,5}{
    \node (v\phi) at (360/5 * \phi:2cm) {$\apsnodei$};
};
\draw[->,>=stealth,thick,green] (v1) -- (v2);
\draw[->,>=stealth,thick,green] (v4) -- (v5);
\draw[opacity=0.5] (v1) -- (v3);
\draw[opacity=0.5] (v1) -- (v4);
\draw[opacity=0.5] (v1) -- (v5);
\draw[opacity=0.5] (v2) -- (v3);
\draw[opacity=0.5] (v2) -- (v4);
\draw[opacity=0.5] (v2) -- (v5);
\draw[opacity=0.5] (v3) -- (v4);
\draw[opacity=0.5] (v3) -- (v5);
%\draw (v4) -- (v5);
\node (a) at (360/5 * 1:2.3cm) {\grow{Aux}};
\node (f) at (360/5 * 2:2.3cm) {\grow{fils...}};
\node (j) at (360/5 * 3:2.3cm) {\grow{j'}};
\node (ap) at (360/5 * 4:2.3cm) {\grow{apporte}};
\node (s) at (360/5 * 5:2.3cm) {\grow{\quad\qquad le salut...}};
\end{tikzpicture}}
\end{center}

In the line scenario, it would look as follows. The abstraction vertices and connections are again not shown.
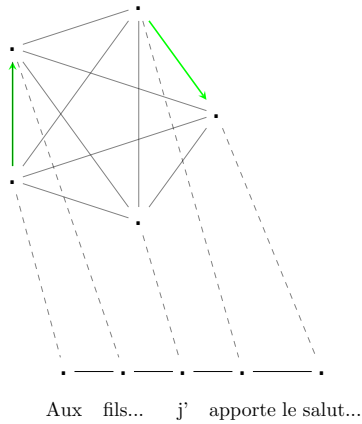
\begin{center}
\scalebox{0.75}{
\begin{tikzpicture}
\tikzstyle{EdgeStyle}=[->,>=stealth,thick,bend left=20]
\begin{scope}[shift={(-5em,-14em)}]
%\node (x1) at (3em,-2em) {$\apsnodei$};
%\node (x2) at (6em,-2em) {$\apsnodei$};
%\node (x3) at (9em,-2em) {$\apsnodei$};
%\node (x4) at (12em,-2em) {$\apsnodei$};
%\node (x5) at (16em,-2em) {$\apsnodei$};
\node (w1) at (3em,1em) {$\apsnodei$};
\node (w2) at (6em,1em) {$\apsnodei$};
\node (w3) at (9em,1em) {$\apsnodei$};
\node (w4) at (12em,1em) {$\apsnodei$};
\node (w5) at (16em,1em) {$\apsnodei$};
\node (a) at (3em,-1em) {\grow{Aux}};
\node (f) at (6em,-1em) {\grow{fils...}};
\node (j) at (9em,-1em) {\grow{j'}};
\node (ap) at (12em,-1em) {\grow{apporte}};
\node (s) at (16em,-1em) {\grow{le salut...}};
\end{scope}
%\draw (w1) -- (x1);
%\draw (w2) -- (x2);
%\draw (w3) -- (x3);
%\draw (w4) -- (x4);
%\draw (w5) -- (x5);
%
\draw (w1) -- (w2);
\draw (w2) -- (w3);
\draw (w3) -- (w4);
\draw (w4) -- (w5);
\foreach \phi in {1,...,5}{
    \node (v\phi) at (360/5 * \phi:2cm) {$\apsnodei$};
};
\draw[->,>=stealth,thick,green] (v3) -- (v2);
\draw[->,>=stealth,thick,green] (v1) -- (v5);
\draw[opacity=0.5] (v1) -- (v2);
\draw[opacity=0.5] (v1) -- (v3);
\draw[opacity=0.5] (v1) -- (v4);
\draw[opacity=0.5] (v2) -- (v3);
\draw[opacity=0.5] (v2) -- (v4);
\draw[opacity=0.5] (v2) -- (v5);
\draw[opacity=0.5] (v3) -- (v4);
\draw[opacity=0.5] (v3) -- (v5);
\draw[opacity=0.5] (v4) -- (v5);
%\draw (v4) -- (v5);
%%%
\draw[opacity=0.5,dashed] (v3) -- (w1); % OK
\draw[opacity=0.5,dashed] (v2) -- (w2); % OK
\draw[opacity=0.5,dashed] (v4) -- (w3);
\draw[opacity=0.5,dashed] (v1) -- (w4);
\draw[opacity=0.5,dashed] (v5) -- (w5);
\end{tikzpicture}}
\end{center}

And finally in the fully connected scenario (again without the abstraction vertices/connections), we would have two isomorphic cliques, connected two each other at their matching points, the first clique (shown at the bottom) representing the fully connected input graph, and the second one the possible parser actions.

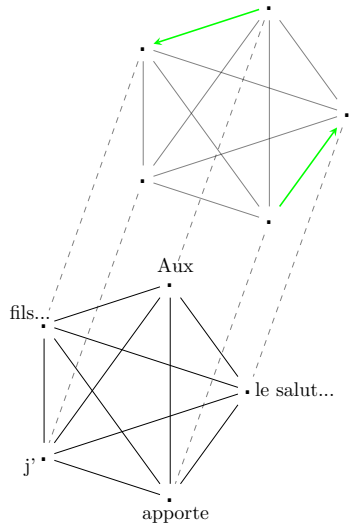
\begin{center}
\scalebox{0.75}{
\begin{tikzpicture}
\tikzstyle{EdgeStyle}=[->,>=stealth,thick,bend left=20]
\begin{scope}[shift={(-5em,-14em)}]
\foreach \phi in {1,...,5}{
    \node (w\phi) at (360/5 * \phi:2cm) {$\apsnodei$};
};
\node (a) at (360/5 * 1:2.3cm) {\grow{Aux}};
\node (f) at (360/5 * 2:2.3cm) {\grow{fils...}};
\node (j) at (360/5 * 3:2.3cm) {\grow{j'}};
\node (ap) at (360/5 * 4:2.3cm) {\grow{apporte}};
\node (s) at (360/5 * 5:2.3cm) {\grow{\quad\qquad le salut...}};
\end{scope}
\draw (w1) -- (w2);
\draw (w1) -- (w3);
\draw (w1) -- (w4);
\draw (w1) -- (w5);
\draw (w2) -- (w3);
\draw (w2) -- (w4);
\draw (w2) -- (w5);
\draw (w3) -- (w4);
\draw (w3) -- (w5);
\draw (w4) -- (w5);
\foreach \phi in {1,...,5}{
    \node (v\phi) at (360/5 * \phi:2cm) {$\apsnodei$};
};
\draw[->,>=stealth,thick,green] (v1) -- (v2);
\draw[->,>=stealth,thick,green] (v4) -- (v5);
\draw[opacity=0.5] (v1) -- (v3);
\draw[opacity=0.5] (v1) -- (v4);
\draw[opacity=0.5] (v1) -- (v5);
\draw[opacity=0.5] (v2) -- (v3);
\draw[opacity=0.5] (v2) -- (v4);
\draw[opacity=0.5] (v2) -- (v5);
\draw[opacity=0.5] (v3) -- (v4);
\draw[opacity=0.5] (v3) -- (v5);
%\draw (v4) -- (v5);
%%%
\draw[opacity=0.5,dashed] (v1) -- (w1);
\draw[opacity=0.5,dashed] (v2) -- (w2);
\draw[opacity=0.5,dashed] (v3) -- (w3);
\draw[opacity=0.5,dashed] (v4) -- (w4);
\draw[opacity=0.5,dashed] (v5) -- (w5);
\end{tikzpicture}}
\end{center}

After we have performed some composition operations, we obtain the configuration shown below. Given that there are only two disconnected substructures, we can still perform a last composition operation with either structure as the functor (in this case, the structure ``Aux fils...'' should be the functor). Alternatively, we can perform a number of abstractions. We have only shown the possibilities from the root node, using a self-loop to help distinguish the computed structure from the parser actions.

There are two possible compose actions, and 16 possible expansion actions: two for each node at both substructures. To keep the figure simple, only a few edges have been shown. The edges which move towards the correct structure are shown in green.
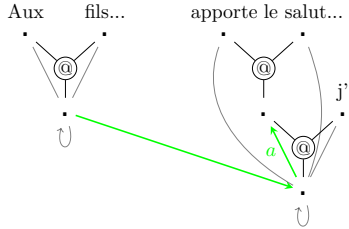
\begin{center}
\scalebox{0.75}{
\begin{tikzpicture}
\node (auxl) at (2em,23.0em) {Aux\proofspace};
\node (aux) at (2em,22em) {$\apsnodei$};
\node (na) at (6em,22em) {$\apsnodei$};
\node [tns] (t) at (4em,20.2em) {};
\node at (4em,20.2em) {\appl};
\node (sspp) at (4em,18em) {$\apsnodei$};
\draw (t)--(aux);
\draw (t)--(sspp);
\draw (t)--(na);
%%%%
\node (filsl) at (6em,23.0em) {fils...\proofspace};
%%%
\node at (18.0em,19.0em) {j'\proofspace};
%%%
\node at (12em,23.0em) {apporte\proofspace};
\node (apporte) at (12em,22em) {$\apsnodei$};
\node (apob) at (16em,22em) {$\apsnodei$};
\node (tvp) at (14em,18em) {$\apsnodei$};
\node (appp) at (18em,18em) {$\apsnodei$};
\node (vp) at (16em,14em) {$\apsnodei$};
% = 18,10.  -24,-12
%
\node [tns] (t1) at (14em,20.2em) {};
\node at (14em,20.2em) {\appl};
\draw (t1) -- (apporte);
\draw (t1) -- (apob);
\draw (t1) -- (tvp);
\node [tns] (t2) at (16em,16.2em) {};
\node at (16em,16.2em) {\appl};
\draw (t2) -- (vp);
\draw (t2) -- (tvp);
\draw (t2) -- (appp);
\node at (16em,23.0em) {le salut...\proofspace};
%%%
\draw[->,>=stealth,thick,green] (sspp) -- (vp);
\draw[->,>=stealth,thick,green] (vp) -- (tvp) node [midway,left] {$a$};
\draw[opacity=0.5] (vp) -- (appp);
\draw[opacity=0.5] (vp) to [in=290,out=70] (apob);
\draw[opacity=0.5] (vp) to [in=250,out=150] (apporte);
\draw[opacity=0.5] (sspp) -- (aux);
\draw[opacity=0.5] (sspp) -- (na);
\draw[opacity=0.5] (sspp) edge [loop below] (sspp);
\draw[opacity=0.5] (vp) edge [loop below] (vp);
\end{tikzpicture}}
\end{center}

\editout{
\begin{center}
\scalebox{0.75}{
\begin{tikzpicture}
\node (auxl) at (2em,23.0em) {Aux\proofspace};
\node (aux) at (2em,22em) {$\apsnodei$};
\node (na) at (6em,22em) {$\apsnodei$};
\node [tns] (t) at (4em,20.2em) {};
\node at (4em,20.2em) {\appl};
\node (sspp) at (4em,18em) {$\apsnodei$};
\draw (t)--(aux);
\draw (t)--(sspp);
\draw (t)--(na);
%%%%
\node (filsl) at (6em,23.0em) {fils...\proofspace};
%%%
\node at (18.0em,19.0em) {j'\proofspace};
%%%
\node at (12em,23.0em) {apporte\proofspace};
\node (apporte) at (12em,22em) {$\apsnodei$};
\node (apob) at (16em,22em) {$\apsnodei$};
\node (tvp) at (14em,18em) {$\apsnodei$};
\node (appp) at (18em,18em) {$\apsnodei$};
\node (vp) at (16em,14em) {$\apsnodei$};
% = 18,10.  -24,-12
%
\node [tns] (t1) at (14em,20.2em) {};
\node at (14em,20.2em) {\appl};
\draw (t1) -- (apporte);
\draw (t1) -- (apob);
\draw (t1) -- (tvp);
\node [tns] (t2) at (16em,16.2em) {};
\node at (16em,16.2em) {\appl};
\draw (t2) -- (vp);
\draw (t2) -- (tvp);
\draw (t2) -- (appp);
\node at (16em,23.0em) {le salut...\proofspace};
%%%
\draw[->,>=stealth,thick,green] (sspp) -- (vp);
\draw[->,>=stealth,thick,green] (vp) -- (tvp) node [midway,left] {$a$};
\draw[opacity=0.5] (vp) -- (appp);
\draw[opacity=0.5] (vp) to [in=290,out=70] (apob);
\draw[opacity=0.5] (vp) to [in=250,out=150] (apporte);
\draw[opacity=0.5] (sspp) -- (aux);
\draw[opacity=0.5] (sspp) -- (na);
\draw[opacity=0.5] (sspp) edge [loop below] (sspp);
\draw[opacity=0.5] (vp) edge [loop below] (vp);
\end{tikzpicture}}
\end{center}
}

This gives the following global strategy for neural proof net generation.
\begin{enumerate}
	\item start with an initial configuration,
	\item\label{it:prop:edge} propagate information using message passing,
	\item predict directed edges,
	\item update the graph with the graph operation corresponding to the best edge (or all edges over a certain weight),
	\item update the action edges: delete action edges which are no longer possible in the updated graph, and add actions which have become possible,
	\item continue to~\ref{it:prop:edge} with the updated graph, or stop if no edges are assigned weights above a threshold.
\end{enumerate}

We can also add `stop' as a global graph property to predict separately, but only when the graph is already connected (and therefore sure to correspond to an appropriate lambda term). This has the advantage that the neural network can learn to calibrate the stop condition against the edge prediction weights.

During training, we take different stages of the graph generation, and in each case, we maximise the weights of the correct edges and minimise the incorrect ones (using values between 0 and 1).

This global strategy still has quite a few blanks to fill in, but we can try to adapt some of the known edge prediction architectures to the current context  \cite[Chapters~10, 19, 24]{wu2022graph}. However, we should be aware of a potential problem, and that is that for some of the main use cases of edge prediction in neural networks, we have a single, large graph and want to predict missing edges and their weights/labels. This is the case for recommender systems and knowledge graphs. By comparison, our graphs will be quite small, but we will have many of them.
% TODO add citations here

\subsubsection*{Parser actions as vertex labelling}

As an alternative to generating parser actions as edge predictions, we can represent parser actions by vertices. There are application vertices labeled `$@$' and connected by an $f$ and $a$ edge to two different vertices in the term representation. Similarly there are vertices labeled `$\lambda_f$' and `$\lambda_a$' for abstraction over functor and arguments respectively, connected to two (not necessarily different) vertices. 

This would represent the initial graph --- for the case with isolated vertices --- and possible parser actions as shown below. Only a few of the possible actions have been shown. The correct actions have been shaded grey.
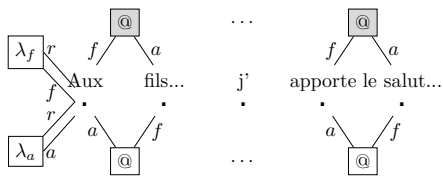
\begin{center}
\scalebox{0.75}{
\begin{tikzpicture}
\node (auxl) at (-6em,23.0em) {Aux\proofspace};
\node (aux) at (-6em,22em) {$\apsnodei$};
\node (na) at (-2em,22em) {$\apsnodei$};
%%%%
\node (filsl) at (-2em,23.0em) {fils...\proofspace};
%%%
\node at (2.0em,23.0em) {j'\proofspace};
\node at (2.0em,22em) {$\apsnodei$};
%%%
\node (apportel) at (6em,23.0em) {apporte\proofspace};
\node (apporte) at (6em,22em) {$\apsnodei$};
\node (apob) at (10em,22em) {$\apsnodei$};
\node (apobl) at (10em,23.0em) {le salut...\proofspace};
%%%
\node[draw,fill=gray!30] (a1) at (-4em,26em) {$@$};
\draw (auxl) -- (a1) node [midway,left] {$f$}; 
\draw (filsl) -- (a1) node [midway,right] {$a$}; 
%%%
\node[draw] (a2) at (-4em,19em) {$@$};
\draw (aux) -- (a2) node [midway,left] {$a$}; 
\draw (na) -- (a2) node [midway,right] {$f$};
%%%
%%%
\node[draw,fill=gray!30] (a3) at (8em,26em) {$@$};
\draw (apportel) -- (a3) node [midway,left] {$f$}; 
\draw (apobl) -- (a3) node [midway,right] {$a$}; 
%%%
\node[draw] (a4) at (8em,19em) {$@$};
\draw (apporte) -- (a4) node [midway,left] {$a$}; 
\draw (apob) -- (a4) node [midway,right] {$f$};
%%%
\node[draw] (l1) at (-9em,19.5em) {$\lambda_a$};
\draw (aux.west) -- (l1.north east) node [near end,above] {$r$}; 
\draw (aux.south west) -- (l1.east) node [near end,below] {$a$}; 
%%%
\node[draw] (l2) at (-9em,24.5em) {$\lambda_f$};
\draw (aux.north west) -- (l2.east) node [near end,above] {$r$}; 
\draw (aux.west) -- (l2.south east) node [near end,below] {$f$}; 
%%%%
\node at (2em,19em) {$\cdots$};
\node at (2em,26em) {$\cdots$};
\end{tikzpicture}}
\end{center}
 
Like before, we want the graph neural network to take an input graph and assign weights between 0 and 1 to all possible actions, maximising the weights assigned to the correct actions and minimising the weights assigned to the incorrect ones. The actions include the `stop' action when the graph is connected.
 
 The structure of the neural proof net generation algorithm does not change much with respect to the edge prediction strategy.
\begin{enumerate}
	\item start with an initial configuration,
	\item\label{it:prop:vertex} propagate information using message passing,
	\item predict weights for all action vertices,
	\item update the graph with the graph operation corresponding to the best action vertex,
	\item continue to~\ref{it:prop:vertex} with the updated graph, until the `stop' action is selected.
\end{enumerate}

 There are graph neural network architectures available in standard libraries which seem well-suited for our node-labeled, vertex-labeled graphs \cite{xie2018crystal,maguire2021xenet}.

 \editout{
 \begin{center}
\scalebox{0.75}{
\begin{tikzpicture}
\node (auxl) at (2em,23.0em) {Aux\proofspace};
\node (aux) at (2em,22em) {$\apsnodei$};
\node (na) at (6em,22em) {$\apsnodei$};
\node [tns] (t) at (4em,20.2em) {};
\node at (4em,20.2em) {\appl};
\node (sspp) at (4em,18em) {$\apsnodei$};
\draw (t)--(aux);
\draw (t)--(sspp);
\draw (t)--(na);
%%%%
\node (filsl) at (6em,23.0em) {fils...\proofspace};
%%%
\node at (18.0em,19.0em) {j'\proofspace};
%%%
\node at (12em,23.0em) {apporte\proofspace};
\node (apporte) at (12em,22em) {$\apsnodei$};
\node (apob) at (16em,22em) {$\apsnodei$};
\node (tvp) at (14em,18em) {$\apsnodei$};
\node (appp) at (18em,18em) {$\apsnodei$};
\node (vp) at (16em,14em) {$\apsnodei$};
% = 18,10.  -24,-12
%
\node [tns] (t1) at (14em,20.2em) {};
\node at (14em,20.2em) {\appl};
\draw (t1) -- (apporte);
\draw (t1) -- (apob);
\draw (t1) -- (tvp);
\node [tns] (t2) at (16em,16.2em) {};
\node at (16em,16.2em) {\appl};
\draw (t2) -- (vp);
\draw (t2) -- (tvp);
\draw (t2) -- (appp);
\node at (16em,23.0em) {le salut...\proofspace};
%%%
\end{tikzpicture}}
\end{center}
}

\subsubsection*{Comments, alternatives, evaluation} 

Both of the previous options to generate the graph structure iterate between an action prediction stage and a graph update stage. The graph update stage has quite a bit of overhead: actions which are no longer possible need to be removed (generally, this is because a vertex is no longer a root node), new actions need to be added: new applications for new root nodes, new abstractions for all new vertices.

To transform a correct sequence of parser action into the input for our training stage, we need to cut the sequences into their individual steps, indicating \emph{all} correct actions in each case. Each step has only the BERT embeddings of the leaves, and the graph labels as its input. However, it would seem advantageous cast the task as a sequence prediction task, where all weights of the previous step are available to be updated. Treating proof net generation as a type of neural network for graph transformation would be a promising possibility \cite[Chapter~12]{wu2022graph}. However, there doesn't appear to be an easy fit between our problem and existing architectures for this case. For example, \citeasnoun{wang2018neural} use a sequence-to-graph neural network for parsing semantic dependencies, but this approach is not easily adapted to our task since it is essentially an edge prediction task between vertices corresponding to words. This means it operates modulo associativity (not distinguishing between $(np\backslash s)/np$ and $np\backslash (s/np)$ for example) and that there is no easy way to incorporate the expansion rules. The parsing model which corresponds best to our needs is the sequence-to-graph model of \citeasnoun{chen2018sequence}, who use a graph generation network to translate sequences of text to parser actions generating a semantic graph. However, even this model would need to be changed to ensure only proof \emph{nets} are generated, otherwise we would lose one of the major attractive points of our current approach to neural proof nets.

Whatever the strategy chosen for graph generation, we want to compare the results with the current state-of-the-art for the `standard' strategy of neural proof nets \cite{deepgrail23}. This means evaluating it on the no-errors in the linear logic proof criterion, or, equivalently the correct (untyped) lambda term criterion.

\subsection{Vertex labelling to obtain the surface structure graph}

For the second task, we have already produced the graph representing the meaning, and we want to obtain the more detailed graph representing the syntactic structure. 

\begin{figure}
\begin{center}
\scalebox{0.75}{
\begin{tikzpicture}
\node (auxl) at (10em,11.0em) {Aux\proofspace};
\node (aux) at (10em,10em) {$\apsnodei$};
\node (na) at (14em,10em) {$\apsnodei$};
\node [tns] (t) at (12em,8.2em) {};
\node at (12em,8.2em) {\appl};
\node (sspp) at (12em,6em) {$\apsnodei$};
\node (auxs) at (10em,10em) {\blb};
\draw[>=latex,->] (t)--(aux);
\draw (t)--(sspp);
\draw (t)--(na);
\node (spp) at (16em,6em) {$\apsnodei$};
\node (auxs) at (14em,2em) {$\apsnodei$};
\node [tns] (tt) at (14em,4.0em) {};
\node at (14em,4em) {\appl};
\draw (tt) -- (auxs);
\draw[>=latex,->] (tt) -- (sspp);
\draw (tt) -- (spp);
\node [par] (par) at (18em,7.8em) {};
\node at (18em,7.8em) {$\textcolor{white}{\small \lambda}$};
%\node (pp) at (20em,6em) {$\atppr{}$};
%\node (ppd) at (20em,6em) {\phantom{x}};
\node (tops) at (18em,10em) {$\apsnodei$};
%\draw (par)--(ppd);
\draw (par)--(tops);
\node (sppd) at (16em,6em) {\phantom{x}};
\path[>=latex,->]  (par) edge (sppd);
%%%%
%\node (fils) at (21em,10em) {$\atnr$};
\node (filsl) at (14em,11.0em) {fils...\proofspace};
%%%
\node (clamel) at (02em,7.0em) {clame\proofspace};
%\node (hs) at (20em,2em) {$\atsr$};
\node (ss) at (04em,2em) {$\apsnodei$};
\node (ssd) at (4em,2em) {\phantom{x}};
\node (cs) at (9em,-2em) {$\apsnodei$};
\node [tns] (t) at (9em,0.2em){};
\node at (9em,0.2em){\appl};
\draw (t) -- (auxs);
\draw[>=latex,->] (t) -- (ssd);
\draw (t) -- (cs);
\node (clsu) at (06em,6em) {$\apsnodei$};
\node (clsud) at (06em,6em) {$\apsnodei$};
\node (clame) at (02em,6em) {$\apsnodei$};
\node (clamed) at (02em,6em) {\phantom{M}};
\node [tns] (tc) at (04em,4.2em) {};
\node at (4em,4.2em) {\appl};
\draw[>=latex,->] (tc) -- (clame);
\draw (tc) -- (clsud);
\draw (tc) -- (ss);
%%%
%\node (il) at (33em,10em) {$\atnpr$};
\node at (06em,7.0em) {-t-il\proofspace};
%%%
%\node (je) at (36em,10em) {$\atnpr$};
\node at (19.8em,15.0em) {j'\proofspace};
%%%
\node at (12em,23.0em) {apporte\proofspace};
\node (apporte) at (12em,22em) {$\apsnodei$};
\node (apob) at (16em,22em) {$\apsnodei$};
\node (tvp) at (14em,18em) {$\apsnodei$};
\node (appp) at (18em,18em) {$\apsnodei$};
\node (vp) at (16em,14em) {$\apsnodei$};
\node (apsu) at (20em,14em) {$\apsnodei$};
\node (apc) at (18em,10em) {\phantom{x}};
% = 18,10.  -24,-12
%
\node [tns] (t1) at (14em,20.2em) {};
\node at (14em,20.2em) {\appl};
\draw[>=latex,->] (t1) -- (apporte);
\draw (t1) -- (apob);
\draw (t1) -- (tvp);
\node [tns] (t2) at (16em,16.2em) {};
\node at (16em,16.2em) {\appl};
\draw (t2) -- (vp);
\draw[>=latex,->] (t2) -- (tvp);
\draw (t2) -- (appp);
\node [tns] (t3) at (18em,12em) {};
\node at (18em,12em) {\appl};
\draw (t3) -- (apc);
\draw[>=latex,->] (t3) -- (vp);
\draw (t3) -- (apsu);
\node at (16em,23.0em) {le salut...\proofspace};
\draw (appp) to [out=50,in=330] (par);
%\node (lesalut) at (49em,10em) {$\atnpr$};
\end{tikzpicture}}
\raisebox{7em}{$\;\;\;\;\;\leadsto$}
\scalebox{0.75}{
\begin{tikzpicture}
\node (auxl) at (10em,11.0em) {Aux\proofspace};
\node (aux) at (10em,10em) {$\apsnodei$};
\node (na) at (14em,10em) {$\apsnodei$};
\node [tns] (t) at (12em,8.2em) {};
\node at (12em,8.2em) {$/$};
\node (sspp) at (12em,6em) {$\apsnodei$};
\node (auxs) at (10em,10em) {\blb};
\draw (t)--(auxs);
\draw (t)--(sspp);
\draw (t)--(na);
\node (spp) at (16em,6em) {$\apsnodei$};
\node (auxs) at (14em,2em) {$\apsnodei$};
\node [tns] (tt) at (14em,4.0em) {};
\node at (14em,4em) {$/$};
\draw (tt) -- (auxs);
\draw (tt) -- (sspp);
\draw (tt) -- (spp);
\node [par] (par) at (18em,7.8em) {};
\node at (18em,7.8em) {\textcolor{white}{\small /}};
%\node (pp) at (20em,6em) {$\atppr{}$};
%\node (ppd) at (20em,6em) {\phantom{x}};
\node (tops) at (18em,10em) {$\apsnodei$};
%\draw (par)--(ppd);
\draw (par)--(tops);
\node (sppd) at (16em,6em) {\phantom{x}};
\path[>=latex,->]  (par) edge (sppd);
%%%%
%\node (fils) at (21em,10em) {$\atnr$};
\node (filsl) at (14em,11.0em) {fils...\proofspace};
%%%
\node (clamel) at (22em,7.0em) {clame\proofspace};
%\node (hs) at (20em,2em) {$\atsr$};
\node (ss) at (24em,2em) {$\apsnodei$};
\node (ssd) at (24em,2em) {\phantom{x}};
\node (cs) at (19em,-2em) {$\atsr$};
\node [tns] (t) at (19em,0.2em){};
\node at (19em,0.2em) {$\backslash$};
\draw (t) -- (auxs);
\draw (t) -- (ssd);
\draw (t) -- (cs);
\node (clsu) at (26em,6em) {$\apsnodei$};
\node (clsud) at (26em,6em) {$\apsnodei$};
\node (clame) at (22em,6em) {$\apsnodei$};
\node (clamed) at (22em,6em) {\phantom{M}};
\node [tns] (tc) at (24em,4.2em) {};
\node at (24em,4.2em) {$/$};
\draw (tc) -- (clamed);
\draw (tc) -- (clsud);
\draw (tc) -- (ss);
%%%
%\node (il) at (33em,10em) {$\atnpr$};
\node at (26em,7.0em) {-t-il\proofspace};
%%%
%\node (je) at (36em,10em) {$\atnpr$};
\node at (16em,15.0em) {j'\proofspace};
%%%
\node at (16em,23.0em) {apporte\proofspace};
\node (apporte) at (16em,22em) {$\apsnodei$};
\node (apob) at (20em,22em) {$\apsnodei$};
\node (tvp) at (18em,18em) {$\apsnodei$};
\node (appp) at (22em,18em) {$\apsnodei$};
\node (vp) at (20em,14em) {$\apsnodei$};
\node (apsu) at (16em,14em) {$\apsnodei$};
\node (apc) at (18em,10em) {\phantom{x}};
% = 18,10.  -24,-12
%
\node [tns] (t1) at (18em,20em) {};
\node at (18em,20em) {$/$};
\draw (t1) -- (apporte);
\draw (t1) -- (apob);
\draw (t1) -- (tvp);
\node [tns] (t2) at (20em,16.0em) {};
\node at (20em,16em) {$/$};
\draw (t2) -- (vp);
\draw (t2) -- (tvp);
\draw (t2) -- (appp);
\node [tns] (t3) at (18em,12em) {};
\node at (18em,12em) {$\backslash$};
\draw (t3) -- (apc);
\draw (t3) -- (vp);
\draw (t3) -- (apsu);
\node at (20em,23.0em) {le salut...\proofspace};
\draw (appp) to [out=50,in=330] (par);
%\node (lesalut) at (49em,10em) {$\atnpr$};
\end{tikzpicture}}
\end{center}
\caption{Figure~\ref{fig:sem} used for vertex labeling}
\label{fig:semlabel}
\end{figure}

Figure~\ref{fig:semlabel} repeats the structure of Figure~\ref{fig:sem}, but with additional arrows pointing to the functor of the application nodes. These arrows are not formally a part of the structure, but they make it easier to see some of its properties. The arrows always point to the main formula of their link, meaning that the arrow is always a formula $A\multimap B$, with $A$ its sister and $B$ its mother node (for tensor links the mother is displayed below the central node, for par links above it).

With this in mind, we can simply compute the generic structure of the formulas which ensures it respects the restrictions on the formulas imposed by the links. This is morally equivalent to computing a principal type in type theory: we make the minimal assumptions about the formulas which guarantees the proof is correct. 
\begin{align*}
\textit{Aux} &= A \multimap (E\multimap F) \multimap G \\
\textit{fils...} &= A \\	
\textit{clame} & = B \multimap G \multimap H \\
\textit{-t-il} & = B \\
\textit{j'} & = C \\
\textit{apporte} & = D\multimap E \multimap C \multimap F \\
\textit{le salut...} &= D \\
\end{align*}
The conclusion is assigned the formula $H$. Since each formula occurs exactly twice, we can uniquely recover the proof net from these formulas, as there is only one way to connect the positive and negative atoms.

To transform these minimal lexical assignments to the fully detailed ones we seek, we need to do two things:
\begin{enumerate}
	\item\label{task:atom} replace the type variables by atomic formulas\footnote{We assume all proofs are in long normal form. Without this restriction, it would be possible to instantiate a variable by a complex formula. For long normal form proofs, we can restrict ourselves to replacing variables by atomic formulas.},
	\item\label{task:direction} replace the linear logic connectives by their directional variants.
\end{enumerate}
We can see task~\ref{task:direction} as an \emph{edge} labelling problem, as shown in Figure~\ref{fig:semlabel} on the right. Each central node now uses the labels `$/$' and `$\backslash$' instead of `$@$' and `$\lambda$'.  

However, it is easier to treat both tasks as vertex labelling problems, allowing us to train them in parallel. For task~\ref{task:atom}, we want to label all vertices where no arrow arrives with one of the fixed set of atomic formulas in our grammar.  For task~\ref{task:direction}, we want to label all vertices which are not leaves of the structure with one of the fixed set of binary connectives. 

From the structure of the input graph, we can determine exactly which vertices need an atom label and which vertices need a connective label. This means that the graph convolutions can propagate the information through vertices which will not be labelled in any way which optimises the information available at the vertices which \emph{will} be labelled.

\begin{figure}
\begin{center}
\scalebox{0.75}{
\begin{tikzpicture}
\node (auxl) at (10em,11.0em) {Aux\proofspace};
\node (aux) at (10em,10em) {$\apsnodei$};
\node (na) at (14em,10em) {$n$};
\node [tns] (t) at (12em,8.2em) {};
\node at (12em,8.2em) {\appl};
\node (sspp) at (12em,6em) {$\apsnodei$};
\node (auxs) at (10em,10em) {\blb};
\draw[>=latex,->] (t)--(aux);
\draw (t)--(sspp);
\draw (t)--(na);
\node (spp) at (16em,6em) {$\apsnodei$};
\node (auxs) at (14em,2em) {$s$};
\node [tns] (tt) at (14em,4.0em) {};
\node at (14em,4em) {\appl};
\draw (tt) -- (auxs);
\draw[>=latex,->] (tt) -- (sspp);
\draw (tt) -- (spp);
\node [par] (par) at (18em,7.8em) {};
\node at (18em,7.8em) {$\textcolor{white}{\small \lambda}$};
%\node (pp) at (20em,6em) {$\atppr{}$};
%\node (ppd) at (20em,6em) {\phantom{x}};
\node (tops) at (18em,10em) {$s$};
%\draw (par)--(ppd);
\draw (par)--(tops);
\node (sppd) at (16em,6em) {\phantom{x}};
\path[>=latex,->]  (par) edge (sppd);
%%%%
%\node (fils) at (21em,10em) {$\atnr$};
\node (filsl) at (14em,11.0em) {fils...\proofspace};
%%%
\node (clamel) at (02em,7.0em) {clame\proofspace};
%\node (hs) at (20em,2em) {$\atsr$};
\node (ss) at (04em,2em) {$\apsnodei$};
\node (ssd) at (4em,2em) {\phantom{x}};
\node (cs) at (9em,-2em) {$s$};
\node [tns] (t) at (9em,0.2em){};
\node at (9em,0.2em){\appl};
\draw (t) -- (auxs);
\draw[>=latex,->] (t) -- (ssd);
\draw (t) -- (cs);
\node (clsu) at (06em,6em) {$np$};
\node (clsud) at (06em,6em) {\phantom{$\apsnodei$}};
\node (clame) at (02em,6em) {$\apsnodei$};
\node (clamed) at (02em,6em) {\phantom{M}};
\node [tns] (tc) at (04em,4.2em) {};
\node at (4em,4.2em) {\appl};
\draw[>=latex,->] (tc) -- (clame);
\draw (tc) -- (clsud);
\draw (tc) -- (ss);
%%%
%\node (il) at (33em,10em) {$\atnpr$};
\node at (06em,7.0em) {-t-il\proofspace};
%%%
%\node (je) at (36em,10em) {$\atnpr$};
\node at (19.8em,15.0em) {j'\proofspace};
%%%
\node at (12em,23.0em) {apporte\proofspace};
\node (apporte) at (12em,22em) {$\apsnodei$};
\node (apob) at (16em,22em) {$np$};
\node (tvp) at (14em,18em) {$\apsnodei$};
\node (appp) at (18em,18em) {$pp$};
\node (vp) at (16em,14em) {$\apsnodei$};
\node (apsu) at (20em,14em) {$np$};
\node (apc) at (18em,10em) {\phantom{x}};
% = 18,10.  -24,-12
%
\node [tns] (t1) at (14em,20.2em) {};
\node at (14em,20.2em) {\appl};
\draw[>=latex,->] (t1) -- (apporte);
\draw (t1) -- (apob);
\draw (t1) -- (tvp);
\node [tns] (t2) at (16em,16.2em) {};
\node at (16em,16.2em) {\appl};
\draw (t2) -- (vp);
\draw[>=latex,->] (t2) -- (tvp);
\draw (t2) -- (appp);
\node [tns] (t3) at (18em,12em) {};
\node at (18em,12em) {\appl};
\draw (t3) -- (apc);
\draw[>=latex,->] (t3) -- (vp);
\draw (t3) -- (apsu);
\node at (16em,23.0em) {le salut...\proofspace};
\draw (appp) to [out=50,in=330] (par);
%\node (lesalut) at (49em,10em) {$\atnpr$};
\end{tikzpicture}}
%%%%%%%%
\scalebox{0.75}{
\begin{tikzpicture}
\node (auxl) at (10em,11.0em) {Aux\proofspace};
\node (aux) at (10em,10em) {$\apsnodei$};
\node (na) at (14em,10em) {$\apsnodei$};
\node [tns] (t) at (12em,8.2em) {};
\node at (12em,8.2em) {\appl};
\node (sspp) at (12em,6em) {$/$};
\node (auxs) at (10em,10em) {\blb};
\draw[>=latex,->] (t)--(aux);
\draw (t)--(sspp);
\draw (t)--(na);
\node (spp) at (16em,6em) {$/_2$};
\node (auxs) at (14em,2em) {$/$};
\node [tns] (tt) at (14em,4.0em) {};
\node at (14em,4em) {\appl};
\draw (tt) -- (auxs);
\draw[>=latex,->] (tt) -- (sspp);
\draw (tt) -- (spp);
\node [par] (par) at (18em,7.8em) {};
\node at (18em,7.8em) {$\textcolor{white}{\small \lambda}$};
%\node (pp) at (20em,6em) {$\atppr{}$};
%\node (ppd) at (20em,6em) {\phantom{x}};
\node (tops) at (18em,10em) {$\backslash$};
%\draw (par)--(ppd);
\draw (par)--(tops);
\node (sppd) at (16em,6em) {\phantom{x}};
\path[>=latex,->]  (par) edge (sppd);
%%%%
%\node (fils) at (21em,10em) {$\atnr$};
\node (filsl) at (14em,11.0em) {fils...\proofspace};
%%%
\node (clamel) at (02em,7.0em) {clame\proofspace};
%\node (hs) at (20em,2em) {$\atsr$};
\node (ss) at (04em,2em) {$/$};
\node (ssd) at (4em,2em) {\phantom{x}};
\node (cs) at (9em,-2em) {$\backslash_1$};
\node [tns] (t) at (9em,0.2em){};
\node at (9em,0.2em){\appl};
\draw (t) -- (auxs);
\draw[>=latex,->] (t) -- (ssd);
\draw (t) -- (cs);
\node (clsu) at (06em,6em) {$\apsnodei$};
\node (clsud) at (06em,6em) {$\apsnodei$};
\node (clame) at (02em,6em) {$\apsnodei$};
\node (clamed) at (02em,6em) {\phantom{M}};
\node [tns] (tc) at (04em,4.2em) {};
\node at (4em,4.2em) {\appl};
\draw[>=latex,->] (tc) -- (clame);
\draw (tc) -- (clsud);
\draw (tc) -- (ss);
%%%
%\node (il) at (33em,10em) {$\atnpr$};
\node at (06em,7.0em) {-t-il\proofspace};
%%%
%\node (je) at (36em,10em) {$\atnpr$};
\node at (19.8em,15.0em) {j'\proofspace};
%%%
\node at (12em,23.0em) {apporte\proofspace};
\node (apporte) at (12em,22em) {$\apsnodei$};
\node (apob) at (16em,22em) {$\apsnodei$};
\node (tvp) at (14em,18em) {$/$};
\node (appp) at (18em,18em) {$\apsnodei$};
\node (vp) at (16em,14em) {$/$};
\node (apsu) at (20em,14em) {$\apsnodei$};
\node (apc) at (18em,10em) {\phantom{x}};
% = 18,10.  -24,-12
%
\node [tns] (t1) at (14em,20.2em) {};
\node at (14em,20.2em) {\appl};
\draw[>=latex,->] (t1) -- (apporte);
\draw (t1) -- (apob);
\draw (t1) -- (tvp);
\node [tns] (t2) at (16em,16.2em) {};
\node at (16em,16.2em) {\appl};
\draw (t2) -- (vp);
\draw[>=latex,->] (t2) -- (tvp);
\draw (t2) -- (appp);
\node [tns] (t3) at (18em,12em) {};
\node at (18em,12em) {\appl};
\draw (t3) -- (apc);
\draw[>=latex,->] (t3) -- (vp);
\draw (t3) -- (apsu);
\node at (16em,23.0em) {le salut...\proofspace};
\draw (appp) to [out=50,in=330] (par);
%\node (lesalut) at (49em,10em) {$\atnpr$};
\end{tikzpicture}}
\end{center}
\caption{The two vertex labelling tasks: atomic formulas (left) and connectives (right)}
\label{fig:label}
\end{figure}
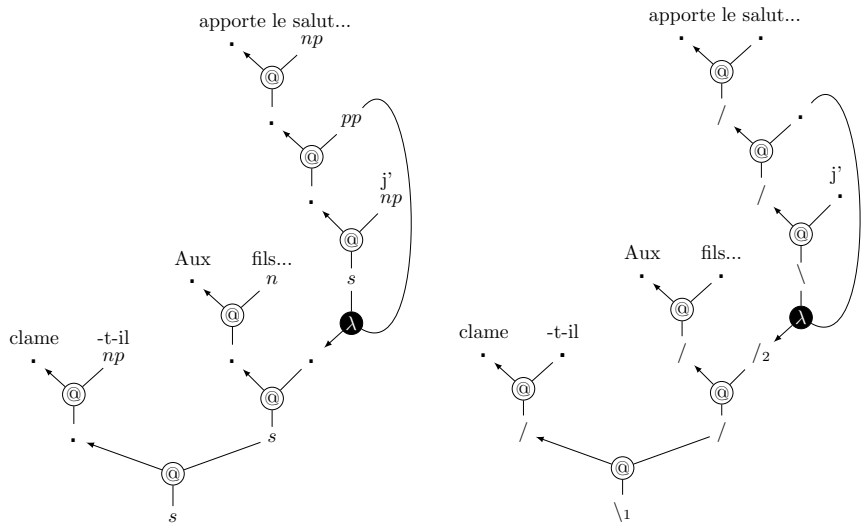

Figure~\ref{fig:label} shows the correct result for the first task on the left of the figure. It amounts to choosing $A=n$, $B=C=D=np$, $E=pp$ and $F=G=H=s$, which produces the following lexical assignments.
\begin{align*}
\textit{Aux} &= n \multimap (pp\multimap s) \multimap s \\
\textit{fils...} &= n \\	
\textit{clame} & = np \multimap s \multimap s \\
\textit{-t-il} & = np \\
\textit{j'} & = np \\
\textit{apporte} & = np\multimap pp \multimap np \multimap F \\
\textit{le salut...} &= np \\
\end{align*}

Figure~\ref{fig:label} shows the correct results for the second task on the right of the figure. It corresponds to replacing linear implication with the directional implications of the Lambek calculus. This task also recovers the labels for `$\backslash_1$' and `$/_2$'. 

\begin{align*}
\textit{Aux} &= (G/(F/_2 E))  /A  \\
\textit{fils...} &= A \\	
\textit{clame} & = (G \backslash_1 H)/B \\
\textit{-t-il} & = B \\
\textit{j'} & = C \\
\textit{apporte} & = ((C\backslash F)/E)/D \\
\textit{le salut...} &= D \\
\end{align*}

Combining these two results together then gives the full formula information.
\begin{align*}
\textit{Aux} &= (s/(s/_2 pp))  /n  \\
\textit{fils...} &= n \\	
\textit{clame} & = (s \backslash_1 s)/np \\
\textit{-t-il} & = np \\
\textit{j'} & = np \\
\textit{apporte} & = ((np\backslash np)/pp)/np \\
\textit{le salut...} &= np \\
\end{align*}

Compared to the graph generation task, the labelling tasks in this section are relatively simple: our vertex labelling needs to decide for the first task that ``fils...'' is a noun $n$ and ``le salut....'' is a noun phrase etc., but also, for the second task, that ``fils...'' is to the immediate right of ``Aux'', and ``le salut'' to the immediate right of ``apporte''.

These seem to be tasks where standard graph architectures for vertex labelling would work well \cite[Chapter~4]{wu2022graph}. 

Generating the labels allows for a direct comparison with the state-of-the-art in supertagging for the same dataset \cite{kogkalidis2022geometry,deepgrail23}.

%\subsection{Discussion}
\section{Conclusion}
\label{sec:conc}

I have sketched a novel way of generating proof nets using graph neural networks. Even though there are a number of implementation details to work out and experiment with, I believe this approach has a number of important advantages over the `standard' way of \citeasnoun{kogkalidis2020neural}.

By changing the split of the task from a formula prediction plus link predication task to a graph generation plus vertex labelling task, we can generate our proof nets in a way which guarantees we always obtain a proof (that is, a proof \emph{net}, not a proof structure) and therefore a lambda term for downstream natural language processing tasks.  

The graph generation task is also easily adapted to a $k$-best or beam search setting: at each prediction step, we can take the $k$ most promising actions (those with highest weight/probability), batch generate the possible next actions for all of them, then prune until we are again left with $k$-best different graphs. This requires some way to ensure different ways of generating the same graph are identified, with only the one with the highest weight kept\footnote{Using some form of canonical vertex numbering (e.g. left-to-right, bottom-to-top), we can avoid having to check for graph isomorphism and simply check for identity.}. 

The parser actions perspective also makes it easier to evaluate partial successes. For example, we can look at two different parse trees for a sentence an compare the two by evaluating how many of the applications and abstractions are correct. 

A weakness of the current proposal is that adding product formulas makes the task quite a bit harder. First of, we need to distinguish between two cases for each tensor link, application and pairing (corresponding to the product introduction rule). Second, the product elimination rule is complicated: we need to expand two vertices (by expanding either one or two vertices) and attach the par link for the product to these two vertices. On the other hand, the product is extremely rare in our dataset \cite{moot15tlgbank} and it appears all occurrences $A\otimes B$ can be replaced by $(A\multimap  (B \multimap s)) \multimap s$, which is both a standard way to implement products as implications in the lambda calculus, and the standard treatment of argument cluster coordination in type-logical grammars.

Given the weakness at treating product types, the skeptic would be right to ask whether we're not simply doing natural deduction proof search, thereby losing some of the most important benefits of using proof nets. I think that in a sense, proof nets --- especially in the version presented here --- \emph{are} natural deduction proofs, but the product elimination rule is tricky in combination with the graph expansion approach advocated here. 

Taken together, I think the benefits largely outweigh the drawbacks. The main problems with standard neural proof nets are that they can fail to produce any meaning at all because of a single supertag error, and that they have problems in the $k$-best scenario because of the task split: it is not hard to obtain the $k$-best supertags but it is hard to obtain the $k$-best proofs.

This new method will of course have to be evaluated against the current state-of-the-art for neural proof nets, in terms of supertagger scores, but, most importantly, in terms of the percentage of sentences receiving the correct meaning.

\bibliographystyle{agsm}
\bibliography{moot.bib}

 \pagebreak
\appendix
\section{Backward chaining proof net generation}
\label{app:backward}

We can do the graph generation procedure as backward chaining proof search as well. While the forward chaining proof search of Section~\ref{sec:gg} is essentially a type of natural deduction proof search, the backward chaining method is more like sequent calculus proof search. At each stage, we have a structure we are constructing about which we know only the hypotheses and the conclusion vertices. This is also essentially the same strategy as the one used by \citeasnoun{kmmt19ded}, with the difference that we do not have the formula information in the current case.

The starting position therefore has exactly one hypothesis for each word in the sentence, and a single conclusion.

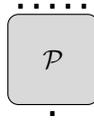
\begin{center}
\begin{tikzpicture}[scale=0.75]
\node[gpn] at (4em,4em) 	{$\mathcal{P}$};
\node at (2.4em,6.8em) {$\apsnodei$};
\node at (3.2em,6.8em) {$\apsnodei$};
\node at (4.0em,6.8em) {$\apsnodei$};
\node at (4.8em,6.8em) {$\apsnodei$};
\node at (5.6em,6.8em) {$\apsnodei$};
%\node at (3.0em,6.2em) {$\apsnodei$};
%\node at (4.0em,6.2em) {$\apsnodei$};
\node at (4.0em,1.4em) {$\apsnodei$};
\end{tikzpicture}
\end{center}

We can stop the proof any time we have a single hypothesis, in which case we turn the proof box with a single hypotheses and a single conclusion into a single vertex.

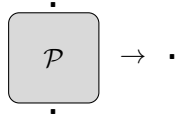
\begin{center}
\begin{tikzpicture}[scale=0.75]
\node[gpn] at (4em,4em) 	{$\mathcal{P}$};
\node at (4.0em,6.8em) {$\apsnodei$};
\node at (4.0em,1.4em) {$\apsnodei$};
\node at (8em,4em) {$\rightarrow$};
\node at (10em,4.2em) {$\apsnodei$};
\end{tikzpicture}
\end{center}

The complicated operation for backward chaining proof search is the elimination rule for linear implication. 
We are computing a structure $\mathcal{P}$ which hypotheses $x_1,\ldots, x_k$ and conclusion $z$. We select a unique hypothesis $y$ as the main formula, then split the remaining hypotheses into two groups $v_1,\ldots,v_n$ and $w_1,\ldots w_m$. We have $n+m+1 = k$, since each $x_i$ is in exactly one of the three groups. The $v_i$, together with the conclusion of the new link, will be the premisses of new structure $\mathcal{Q}$ whereas the $w_j$ will be the premisses of $\mathcal{R}$.

If this sounds complicated, it is essentially the $\multimap L$ rule from sequent calculus, where we similarly divide the formulas into $\Gamma$ (our $v$ formulas), $\Delta$ (our $w$ formulas) and $A\multimap B$ ($y$). 

\[
\infer[\multimap L]{\Gamma,\Delta,A\multimap B\vdash C}{\Delta\vdash A & \Gamma,B\vdash C}
\]

%\infer[\multimap E]{\Gamma,\Delta\vdash C}{\Gamma\multimap }

There are many possibilities here, since for $k$ formulas, there are $k$ choices for the main formula, then for the remaining $k-1$ formulas we assign them to one of two groups. This gives a total of $2^k$ different ways to apply this rule. However, this might still be an effective strategy for neural parsing. As indicated before, the backward chaining proof search procedure is essentially the same as the one of \citeasnoun{kmmt19ded} without explicit formula information at the vertices during the graph construction stage.

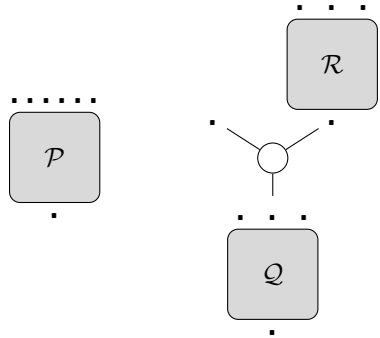
\begin{center}
\begin{tikzpicture}[scale=0.75]
\node[gpn] at (-8em,13.5em) 	{$\mathcal{P}$};
\node at (-10.0em,16.5em) {$\apsnodei$};
\node at (-9.2em,16.5em) {$\apsnodei$};
\node at (-8.4em,16.5em) {$\apsnodei$};
\node at (-7.6em,16.5em) {$\apsnodei$};
\node at (-6.8em,16.5em) {$\apsnodei$};
\node at (-6.0em,16.5em) {$\apsnodei$};
\node at (4.4em,21.2em) {$\apsnodei$};
\node at (6em,21.2em) {$\apsnodei$};
\node at (7.6em,21.2em) {$\apsnodei$};
\node at (1.4em,10.6em) {$\apsnodei$};
\node at (4.6em,10.6em) {$\apsnodei$};
%\node at (3.0em,6.2em) {$\apsnodei$};
%\node at (4.0em,6.2em) {$\apsnodei$};
\node at (-8.0em,10.7em) {$\apsnodei$};
%\node [gpn] at (6em,-3.3em) {$\Theta$};
\node [gpn] at (6em,18.2em) {$\mathcal{R}$};
%\node [pn] at (4.5em,7.6em) {$\Delta$};
\node [gpn] at (3em,7.6em) {$\mathcal{Q}$};
\node (ab2) at (3em,10.6em) {$\apsnodei$};
\node (ab) at (3em,4.8em) {$\apsnodei$};
\node (a) at (6em,15.4em) {$\apsnodei$};
\node (b) at (0em,15.4em) {$\apsnodei$};
\node[tns] (c) at (3em,13.468em) {};
%\node (clab) at (3em,7.668em) {$i$};
\draw (c) -- (ab2);
\draw (c) -- (a);
\draw (c) -- (b);
%
%\node (pa) at (6em,0em) {$\apsnodei$};
%\node[par] (pc) at (3em,1.732em) {};
%\node (pclab) at (3em,1.732em) {\textcolor{white}{$i$}};
%\draw (pc) -- (ab);
%\path[>=latex,->]  (pc) edge (pa);
%\draw (a) ..controls(-8em,38em)and(-8em,-5em).. (pc);
%\draw (b) to [out=130,in=210] (pc);
%\node (labl) at (3em,-2.5em) {$[\ldr I]$};
%%%%
\end{tikzpicture}
\end{center}

The final operation is the introduction rule. We start with a box $\mathcal{P}$ with premisses $x_1,\ldots,x_n$ for $n>0$, add a new hypothesis $x_0$ and attach a new link withdrawing this hypothesis. Like the graph expansion rules, this rule increases the number of hypotheses, but unlike the expansion rule it competes with the `stop' rule only when $n=1$. 

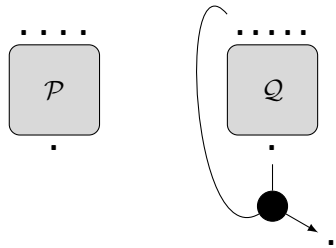
\begin{center}
\begin{tikzpicture}[scale=0.75]
\node at (-9.5em,10.6em) {$\apsnodei$};
\node at (-6.5em,10.6em) {$\apsnodei$};
\node at (-7.5em,10.6em) {$\apsnodei$};
\node at (-8.5em,10.6em) {$\apsnodei$};
\node at (-8em,4.8em) {$\apsnodei$};
\node[gpn] at (-8em,7.6em) 	{$\mathcal{P}$};
%\node [pn] at (4.5em,7.6em) {$\Delta$};
\node [gpn] at (3em,7.6em) {$\mathcal{Q}$};
\node (ab2) at (1.4em,10.6em) {$\apsnodei$};
\node at (2.2em,10.6em) {$\apsnodei$};
\node at (3.0em,10.6em) {$\apsnodei$};
\node at (3.8em,10.6em) {$\apsnodei$};
\node at (4.6em,10.6em) {$\apsnodei$};
\node (ab) at (3em,4.8em) {$\apsnodei$};
%\node (a) at (6em,15.4em) {$\apsnodei$};
%\node (b) at (0em,15.4em) {$\apsnodei$};
%\node[tns] (c) at (3em,13.468em) {};
%\node (clab) at (3em,7.668em) {$i$};
%\draw (c) -- (ab2);
%\draw (c) -- (a);
%\draw (c) -- (b);
%
\node (pa) at (6em,0em) {$\apsnodei$};
\node[par] (pc) at (3em,1.732em) {};
%\node (pclab) at (3em,1.732em) {\textcolor{white}{$i$}};
\draw (pc) -- (ab);
\path[>=latex,->]  (pc) edge (pa);
%\draw (a) ..controls(-8em,38em)and(-8em,-5em).. (pc);
\draw (ab2) to [out=130,in=210] (pc);
%\node (labl) at (3em,-2.5em) {$[\ldr I]$};
%%%%
\end{tikzpicture}
\end{center}

The backward chaining graph generation strategy therefore has two major drawbacks compared to the forward chaining strategy.
\begin{enumerate}
	\item it front-loads the combinatorics for the $\multimap E$ rule; where the forward chaining strategy needs to consider $n^2$ combinations for this rule (it selects a functor from $n$ roots and an argument from the $n-1$ remaining ones), the backward chaining strategy needs to consider $2^n$. We are also handicapped because we cannot use any formula-based ways to restrict our partitions, such as the `count check'.  
	\item it can more easily get stuck in a loop; in the forward chaining strategy, the $\multimap I$ rule always has the `stop' rule as an alternative, whereas this is not the case for backward chaining.
\end{enumerate}

For these reasons, the forward chaining strategy appears to be preferred, but ultimately we would need to compare the performance of both approaches, and study other graph construction methods which mix ideas from forward and backward chaining.

\end{document}